\documentclass[manuscript, screen]{acmart}
\usepackage{algorithmic}
\usepackage{array}
\usepackage[mathscr]{eucal}
\usepackage[caption=false,font=normalsize,labelfont=sf,textfont=sf]{subfig}
\usepackage{textcomp}
\usepackage{stfloats}
\usepackage{url}
\usepackage{verbatim}
\usepackage{graphicx}
\usepackage{multirow}
\usepackage{rotating}
\usepackage{color}
\allowdisplaybreaks

\newcommand{\mf}[1]{\mathbf{#1}}
\newcommand{\tr}{\mathrm{Tr}}
\newcommand{\inc}{\textbf{\checkmark}}
\newcommand{\incc}{\textbf{\checkmark}}
\newcommand{\notc}{$\mathbf{\times}$}

\usepackage[switch]{lineno}
\AtBeginDocument{%
  }

\setcopyright{acmlicensed}
\copyrightyear{2024}
\acmYear{2024}
\acmDOI{XXXXXXX.XXXXXXX}

\acmJournal{POMACS}
\acmVolume{37}
\acmNumber{4}
\acmArticle{111}
\acmMonth{8}

\begin{document}

\title{Nonnegative Matrix Factorization in Dimensionality Reduction: A Survey}

\author{Farid Saberi-Movahed}
\email{f.saberimovahed@kgut.ac.ir}
\affiliation{%
  \institution{Department of Applied Mathematics, Faculty of Sciences and Modern Technologies, Graduate University of Advanced Technology}
  \city{Kerman}
  \country{Kerman, Iran}
}

\author{Kamal Berahmand}
\affiliation{%
  \institution{School of Computer Sciences, Science and Engineering Faculty, Queensland University of Technology (QUT)}
  \city{Brisbane}
  \country{Brisbane, Australia}}
\email{kamal.berahmand@hdr.qut.edu.au}

\author{Razieh Sheikhpour}
\authornote{Corresponding author: Razieh Sheikhpour}
\affiliation{%
  \institution{Department of Computer Engineering, Faculty of Engineering, Ardakan University, P.O. Box 184}
  \city{Ardakan}
  \country{Ardakan, Iran}
}
\email{rsheikhpour@ardakan.ac.ir}

\author{Yuefeng Li}
\affiliation{%
  \institution{School of Computer Sciences, Science and Engineering Faculty, Queensland University of Technology (QUT)}
  \city{Brisbane}
  \country{Brisbane, Australia}}
\email{y2.li@qut.edu.au}

\author{Shirui Pan}
\affiliation{%
  \institution{School of Information and Communication Technology, Griffith University, Gold Coast 4222}
  \country{Australia}}
\email{s.pan@griffith.edu.au}

\renewcommand{\shortauthors}{Saberi-Movahed et al.}

\begin{abstract}
Dimensionality Reduction plays a pivotal role in improving feature learning accuracy and reducing training time by eliminating redundant features, noise, and irrelevant data. Nonnegative Matrix Factorization (NMF) has emerged as a popular and powerful method for dimensionality reduction. Despite its extensive use, there remains a need for a comprehensive analysis of NMF in the context of dimensionality reduction. To address this gap, this paper presents a comprehensive survey of NMF, focusing on its applications in both feature extraction and feature selection. We introduce a classification of dimensionality reduction, enhancing understanding of the underlying concepts. Subsequently, we delve into a thorough summary of diverse NMF approaches used for feature extraction and selection. Furthermore, we discuss the latest research trends and potential future directions of NMF in dimensionality reduction, aiming to highlight areas that need further exploration and development.
\end{abstract}


\received{... ... 2024}
\received[revised]{...}
\received[accepted]{...}

\maketitle

\section{Introduction}\label{sec1}

 Dimensionality reduction is a critical step in the machine learning process, as the dataset's dimension can significantly influence the performance of a machine learning algorithm \cite{xu2019review, anowar2021conceptual,cunningham2015linear,ayesha2020overview}. By discriminatively reducing the data size, the efficiency of the machine learning algorithm experiences a notable improvement. Consequently, dimension reduction has become an integral and indispensable step in both supervised and unsupervised learning tasks. The ultimate goal of dimensional reduction is to transform high-dimensional data into a lower-dimensional representation that preserves discriminatory power and relevant information. This process allows us to eliminate noise, remove redundant features, and focus on the most informative aspects of the data \cite{huang2023unsupervised,li2021semisupervised,alhassan2021review}.

 When it comes to reducing the dimensionality of high-dimensional data, two primary approaches stand out: feature extraction and feature selection. Through feature extraction, the original data is transformed into a new, reduced feature space while retaining the essential information content and eliminating irrelevant features \cite{nie2022discrete}. Conversely, feature selection aims to identify and select the most relevant subset of features from the input data that provide the most informative insights for a given problem \cite{saeys2007review}. Each of these approaches offers a variety of methods. Feature extraction is divided into two categories: linear and non-linear methods, while feature selection is further categorized into three types: filter methods, wrapper methods, and embedded methods.

 Nonnegative Matrix Factorization (NMF) is a powerful technique for dimension reduction, offering several advantages over traditional linear methods and other dimension reduction techniques \cite{gan2021nonnegative, he2011symmetric, yang2010linear,yi2020nonnegative}. One of its key strengths lies in its ability to handle non-negativity constraints, making it particularly well-suited for datasets with only positive values, such as images, text, and audio signals. By leveraging this property, NMF can extract parts-based and additive representations, revealing underlying patterns and features within the data \cite{7902136}. Moreover, NMF inherent sparsity-promoting nature allows it to automatically select relevant features, effectively reducing data dimensionality while preserving critical information. Unlike certain linear methods that might struggle with high-dimensional and complex datasets, NMF demonstrates robustness and scalability in managing such scenarios \cite{Pei7748469}. Additionally, NMF interpretability is another valuable aspect, as it empowers researchers to gain meaningful insights into the latent structure of the data, facilitating data exploration and analysis \cite{PENG2022571}. Overall, the combination of nonnegativity, sparsity, interpretability, and scalability renders NMF a versatile and compelling technique for dimension reduction tasks, providing an attractive alternative to other methods in the field.

 Over the past two decades, a plethora of NMF versions have emerged, each contributing to advancements in feature extraction and feature selection methods. However, a comprehensive review encompassing all of these contributions is currently lacking. To address this gap, this paper presents a focused summary article centered on the two primary methods of dimensionality reduction: feature extraction and feature selection.

 The paper begins by providing a classification of dimensionality reduction based on feature extraction and feature selection, with particular emphasis on NMF role in both approaches. Subsequently, a meticulous examination of NMF in the context of dimension reduction is presented. From the perspective of feature extraction, NMF is organized into four main categories: variety, regularization, generalized, and robust, with each category further branching out into sub-categories, effectively capturing the diverse applications of NMF in feature extraction. Additionally, the paper analyzes NMF from six different perspectives concerning feature selection: Standard NMF Problem, convex-NMF Problem, graph-Based, dual Graph-Based, sparsity, and orthogonality Constraint. This comprehensive exploration provides valuable insights into the various ways NMF can be effectively applied for feature selection in dimension reduction tasks. To achieve this, the review addresses four fundamental research questions:
\begin{itemize}
	\item What is NMF, and how does it function in dimension reduction?
	\item What are the various variations of NMF and their applications in dimension reduction?
	\item What are the advantages and limitations of NMF compared to other dimension reduction techniques?
	\item What are the future directions for NMF in dimension reduction?
	By addressing these questions, the review endeavors to illuminate the potential of NMF as a powerful and versatile tool in dimension reduction, while also identifying areas for future research and exploration.
\end{itemize}
The paper is organized into distinct sections to offer a comprehensive perspective on NMF in dimensionality reduction. In Section \ref{sec2}, we delve into various dimensionality reduction methods and their respective variants, aiming to comprehend their implications and advantages. Section \ref{sec3} offers an in-depth definition of NMF and explores its diverse methods employed in both feature extraction and feature selection. Future directions concerning NMF in dimensional reduction are examined in Section \ref{sec4}. In conclusion, we summarize the findings and implications of our study.

 \section{Dimensionality Reduction}\label{sec2}
\begin{figure*}[!h]
	\centering
	\includegraphics[scale=0.45]{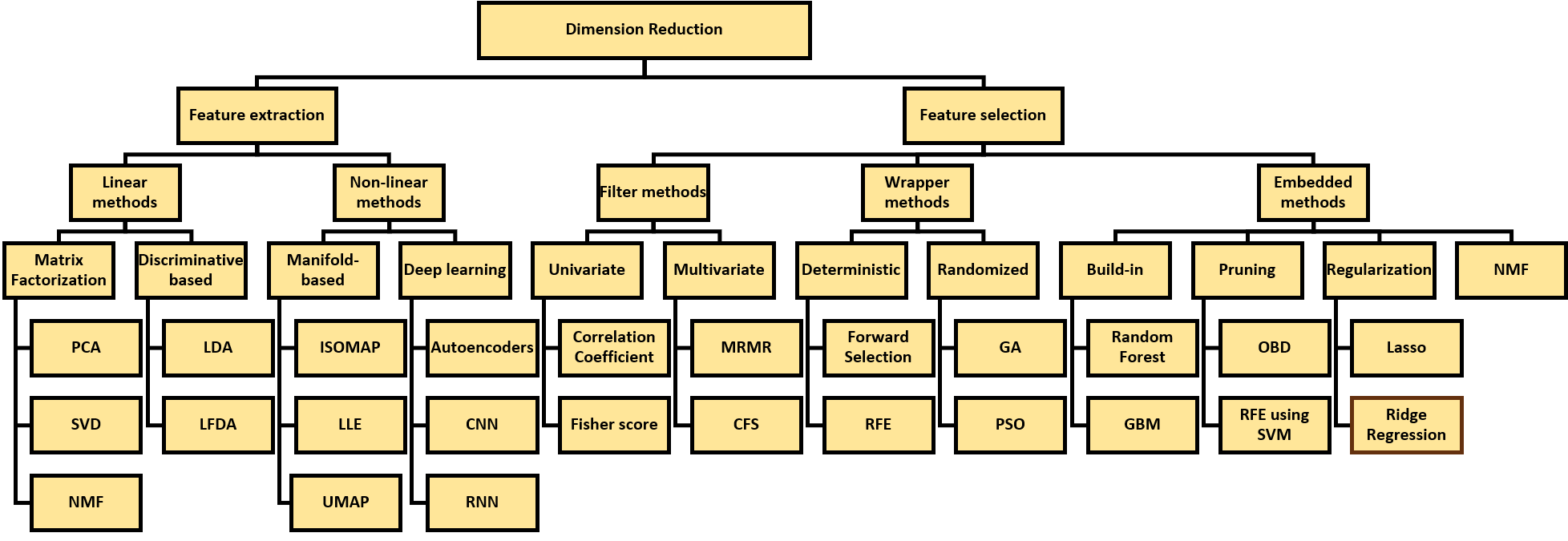}
	\caption{Categorizing dimensionality reduction techniques: a clear division into feature extraction and feature selection approaches}
\label{figDR}
	\end{figure*}

Dimensionality reduction is a technique used in machine learning to reduce the number of features while preserving as much relevant information as possible \cite{xu2019review,wang2022robust,nie2022discrete}. This technique improves the performance of machine learning algorithms, reduces the computational complexity of learning models, and prevents overfitting \cite{sheikhpour2020robust,eskandari2023online}. Feature extraction and feature selection are two main approaches in dimension reduction. In this paper, we present a classification of the dimensionality reduction methods as shown in Fig. \ref{figDR}.
Moreover, the most commonly notations used in this paper are demonstrated in Table \ref{notationlist}.

 \begin{table}[!h]
\centering
\caption{Description of the notations used in this paper.}
\scalebox{1}{\begin{tabular}{|l||l|}
\hline
Notations                         & Description                                       \\
\hline
$\mathbb{R}^{m \times n}$     & Set of matrices of size $m \times n$              \\
$\mathbb{R}_{+}^{m \times n}$     & Set of non-negative matrices of size $m \times n$ \\
$p$                               & Number of features                                \\
$q$                               & Number of samples                                 \\
$k$                               & Size of dimensionality reduction                  \\
$\mf{A}$ & Non-negative data matrix in $\mathbb{R}_+^{p\times q}$           \\
$\mf{B}$ & Basis matrix                                      \\
$\mf{C}$ & Coefficient representation matrix            \\
$\mathcal{A}$ & Non-negative data tensor \\
$\mf{W}$ & Transformation matrix for dimensionality reduction \\
$\mathbf{L}$ & Laplacian matrix of the data graph\\
$\widetilde{\mathbf{L}}$ & Laplacian matrix of the feature graph\\
$\mathbf{Z}^\pm$                  & Matrix with mixed sign values  elements                      \\
$\mf{Z}^T$ & Transpose of the matrix $\mf{Z}$\\
$\mf{I}_m$ & Identity matrix with the size $m$ \\
$\mf{0}_{m\times n}$ & Zero matrix with the size $m\times n$\\
$\mf{z}_{ij}$ &  Entries of the matrix $\mf{Z}$\\
$\mathrm{Tr}$ & Trace operator\\
$\Vert\cdot\Vert_{F}$ & Frobenius norm \\
$\Vert\cdot\Vert_{2,1}$ & $L_{2,1}$ norm \\
$\Vert\cdot\Vert_2$ & Vector Euclidean norm \\
$\{\mf{a}_1,\mf{a}_2,\ldots,\mf{a}_q\}$ & Set of data samples \\
$\{\mf{f}_1,\mf{f}_2,\ldots,\mf{f}_p\}$ & Set of feature vectors \\
$\mathrm{coni}\left(\mf{z}_{1}, \ldots, \mf{z}_{m}\right)$ & Conic hull of the set of vectors $\{\mf{z}_{1}, \ldots, \mf{z}_{m}\}$\\
$\alpha,\beta,\gamma$ &  Positive regularization parameters \\
\hline 
\end{tabular}}
\label{notationlist}
\end{table}
\subsection{Feature Extraction}
Feature extraction learns a low-dimensional combination of the original features and transforms the original features into a new set of features that keep the most important information in the dataset \cite{chao2019recent}. Feature extraction methods are divided into linear and nonlinear methods \cite{xu2019review,chao2019recent,ran2020general,wang2021unsupervised}.
\subsubsection{Linear Feature Extraction Methods}
Linear methods create a low-dimensional linear mapping of the original features \cite{cunningham2015linear,ayesha2020overview,anowar2021conceptual}. The linear methods can be classified into matrix factorization and discriminative-based methods.
\paragraph {Matrix Factorization-based Feature Extraction Methods}
Matrix factorization-based methods extract significant features by decomposing a high-dimensional data matrix into lower-dimensional matrices. These lower-dimensional representations retain essential data structure and relationships, resulting in more interpretable and compact feature sets \cite{cunningham2015linear}.

$\bullet$\,\,\textbf{Principal Component Analysis (PCA).}
PCA finds the orthogonal projection of the data onto a lower dimensional linear subspace such that the projected data have the maximum variance \cite{ayesha2020overview}. PCA can be formulated as:
\begin{equation}\label{PCArel1}
\max_{\mf{W}} \tr({\mf{W}^{T}\mf{S}\mf{W}})\,\,\
{\mathrm{subject\,\,to}\,\,\mf{W}^T\mf{W}=\mf{I}},
\end{equation}
where $\mf{S}=\mf{A}\mf{A}^{T}$ is the covariance matrix of data matrix $\mf{A}$.
In summary, PCA calculates the covariance matrix of data, and conducts an eigenvalue decomposition on this matrix to generate eigenvectors and eigenvalues. The eigenvalues are sorted in descending order and the eigenvectors corresponding to the highest eigenvalues are selected to form the transformation matrix $\mf{W}$.


 $\bullet$\,\,\textbf{Singular Value Decomposition (SVD).}
SVD is a matrix decomposition technique that can be used for feature extraction and is closely related to PCA \cite{ayesha2020overview}. SVD decomposes the data matrix $\mf{A}\in\mathbb{R}^{p\times q}$ as:
\begin{equation}\label{SVDrel1}
\mf{A}=\mf{U}\mf{Z}\mf{V}^{T},
\end{equation}
where $\mf{U}\in\mathbb{R}^{p\times p}$ and $\mf{V}\in\mathbb{R}^{q\times q}$ are two orthogonal matrices, and $\mf{Z}\in\mathbb{R}_+^{p\times q}$ is a rectangular diagonal matrix that holds the singular values of $\mf{A}$. For a given value of $k$, SVD can also finds a lower dimension of the matrix $\mf{A}$ by retaining only the top $k$ largest singular values and their corresponding singular vectors.

$\bullet$\,\,\textbf{Nonnegative Matrix Factorization (NMF).}
NMF is a matrix factorization-based feature extraction method utilized to decompose a non-negative data matrix into two lower-dimensional non-negative matrices \cite{paatero1994positive}. The details of NMF in feature extraction is fully reviewed in Section \ref{FENMF}.

\paragraph {Discriminative-based Feature Extraction Methods}
Discriminative-based methods identify the most informative and discriminative features from input data. The aim of these methods is to extract features from data that maximize the separation between classes while minimize the intra-class variations \cite{ding2016locality}.


$\bullet$\,\,\textbf{Linear Discriminant Analysis (LDA).}
LDA is a supervised feature extraction method
aimed at finding a linear transformation of the data that maximizes the separation between classes \cite{nie2020unsupervised,wang2021novel}. The objective function of LDA can be defined as:
\begin{equation}\label{LDArel1}
\max_\mf{W} { \frac{\tr(\mf{W}^T\mf{S}_b\mf{W})}{\tr(\mf{W}^T\mf{S}_w\mf{W})}},
\end{equation}
where $\mf{S}_b\in\mathbb{R}^{p\times p}$ is the within-class scatter matrix and $\mf{S}_w\in\mathbb{R}^{p\times p}$ is the between-class scatter matrix.

$\bullet$\,\,\textbf{Local Fisher Discriminant Analysis (LFDA).}
LFDA is a feature extraction method that combines the idea of LDA and locality-preserving projection (LPP) to find a transformation matrix that maximizes the Fisher discriminant ratio and retains the local structure of the data \cite{sugiyama2006local}. The transformation matrix of FLDA is defined as 

 \begin{equation}\label{LFDArel1}
\max_\mf{W} { \frac{\tr(\mf{W}^T\mf{\bar{S}}_b\mf{W})}{\tr(\mf{W}^T\mf{\bar{S}}_w\mf{W})}},
\end{equation}
where $\mf{\bar{S}}_b\in\mathbb{R}^{p\times p}$ is the local within class scatter matrix and $\mf{\bar{S}}_w\in\mathbb{R}^{p\times p}$ is the local between class scatter matrix.

\subsubsection{Nonlinear Feature Extraction Methods}
Nonlinear methods seeks to extract features from the original data by retaining the nonlinear relationships and structures in the data \cite{anowar2021conceptual}. These methods are divided into manifold-based methods and deep learning.
\paragraph{Manifold-based Feature Extraction Methods}
Manifold-based methods are utilized to reveal the underlying nonlinear structure of high-dimensional data that lies on or near a lower-dimensional manifold. These methods maintain the intrinsic geometry and relationships between data points to construct compact representations in a lower-dimensional space
\cite{chao2019recent,nie2021adaptive}.

$\bullet$\,\,\textbf{Isometric Feature Mapping (Isomap).}
Isomap is an unsupervised feature extraction method based on nonlinear manifold learning that aims to get the low-dimensional representation of data while retaining the geodesic distances and the manifold structure of data \cite{anowar2021conceptual}.

$\bullet$\,\,\textbf{Locally Linear Embedding (LLE).}
LLE is an unsupervised nonlinear feature extraction method which aims to retain the local data structure through linear combinations. It constructs a neighborhood graph and assumes that each data point can be represented as a convex linear combination of its nearest neighbors \cite{ayesha2020overview}.
For dimesionality reduction, LLE identifies the ${k}$-nearest neighbors of each data point $\mf{a}_i$, and computes the local weights for each data using the linear combination of its neighbors. This process creates a weight matrix $\mf{W}=[\mf{w}_{ij}]$ by minimizing the reconstruction error as shown in \eqref{LLEresl1}:
\begin{equation}\label{LLEresl1}
\min_\mf{W} { \sum_{i=1}^q \|\mf{a}_i - \sum_{j=1}^q \mf{w}_{i,j}\mf{a}_j\|_2^2},
\end{equation}
where $w_{i,j}$ is the weight between $\mf{a}_i$ and $\mf{a}_j$.
Finally, LLE calculates the low-dimensional embedding of data points $\mf{Z}$ to map each data point $\mf{a}_i$ from $p$-dimension to $\mf{z}_i$ in $k$-dimension by considering the following minimization problem:
\begin{equation}\label{LLEresl2}
\min_\mf{Z} { \sum_{i=1}^q \|\mf{z}_i - \sum_{j=1}^q\mf{w}_{i,j}\mf{z}_j\|_2^2}.
\end{equation}


$\bullet$\,\,\textbf{Uniform Manifold Approximation and Projection (UMAP).}
UMAP is a manifold learning technique for dimensionality reduction that can effectively maintain both global and local structures of data. It constructs a neighborhood graph and creates a topological representation of the high-dimensional data based on the graph. Then, it computes a low-dimensional layout of data representation and optimize the layout of the data representation in the low-dimensional space \cite{mcinnes2018umap}.
\paragraph{Deep Learning-based Methods for Feature Extraction}
Deep learning-based methods for feature extraction captures complex structures in high-dimensional data and project them into lower-dimensional space. They employ representation learning capabilities of deep neural networks to extract nonlinear features from data and learn data representations that maintain important information while reducing the dimensionality of data \cite{latif2021survey}.

$\bullet$\,\,\textbf{Autoencoders (AEs).}
Autoencoders learn compact representation of the input data, aiding in dimensionality reduction. The compact representation maintains the most important features and eliminates the irrelevant or noisy ones. Autoencoders consist of an encoder network that maps the input data to a lower-dimensional representation (encoding), and a decoder network that reconstructs the original input from the encoding \cite {latif2021survey} .

$\bullet$\,\,\textbf{Convolutional Neural Networks (CNNs).}
CNNs can be utilized for dimensionality reduction, especially in high-dimensional image data scenarios. They achieve this through a combination of convolution layers, pooling layers, and fully connected layers, where dimensionality reduction occures automatically as a result of the network operation and architecture \cite {latif2021survey}.

$\bullet$\,\,\textbf{Recurrent Neural Networks (RNNs).}
RNNs are a form of neural networks designed for processing time-series or sequential data, making them suitable for dimensionality reduction tasks. They have an internal memory to maintain processed information. RNNs conduct dimensionality reduction on sequential or time-series data by using temporal dependencies and their memory capabilities \cite {latif2021survey}.

\subsection{Feature Selection}
Feature selection aims to select a useful subset of features from a large number of original features to improve the learning performance \cite{zhang2020unsupervisedfeature,fan2021adaptive,huang2023unsupervised}. Feature selection methods are divided into filter, wrapper and embedded methods \cite{xu2021general,wang2021joint,zhang2020unsupervised,li2021semisupervised}.

 \subsubsection{Filter Feature Selection Methods}
Filter methods select the features based on a performance criterion prior to using the learning algorithm \cite{li2021semisupervised,karimi2023semiaco}. They are computationally simple, inexpensive, fast, and easily scalable for high dimensional datasets \cite{sheikhpour2023local}. Filters can be split into univariate and multivariate categories \cite{saeys2007review,xu2023graph}.
\paragraph{Univariate Filter Feature Selection Methods}
Univariate methods evaluate and rank the feature individually according to specific criteria and select the top-ranked features. These methods ignore feature dependencies, which are their main concern \cite{saeys2007review,wang2021semisupervised}. 
\paragraph{Multivariate Filter Feature Selection Methods}
Multivariate methods take into account the feature dependencies and evaluate the feature subsets according to joint influence of their features on the target variable \cite{saeys2007review,xu2023graph}. 
\subsubsection{Wrapper Feature Selection Methods}
Wrapper feature selection involves searching the feature space to generate candidate feature subsets using a search procedure, then assessing these subsets with a learning algorithm using an evaluation criterion \cite{zhang2020unsupervisedfeature,sheikhpour2023hessian}. Wrapper methods usually outperform filter methods \cite{sheikhpour2023hessian}, but are computationally expensive and have a higher risk of overfitting \cite{sheikhpour2023local,sheikhpour2023hessian,alhassan2021review}. Wrapper methods can be divided into deterministic and randomized categories.

 \paragraph{Deterministic Wrapper Feature Selection Methods}
Deterministic methods systematically and exhaustively explore the feature space to evaluate all possible feature combinations, ensuring the discovery of the optimal subset. However, they need assessing a large number of feature combinations, leading to significant computational costs \cite{saeys2007review}. 

 \paragraph{Randomized Wrapper Feature Selection Methods}
Randomized methods utilize randomization or heuristics to effectively search for optimal or near-optimal feature subsets without generating all possible combinations of feature sets. These methods are computationally more effective than the deterministic methods \cite{saeys2007review}.

 \subsubsection{Embedded Feature Selection Methods}
Embedded methods integrate feature selection into model building \cite{zhang2020unsupervisedfeature}. These methods are computationally more effective and less expensive than wrapper methods while they have similar performance \cite{sheikhpour2023hessian}. Embedded methods can be divided into regularization, built-in, pruning and NMF methods.
\paragraph{Regularization-based Embedded Feature Selection Methods}
Regularization-based methods combine feature selection with model training. These methods add penalty terms to the objective function of the model to reduce the model complexity and promote the sparsity \cite{huang2015supervised}. 

 \paragraph{Built-in Embedded Feature Selection Methods}
Built-in feature selection refers to feature selection methods that are integrated into specific learning algorithms \cite{huang2015supervised}. 

 \paragraph{Pruning Embedded Feature Selection Methods}
Pruning feature selection refer to methods utilized to decrease the complexity of a model by excluding unnecessary or redundant features. It creates a model with all features and then eliminate less important features \cite{huang2015supervised}. 

 \paragraph{Nonnegative Matrix Factorization (NMF)}
NMF decomposes a nonnegative data matrix into a basis matrix and a coefficient representation matrix \cite{paatero1994positive}. It serves as an embedded feature selection method by identifying a reduced set of latent features that retain the most important data patterns. 
The details of NMF in feature selection is discussed in Section \ref{FSNMF}.\\
A summery of dimensionality reduction methods is presented in Table \ref{tableDR}.

 {\renewcommand{\baselinestretch}{1}
\begin{table*}[!htbp]
\centering
\caption{A summary of the dimensionality reduction methods.}
\scalebox{0.65}{\begin{tabular}{|c||c|c|c|c|l|}
\hline
\multicolumn{1}{|>{\centering}m{2cm}||}{\textbf{Dimensionality Reduction Method}} &
\textbf{Type} &	
\textbf{Method	} &
\textbf{Formulation} &
\textbf{Description} \\
\hline
\multirow{24}{*}{\large\bfseries\begin{turn}{90} Feature Extraction \end{turn}} &
\multirow{10}{*}{\bfseries Linear} &
PCA \cite{ayesha2020overview} &
$\max_{\mf{W}^T\mf{W}=\mf{I}}\tr(\mf{W}^T\mf{A}\mf{W})$ &
\multicolumn{1}{>{\centering}m{8cm}|}{Maximizes the variance of the data projected onto the principal components} \\[2ex]
\cline{3-5}
&
&
SVD \cite{ayesha2020overview} &
$\mf{A}=\mf{UZ{V}^T}$ &
Factorizes the data matrix into three matrices. \\[2ex]
\cline{3-5}
&
&
NMF \cite{paatero1994positive} &
$\min_{\mf{B,C}\geq 0} \Vert \mf{A}-\mf{BC}\Vert_F^2$ &
\multicolumn{1}{>{\centering}m{8cm}|}{Represents the data matrix as the product of two non-negative matrices.} \\[2ex]
\cline{3-5}
&
&
LDA \cite{wang2021novel} &
$\max_{\mf{W}}\frac{\tr(\mf{W}^T\mf{S}_b\mf{W})}{\tr(\mf{W}^T\mf{S}_w\mf{W})}$ &
\multicolumn{1}{>{\centering}m{8cm}|}{Maximizes the ratio of between-class scatter to within-class scatter.} \\[2ex]
\cline{3-5}
&
&
LFDA \cite{sugiyama2006local} &
$\max_{\mf{W}}\frac{\tr(\mf{W}^T\bar{\mf{S}}_b\mf{W})}{\tr(\mf{W}^T\bar{\mf{S}}_w\mf{W})}$ &
\multicolumn{1}{>{\centering}m{8cm}|}{Combines the local structure of data with the discriminative information of LDA.} \\[2ex]
\cline{2-5}
&
\multirow{14}{*}{\bfseries Nonlinear} &
Isomap \cite{anowar2021conceptual} &
$\sum_{i=1}^{q} \sum_{j=1}^{q} (d_G({x}_i,{x}_j) - \|\mf{y}_i - \mf{y}_j\|_2)^2$ &
\multicolumn{1}{>{\centering}m{8cm}|}{Approximates the pairwise geodesic distances on the underlying manifold.} \\[2ex]
\cline{3-5}
&
&
LLE \cite{ayesha2020overview} &
$ \min_\mf{Z}{\sum_{i=1}^q \|\mf{z}_i - \sum_{j=1}^q \mf{w}_{i,j}\mf{z}_j\|_2^2}$ &

 \multicolumn{1}{>{\centering}m{8cm}|}{Reconstructs each data point as a weighted sum of its neighbors and preserves the local linear structure.} \\[2ex]
\cline{3-5}
&
&
UMAP \cite{mcinnes2018umap} &
- &
\multicolumn{1}{>{\centering}m{8cm}|}{Retains local and global structures by minimizing the cross-entropy between high-dimensional and low-dimensional distances.} \\[2ex]
\cline{3-5}
&
&
AEs \cite{latif2021survey} &
$\min \sum_{i=1}^{q} \|a_i - a'_i\|_2^2 $ &
\multicolumn{1}{>{\centering}m{8cm}|}{Consist of encoder and decoder networks and learn a compressed representation of the data.} \\[2ex]
\cline{3-5}
&
&
CNN \cite{latif2021survey} &
- &
\multicolumn{1}{>{\centering}m{8cm}|}{Uses convolutional layers and pooling to extract features from images and visual data.} \\[2ex]
\cline{3-5}
&
&
RNN \cite{latif2021survey} &
- &
\multicolumn{1}{>{\centering}m{8cm}|}{Has an internal memory to maintain the processed information and is suitable for sequential data.} \\[2ex]
\hline\hline
\multirow{30}{*}{\large\bfseries\begin{turn}{90} Feature Selection \end{turn}} &
\multirow{12}{*}{\bfseries Filter} &
Correlation Coefficient \cite{alhassan2021review} &
$R_i = \frac{\sum_{i}(\mf{a}_i - \mf{\bar{a}})(\mf{y}_i - \mf{\bar{y}})}{\sqrt{\sum_{i}(\mf{a}_i - \mf{\bar{a}})^2 \sum_{i}(\mf{y}_i-\mf{\bar{y}})^2}}$ &
\multicolumn{1}{>{\centering}m{8cm}|}{Calculates linear correlation between two features.} \\[2ex]

\cline{3-5}
&
&
Fisher Score \cite{alhassan2021review} &
$R_{j} = \frac{\sum_{i=1}^{c} q_i (\mu_j^i - \mu_j)^2}{\sum_{i=1}^{c} q_i (\sigma_j^i)^2}
$ & \multicolumn{1}{>{\centering}m{8cm}|}{Ranks features based on their discriminating ability.}\\[2ex]
\cline{3-5}
&
&
MRMR \cite{ding2005minimum} &$ \begin{aligned}
\max_{a_i} \left[ \frac{1}{|F|} \sum_{a_i \in F} I(a_i,c) -
\frac{1}{|F|^2} \sum_{a_i,a_j \in F} I(a_i,a_j) \right]
\end{aligned}$ &
\multicolumn{1}{>{\centering}m{8cm}|}{Maximizes the relevance and minimizes redundancy.} \\[2ex]
\cline{3-5}
&
&
CFS \cite{hall1999correlation} &
$\text{Merit}_F = \frac{k (\bar{r}_{ci})}{\sqrt{k + k(k-1) \bar{r}_{ii}}}$ &
\multicolumn{1}{>{\centering}m{8cm}|}{Assesses feature subsets based on correlation with the target and redundancy among features.} \\[2ex]
\cline{2-5}
&
\multirow{6}{*}{\bfseries Wrapper} &
Forward Selection \cite{deng1998greediness}&
- &
\multicolumn{1}{>{\centering}m{8cm}|}{Iteratively adds features that improve model performance.} \\[2ex]
\cline{3-5}
&
&
RFE \cite{guyon2002gene} & - &
\multicolumn{1}{>{\centering}m{8cm}|}{Recursively removes the least important features.} \\[2ex]
\cline{3-5}
&
&
GA \cite{alhassan2021review} & - &
\multicolumn{1}{>{\centering}m{8cm}|}{Utilizes principles of evolution to select optimal features.} \\[2ex]
\cline{3-5}
&
&
PSO \cite{alhassan2021review} & - &
\multicolumn{1}{>{\centering}m{8cm}|}{Utilizes swarm intelligence to find feature subsets.} \\[2ex]

\cline{2-5}
&
\multirow{12}{*}{\bfseries Embedded} &
Lasso \cite{alpaydin2014introduction} &
$\min_{\mf{w}} \|\mf{Y} - \mf{Aw}\|_2^2 + \lambda \|\mf{w}\|_1$ &
\multicolumn{1}{>{\centering}m{8cm}|}{Linear regression with L1 regularization.} \\[2ex]
\cline{3-5}
&
&
Ridge Regression \cite{alpaydin2014introduction} &
$\min_{\mf{w}} \|\mf{Y} - \mf{Aw}\|_2^2 + \lambda \|\mf {w}\|_2^2$ &
\multicolumn{1}{>{\centering}m{8cm}|}{Linear regression with L2 regularization..} \\[2ex]
\cline{3-5}
&
&
GBM \cite{friedman2001greedy} & - &
\multicolumn{1}{>{\centering}m{8cm}|}{Builds decision trees sequentially where each tree correct the errors of the previous one.} \\[2ex]
\cline{3-5}
&
&
Rndom Forest \cite{alpaydin2014introduction} & - &
\multicolumn{1}{>{\centering}m{8cm}|}{Builds multiple decision trees and uses feature importance from the trees.} \\[2ex]
\cline{3-5}
&
& OBD \cite{le1990optimal} & - &
\multicolumn{1}{>{\centering}m{8cm}|}{Removes the least important weights in a neural network to prune the network and improve generalization.} \\[2ex]
\cline{3-5}
&
&
RFE using SVM \cite{guyon2002gene} & - &
\multicolumn{1}{>{\centering}m{8cm}|}{Recursively removes features based on SVM weights.} \\[2ex]
\cline{3-5}
&
&
NMF \cite{paatero1994positive} & $\min_{\mf{B,C}\geq 0}\Vert \mf{A}-\mf{BC}\Vert_F^2$ &
\multicolumn{1}{>{\centering}m{8cm}|}{Uses the product of two non-negative matrices to represent the data matrix.} \\[2ex]
\hline
\end{tabular}}
\label{tableDR}
\end{table*}}

\section{NMF in Dimensionality Reduction}\label{sec3}

 \subsection{Background of NMF}
Non-negative Matrix Factorization (NMF) is a mathematical approach, which stems particularly from the concept of matrix factorization. It aims to discover a reduced representation--specifically, a low-rank, part-based representation--of a nonnegative matrix \cite{lee2000algorithms}. Let $\mf{A}= [\mf{a}_1,\mf{a}_2,\ldots,\mf{a}_q]$ be a non-negative data matrix in $\mathbb{R}_+^{p\times q}$, where $p$ is the dimension and $q$ is the number of data samples. The goal of NMF is to break down $\mf{A}$ into two low-rank matrices: a basis matrix $\mf{B}\in\mathbb{R}_+^{p\times k}$ and a coefficient representation matrix $\mf{C}\in\mathbb{R}_+^{k\times q}$, where $k$ is called the dimensionality reduction parameter and is much smaller than $p$ and $q$, i.e., $k\ll\min\{p,q\}$. Formally, the NMF problem is expressed as follows:
\begin{equation}\label{lossnmf}
	\min_{\mf{B}\in\mathbb{R}_+^{p\times k}, \mf{C}\in\mathbb{R}_+^{k\times q}} \mathcal{L}(\mf{A},\mf{BC}),
\end{equation}
where $\mathcal{L}$ represents the loss function that evaluates the similarity between the data matrix $\mf{A}$ and the factorization $\mf{BC}$.

The loss function of NMF as provided in Problem \eqref{lossnmf} can be formulated using a mathematical metric that measures the extent of dissimilarity between the elements of the initial data matrix $\mf{A}$ and the elements of the factorization $\mf{BC}$ derived from NMF.
Among the loss functions effectively applied in NMF, the $\beta$-divergence functions stand out. Previous research demonstrates that incorporating the $\beta$-divergence into NMF can enhance its applicability to handle various data distributions such as Gaussian, Poisson, and Gamma \cite{fevotte2011algorithms}. Considering the flexibility and robustness offered by the $\beta$-divergence loss functions, their utilization in NMF can lead to meaningful factorizations that closely match the inherent data characteristics \cite{marmin2023majorization}.
Let $\beta\in\mathbb{R}$. The $\beta$-divergence function $d_\beta$ is defined for any $a$ and $b$ in $\mathbb{R}_+\setminus\{0\}$ as follows \cite{yuan2023beta}:
\begin{equation}
	d_\beta(a,b)=
	\begin{cases}
		\frac{a}{b}-\mathrm{log}\left(\frac{a}{b}\right)-1, & \beta = 0,\\
		a(\mathrm{log} a-\mathrm{log} b)+(b-a), & \beta = 1,\\
		\frac{a^\beta+(\beta-1)b^\beta+\beta ab^{\beta-1}}{\beta(\beta-1)}, & \beta\in\mathbb{R}_+\setminus\{0,1\}.
	\end{cases}
\end{equation}
In consequence, the NMF loss function utilizing the $\beta$-divergence function can be formulated on an element-wise form, as presented below:
\begin{equation}\label{lossnmf1}
	\mathcal{L}(\mf{A},\mf{BC})=\sum_{i=1}^p\sum_{j=1}^qd_\beta(\mf{a}_{ij}, \sum_{z=1}^k\mf{b}_{iz}\mf{c}_{zj}).
\end{equation}
It is noteworthy to state that the $\beta$-divergence function is strictly convex within the range of $\beta\in [1,2]$ \cite{10.1162NECOa00168}. Furthermore, the $\beta$-divergence function corresponds to the Euclidean distance (at $\beta = 2$), the generalized Kullback-Leibler divergence (at $\beta = 1$), and the Itakura-Saito divergence (at $\beta = 0$).

This paper centers around the specific variant of the $\beta$-divergence, where $\beta$ is set to 2. Under this circumstance, the NMF problem takes on the well-known form of NMF using the Frobenius norm, and this presentation formulates the NMF problem as follows:
\begin{equation}\label{pnmf}
	\min_{\mf{B}\in\mathbb{R}_+^{p\times k}, \mf{C}\in\mathbb{R}_+^{k\times q}} \Vert \mf{A}-\mf{BC}\Vert_F^2.
\end{equation}
Finally, it is worth mentioning that over the past decade, numerous optimization algorithms have been developed to address NMF problems. Notable among them are Multiplicative Update Rules (MUR), Alternating Non-Negative Least Squares Method (ANLS), Hierarchical Alternating Least Squares (HALS), and Alternating Direction Multiplier Method (ADMM). For a thorough examination of these methods, please refer to \cite{gan2021nonnegative}.

\subsection{Taxonomy of NMF in Dimensionality Reduction}
Thanks to the capability of NMF to unveil underlying structures present in data, it becomes applicable to two essential approaches in dimensionality reduction: feature extraction and feature selection. In the upcoming section, we delve into the capabilities of NMF within these categories, which reveals how diverse NMF variants provide valuable perspectives on the process of decreasing the dimensionality of high-dimensional data.

 \subsubsection{NMF in Feature Extraction}\label{FENMF}
The current approaches utilizing NMF for feature extraction can be organized into the four following categories.
(1) \textbf{Variants of NMF:} which refers to diverse modifications or adjustments introduced to the original framework of NMF.
(2) \textbf{Regularized NMF:} which involves incorporating regularization techniques within the NMF framework to enhance the quality and robustness of the NMF factorization process.
(3) \textbf{Generalized NMF:} Under this category, the NMF structure is expanded to deal with a wider range of data types such as multi-dimensional data and specialized domains such as hierarchically learning.
(4) \textbf{Robust NMF:} This category refers to a distinct variation of NMF, which is adapted to efficiently manage data that is corrupted by noise and corruption.
The aforementioned categories are illustrated in Figure \eqref{figFigNMFvarints}. Additionally, comprehensive discussions for each of these categories are presented below.
\begin{figure*}[!htbp]
	\centering
	\includegraphics[scale=0.6]{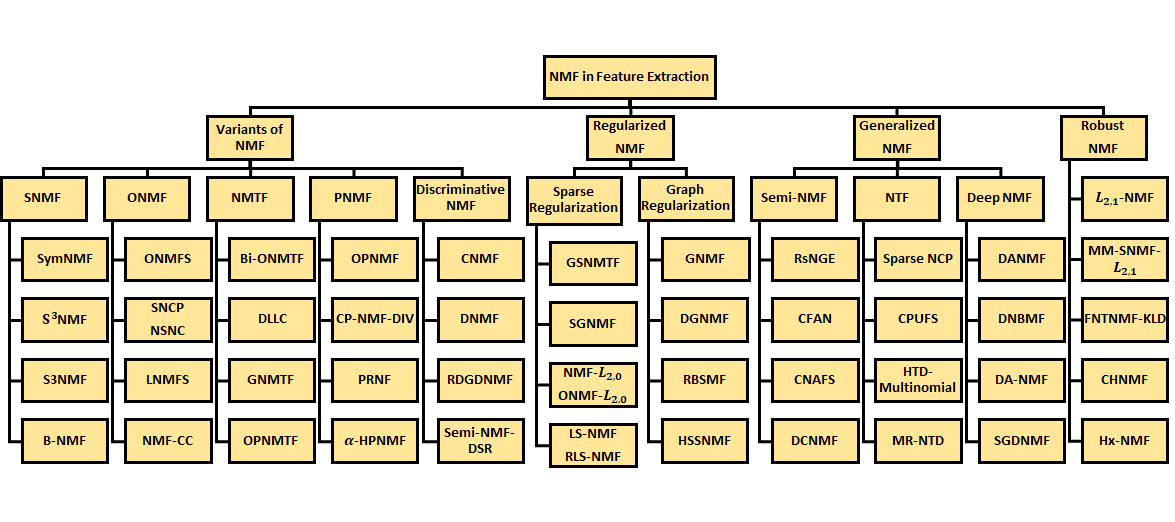}
	\caption{Classifying current NMF approaches for feature extraction into four groups: Variants of NMF, Regularized NMF, Generalized NMF, and Robust NMF.}
	\label{figFigNMFvarints}
\end{figure*}

 $\bullet$\,\,\textbf{Variants of NMF.} This section delves into a range of well-known NMF variants, each representing a unique adjustment to the original NMF framework. Some of these variants, such as Symmetric Nonnegative Matrix Factorization (SNMF), Orthogonal NMF (ONMF), Nonnegative Matrix Tri-Factorization (NMTF), Projective NMF (PNMF), and Discriminative NMF, are reviewed in the following.

 \paragraph{SNMF}
It is a variation of NMF that makes use of the symmetry constraint on the NMF framework \cite{he2011symmetric}. Consider $\mathbf{S}\in\mathbb{R}_+^{q\times q}$ as the similarity matrix produced using the data matrix $\mathbf{A}$. Mathematically, the purpose of SNMF is to express the symmetric matrix $\mf{S}$ in the form of $\mf{B}\mf{B}^T$, which is equivalent to say that
\begin{equation}\label{psnmf}
	\min_{\mf{B}\in\mathbb{R}_+^{q\times k}} \Vert \mf{S}-\mf{B}\mf{B}^T\Vert_F^2,
\end{equation}
where $k\ll q$. When it comes to clustering data that cannot be linearly separated, SNMF is a more suitable choice compared to NMF, thanks to its ability to handle non-linear relationships effectively \cite{kuang2015symnmf}. It is also shown that there is an equivalence between SNMF and the kernel $k$-means clustering when the matrix $\mf{B}$ satisfies the orthogonality constraint $\mf{B}^T\mf{B}=\mf{I}_k$ \cite{ding2005equivalence}.

 In recent years, there has been significant effort devoted to advancing the utilization of SNMF and rectify its limitations. As an example, Kuang et al. \cite{kuang2015symnmf} presented the SymNMF approach, which serves as a means of approximating a low-rank version of the similarity matrix used in graph clustering. This method is well-suited for the clustering of data points located within both linear and nonlinear manifolds. Inspired by ensemble clustering, Jia et al. \cite{jia2021selfsupervised} introduced the self-supervised SNMF (S$^{3}$NMF) approach, designed to mitigate the sensitivity of SNMF to variable initialization. Qin et al. \cite{9543530} presented S3NMF, a semi-supervised adaptation of SNMF. Within this approach, a block-diagonal strategy is utilized to construct the similarity matrix, a crucial step that significantly enhances the generation of an optimal assignment matrix for SNMF.
Most recently, Li et al. \cite{9606619} proposed the transformation of Problem \eqref{psnmf} into a penalized NMF problem as follows:
\begin{equation}\label{psnmf1}
	\min_{\mf{B}\in\mathbb{R}_+^{q\times k},\mf{V}\in\mathbb{R}_+^{q\times k}} \Vert \mf{S}-\mf{B}\mf{V}^T\Vert_F^2+\alpha\Vert\mf{B}-\mf{V}\Vert_F^2.
\end{equation}
Here, $\alpha$ is a non-negative penalty parameter. Subsequently, the authors of \cite{9606619} assessed the effectiveness of three alternating-type algorithms, namely SymANLS, SymHALS, and A-SymHALS, in solving the SNMF problem using the newly devised framework \eqref{psnmf1}. SNMF has another specific merit in handling symmetric data structures like graphs or networks, as it can identify more meaningful and comprehensible characteristics. As an instance, Liu et al. \cite{10072010} introduced an approach named B-NMF, which formulates a bi-regularized version of SNMF to create interpretable latent factor matrices for addressing large-scale undirected networks.
	Further examples of SNMF can be explored by referring to the following references:
	fast SNMF \cite{zhu2018dropping},
	CDSSNMF \cite{belachew2019efficient},
	and $\alpha,\beta$-SNMF/$\alpha,\beta$-GSNMF \cite{luo2021symmetric}.

\paragraph{ONMF}
It represents a deviation from traditional NMF, introducing an orthogonality constraint to either the basis matrix $\mf{B}$ or the coefficient representation matrix $\mf{C}$. It is demonstrated that ONMF is equivalent to the $k$-means clustering \cite{ding2006orthogonal}. In general, the optimization problem for ONMF, defined by Ding et al. \cite{ding2006orthogonal}, is expressed as follows:
\begin{align}\label{ponmf}
	\min_{\mf{B}\in\mathbb{R}_+^{p\times k}, \mf{C}\in\mathbb{R}_+^{k\times q}}\Vert \mf{A}-\mf{BC}\Vert_F^2,\\
	\mathrm{subject\,\,to}\,\,\mf{C}\mf{C}^T=\mf{I}_k,\nonumber
\end{align}
where $k\ll\min\{p,q\}$. Another variant of NMF that relies on the orthogonality is Bi-Orthogonal NMF (Bi-ONMF) \cite{ding2006orthogonal} which involves constraining both $\mf{B}$ and $\mf{C}$ to be orthogonal, which results in the following constrained optimization problem:
\begin{align}\label{pdonmf}
	&\min_{\mf{B}\in\mathbb{R}_+^{p\times k}, \mf{C}\in\mathbb{R}_+^{k\times q}}\Vert \mf{A}-\mf{BC}\Vert_F^2\\
	&\mathrm{subject\,\,to}\quad\mf{B}^T\mf{B}=\mf{I}_k,\quad \mf{C}\mf{C}^T=\mf{I}_k,\nonumber
\end{align}
where $k\ll\min\{p,q\}$. Here, the bi-orthogonality constraint on $\mathbf{B}$ and $\mathbf{C}$ is analogous to cluster the rows or columns of the original data matrix, respectively. In fact, by imposing this bi-orthogonality in \eqref{pdonmf}, Bi-ONMF aims to reduce the redundancy between the vectors in $\mf{B}$ and $\mf{C}$ simultaneously.

 Numerous enhancements have been suggested in the context of ONMF, with differences in their approaches for addressing the orthogonality condition in solving ONMF problem or leveraging it to enhance the efficiency of ONMF. For example,
Asteris et al. \cite{asteris2015orthogonal} introduced a novel ONMF framework, called ONMFS, with the goal of providing a provable solution for ONMF by solving a low-dimensional non-negative PCA problem.
He et al. \cite{he2020lowrank} introduced a low-rank NMF technique, known as LNMFS, which incorporates an orthogonality constraint on the basis matrix under the assumption that it is a member of a Stiefel manifold.
Inspired by the improved clustering performance achieved by applying the orthogonality conditions to both the basis and coefficient representation matrices, Liang et al. \cite{liang2020multi} introduced an ONMF model, called NMF-CC, which employs the co-orthogonal constraints on both these matrices to address the problem of multi-view clustering.
Wang et al. \cite{wang2021clustering} introduced two innovative ONMF variations, SNCP and NSNC. These alternatives convert the orthogonality constraint into a set of non-convex penalty constraints, potentially enhancing the efficiency of these two proposed variants when compared to the traditional ONMF approach.
Other noteworthy applications of ONMF are discussed in the literature by discriminative ONMF \cite{li2014discriminative}, KOGNMF \cite{tolic2018nonlinear}, sparse ONMF \cite{li2020fast}, and ONMF-apx \cite{pmlr-v130-charikar21a}.

 \paragraph{NMTF}
Ding et al. \cite{ding2006orthogonal} were the first to introduce NMTF and its bi-orthogonal variant, with the aim of simultaneously clustering words and documents. From an algebraic perspective, NMTF decomposes the data into three nonnegative matrices, the basis matrix $\mf{B}$, the coefficient representation matrix $\mf{C}$, and the block matrix $\mf{D}$, using the following problem:
\begin{equation}\label{ptrinmf}
	\min_{\mf{B}\in\mathbb{R}_+^{p\times k}, \mf{D}\in\mathbb{R}_+^{k\times l}, \mf{C}\in\mathbb{R}_+^{l\times q}} \Vert \mf{A}-\mf{BDC}\Vert_F^2,
\end{equation}
where $k, l\ll \min\{p,q\}$. Compared to NMF, where the value of the dimensionality reduction parameter $k$ indicates the number of columns in the basic matrix and the number of rows in the coefficient matrix, the NMTF framework employs an alternate strategy. In simpler terms, within the NMTF framework, the number of columns in the basic matrix is referred to as $k$, while the number of rows in the coefficient matrix is referred to as $l$. Therefore, the NMTF framework, which incorporates the additional matrix $\mf{D}$, allows for more flexibility in the size reduction compared to NMF.

 Another achievement made possible by NMTF lies within the co-clustering problem. Conventional clustering techniques, such as NMF, are typically employed for one-sided clustering, aiming to cluster data by leveraging the similarities observed among data samples. However, these one-sided clustering methods tend to neglect the inherent duality that exists between data samples and features \cite{gu2009co}. To address this limitation, co-clustering techniques have been established and proven to outperform the conventional approach of one-sided clustering \cite{lin2019overview}. An example of such techniques can be found in the bi-orthogonal version of NMTF (Bi-ONMTF) \cite{ding2006orthogonal}, through which a co-clustering problem is established as follows:
\begin{align}\label{Bi-OTNMF}
	&\min_{\mf{B}\in\mathbb{R}_+^{p\times k}, \mf{D}\in\mathbb{R}_+^{k\times l}, \mf{C}\in\mathbb{R}_+^{l\times q}} \Vert \mf{A}-\mf{BDC}\Vert_F^2\\
	&\mathrm{subject\,\,to}\quad\mf{B}^T\mf{B}=\mf{I}_k,\quad \mf{C}\mf{C}^T=\mf{I}_l,\nonumber
\end{align}
where $k, l\ll \min\{p,q\}$.
In Bi-ONMTF, the matrices $\mathbf{B}$ and $\mathbf{C}$ function as indicators to identify the clustering for features and data samples, respectively. Additionally, these matrices are subject to constraints of orthogonality and nonnegativity, which facilities the simultaneous hard co-clustering of both features and data samples in data \cite{ding2006orthogonal}.
In line with this, Wang and Guo \cite{wang2017robust} presented the DLLC approach, which combines the geometric characteristics of both features and data samples within the data into the Bi-ONMTF problem defined in \eqref{Bi-OTNMF}.
Jin et al. \cite{Jin2019} suggested the GNMTF method, a graph-regularized adaptation of NMTF, to address the community detection problem.
This method makes use of the intrinsic geometric traits of the network to discern the community membership of nodes and grasp the interconnections among diverse communities.
Hoseinipour et al. \cite{hoseinipour2023orthogonal} propounded a new version of Bi-ONMTF, known as OPNMTF, in which the $\alpha$-divergence function \cite{cichocki2008nonnegative} is employed to manage the bi-orthogonality constraint, instead of utilizing the Frobenius norm. This integration allows for greater adaptability in selecting divergence metrics by adjusting the $\alpha$ value.
Some other improvements of NMTF include the following: NMTF for co-clustering \cite{delbuono2015nonnegative}, PNMT \cite{wang2017penalized}, CNMTF \cite{peng2020robust}, and triFastSTMF \cite{tri10156842}.

\paragraph{PNMF}
The concept of orthogonal projection forms the foundation for defining the PNMF method \cite{yuan2005projective}, which can be described as a projective adaptation of NMF. The goal of PNMF is to minimize the error between the data matrix and its orthogonal projection onto the subspace constructed by the data samples. Moreover, when compared with the conventional NMF technique, PNMF leverages the following advantages \cite{yang2007projective,yang2010linear}: (1) a significantly sparse approximation, (2) an almost orthogonal factorization matrix, (3) reduced computational cost in the learning process, and (4) a close connection to the soft $k$-means clustering. The optimization problem for PNMF can be stated as follows:
\begin{align}\label{pornmf}
	&\min_{\mf{B}\in\mathbb{R}_+^{p\times k}}\Vert \mf{A}-\mf{B}\mf{B}^T\mf{A}\Vert_F^2,
\end{align}
where $k\ll\min\{p,q\}$, and the matrix $\mf{B}\mf{B}^T$ is the operator that projects the original data matrix. In the initial version of PNMF, the necessity for the orthogonality condition on the matrix $\mf{B}$ to establish an orthogonal projection was overlooked. In this case, the learning process of PNMF yields a matrix $\mf{B}$ which is nearly orthogonal, although not precisely. Later on, Yang and Oja \cite{yang2010linear} explored the impact of imposing the orthogonality condition $\mf{B}^T\mf{B}=\mf{I}_k$ on improving the optimization solution for PNMF. Following this, they developed the non-linear version of PNMF, named OPNMF, wherein the loss function is defined based on either the Frobenius norm or the generalized Kullback-Leibler divergence, and it is also subject to the constraint of orthogonality on $\mf{B}$.
To solve the PNMF problem using the Kullback-Leibler Divergence loss function, Hu et al. \cite{hu2014convergent} introduced an approach called CP-NMF-DIV. This method not only demonstrates superior performance compared to NMF but also generates basis matrices that exhibit improved levels of both orthogonality and sparsity.
Besides, Lu et al. \cite{lu2016projective} designed a robust adaptation of PNMF known as PRNF. It integrates a similarity graph regularizer to capture the inherent data structure and employs sparsity constraints based on the $L_p$ and $L_{2,p}$ matrix norms to enhance the resilience of PNMF.
Belachew and Del Buono \cite{belachew2020hybrid} formulated the loss function of PNMF using the $\alpha$-divergence function and subsequently introduced the $\alpha$-HPNMF method to solve this proposed framework.
Additional instances of PNMF can be found in the works of $\alpha$-PNMF \cite{yuan2009advances}, BPNMF \cite{zhang2014boxconstrained}, and BLO-PNMF \cite{chen2023bilevel}.

 \paragraph{Discriminative NMF}
Prior studies have shown that combining unlabeled data with a small amount of labeled data as a discriminative constraint can lead to significant improvements in learning accuracy of NMF. This fusion has given rise to a variant of NMF known as Discriminative NMF, which effectively utilizes both the intrinsic patterns extracted through NMF and the information offered by labeled data.
One leading example of Discriminative NMF is Constrained NMF (CNMF) established by Liu et al. \cite{liu2011constrained}. Suppose that the $r$ data samples $\mf{a}_1,\ldots,\mf{a}_r$ have labels and the rest of $q-r$ samples are unlabeled. Also, let the total number of these labels be $s$. In order to incorporate the available label information, CNMF exploits the following constraint matrix:
\begin{equation}\label{defM}
	\mf{M}=\left[
	\begin{array}{cc}
		\mf{U} & \mf{0}_{s\times (q-r)}\\
		\mf{0}_{(q-r)\times r} & \mf{I}_{q-r}
	\end{array}
	\right]\in\mathbb{R}_+^{(q-r+s)\times q},
\end{equation}
in which $\mf{U}=[\mf{u}_{ij}]\in\mathbb{R}_+^{s\times r}$ such that $\mf{u}_{ij} = 1$ (for $i=1,\ldots,s$ and $j=1,\ldots,r$), if $a_j$ is associated with the class $i$, and $\mf{u}_{ij} = 0$ otherwise. Then, the optimization problem of CNMF is defined as:
\begin{equation}\label{probcnmf}
	\min_{\mf{B}\in\mathbb{R}_+^{p\times k}, \mf{Z}\in\mathbb{R}_+^{k\times (q-r+s)}} \Vert \mf{A}-\mf{BZM}\Vert_F^2,
\end{equation}
where $k\ll\min\{p,q-r+s\}$. Compared to the primary problem of NMF, the major property of CNMF is that the coefficient representation matrix $\mf{C}$ is expressed as a linear combination form $\mf{C}=\mf{ZM}$.
However, when two given data samples share the same label, their corresponding representation vectors can be merged into a single point, which may be viewed as a drawback of CNMF. To address this issue, Babaee et al. \cite{babaee2016discriminative} introduced the DNMF method, which abandons the assumption of $\mf{C}=\mf{ZM}$ and instead learns the constraint matrix $\mf{M}$ as a free variable. With this description, the optimization problem of DNMF is defined as:
\begin{equation}\label{probcnmfbaba}
	\min_{\mf{B}\in\mathbb{R}_+^{p\times k}, \mf{C}\in\mathbb{R}_+^{k\times q},\mf{M}\in\mathbb{R}^{s\times k}} \Vert \mf{A}-\mf{B}\mf{C}\Vert_F^2+\alpha\Vert \mf{U}-\mf{M}\mf{C}_L\Vert_F^2,
\end{equation}
where $k\ll\min\{p,q\}$ and $\alpha$ is the regularization parameter. Moreover, $\mf{C}_L \in\mathbb{R}_+^{k\times r}$ represents a specific portion of the coefficient representation matrix $\mf{C}$, corresponding to the $r$ data samples having label, and the matrix $\mf{U}\in\mathbb{R}_+^{s\times r}$ was defined in \eqref{defM}. The key merit of DNMF is its ability to ensure that the vectors of data samples belonging to the same class are represented in a common axis during the transformation of high-dimensional data into a lower-dimensional space. Nonetheless, both CNMF and DNMF fail to take into account the fundamental geometric connections between data points, and as a consequence, they do not accurately capture crucial data patterns. This limitation can significantly impair the effectiveness of dimensionality reduction.
To counter this issue, Lu et al. \cite{lu2023robust} brought the idea of dual graph regularization into the framework of DNMF and proposed the RDGDNMF method. Moreover, RDGDNMF adopts its loss function based on the $L_{2,1/2}$ norm to mitigate the influence of outliers.
Jia et al. \cite{jia2019semi} introduced another potent discriminative NMF technique described as Semi-NMF-DSR. This approach leverages available label information within the data and discerns the impacts of samples originating from the same class versus different classes. Consequently, this method can generate distinctive low-dimensional representations of data samples.
Alongside these methods, other cases of discriminative NMF can be highlighted, such as
RSNMF \cite{7902136},
DSSNMF \cite{xing2021discriminative},
DCNMF \cite{PENG2022571}, and
S$^4$NMF \cite{CHAVOSHINEJAD2023109282}.

$\bullet$\,\,\textbf{Regularized NMF.}
To enhance the interpretability of factor matrices obtained from NMF, two prominent regularization techniques such as sparsity and graph regularizations can be incorporated into NMF.
\paragraph{NMF with Sparse Regularization}\label{Sparsesec}
The property of sparsity in NMF refers to the idea that the matrix factors generated by the NMF process should contain a large number of zero values. The aspect of sparsity holds significance in NMF for various reasons, including interpretability, dimensionality reduction, and computational efficiency.
In order to introduce a sparse NMF model, the NMF problem can be augmented with some sparsity constraints, which can be expressed as a function on the basis matrix $\mf{B}$ and/or a function on the coefficient representation matrix $\mf{C}$. The general form of a sparse NMF model can be written mathematically as:
\begin{equation}\label{spsnmf}
	\min_{\mf{B}\in\mathbb{R}_+^{p\times k}, \mf{C}\in\mathbb{R}_+^{k\times q}} \Vert \mf{A}-\mf{BC}\Vert_F^2+\alpha \mathcal{S}_{\text{reg}}(\mf{B})+\beta \mathcal{S}_{\text{reg}}(\mf{C}),
\end{equation}
where
$\mathcal{S}_{\text{reg}}(\mf{B})$ is a sparsity regularization function in terms of $\mf{B}$,
$\mathcal{S}_{\text{reg}}(\mf{C})$ is a sparsity regularization function in terms of $\mf{C}$,
and $\alpha,\beta>0$ are the sparsity regularization parameters. Table \ref{tablenorms} provides a summary of various common sparsity approaches for defining the sparsity regularization, along with their respective advantages and disadvantages.

 \begin{table}[!htbp]
	\centering
	\caption{A summary of the advantages and disadvantages of some well-known matrix norms used in the NMF problem.}
	\scalebox{1}{\begin{tabular}{|c||m{4cm}|m{4cm}|}
			\hline
			\textbf{Norm} &	\multicolumn{1}{>{\centering}m{4cm}|}{\textbf{Advantages}} 	& \multicolumn{1}{>{\centering}m{4cm}|}{\textbf{Disadvantages}} \\
			\hline
			Frobenius	
			&
			\begin{itemize}
				\itemsep0em
				\item Easy to compute	
				\item Has a geometric interpretation as the Euclidean distance
				\item Convex
			\end{itemize}
			&
			\begin{itemize}
				\itemsep0em
				\item
				Less sparse than $L_1$ and $L_{2,1}$
				\item
				Less robust to outliers than $L_1$
			\end{itemize} \\
			\hline
			$L_{0}$	
			&
			\begin{itemize}
				\itemsep0em
				\item Exact sparsity control
			\end{itemize}
			&
			\begin{itemize}
				\itemsep0em
				\item
				NP-hard to optimize
				\item
				Not convex
				\item
				Computationally expensive
				\item
				Not a valid norm
			\end{itemize} \\
			\hline
			$L_1$	
			&
			\begin{itemize}
				\itemsep0em
				\item Promotes sparsity
				\item Robust to outliers
				\item Convex
			\end{itemize}
			&
			\begin{itemize}
				\itemsep0em
				\item
				Non-differentiable at zero
				\item
				Difficult to optimize
			\end{itemize} \\
			\hline
			$L_{p}$ ($1<p$)	
			&
			\begin{itemize}
				\itemsep0em
				\item Generalizes the $L_1$-norm
				\item Smooth and differentiable
				\item Convex
			\end{itemize}
			&
			\begin{itemize}
				\itemsep0em
				\item
				Less sparse than $L_1$
			\end{itemize} \\
			\hline
			$L_{p}$ ($0<p<1$)	
			&
			\begin{itemize}
				\itemsep0em
				\item More sparse than $L_1$
			\end{itemize}
			&
			\begin{itemize}
				\itemsep0em
				\item
				Difficult to optimize
				\item
				Not convex
				\item
				Computationally expensive
				\item
				Not a valid norm
			\end{itemize} \\
			\hline
			$L_{2,0}$	
			&
			\begin{itemize}
				\itemsep0em
				\item Promotes row-wise sparsity
			\end{itemize}
			&
			\begin{itemize}
				\itemsep0em
				\item
				NP-hard to optimize
				\item
				Not convex
				\item
				Computationally expensive
				\item
				Not a valid norm
			\end{itemize} \\
			\hline
			$L_{2,1}$	
			&
			\begin{itemize}
				\itemsep0em
				\item Promotes row-wise sparsity
				\item More robust to outliers than $L_1$
				\item Convex
			\end{itemize}
			&
			\begin{itemize}
				\itemsep0em
				\item
				Non-differentiable at zero
				\item
				Difficult to optimize
				\item
				More computationally expensive than $L_1$
			\end{itemize} \\
			\hline
			$L_{2,p}$	 ($0<p<1$)	
			&
			\begin{itemize}
				\itemsep0em
				\item Promotes row-wise sparsity
				\item More sparse than $L_{2,1}$
			\end{itemize}
			&
			\begin{itemize}
				\itemsep0em
				\item
				Difficult to optimize
				\item
				Not convex
				\item
				Computationally expensive
				\item
				Not a valid norm
			\end{itemize} \\
			\hline
			\end{tabular}}
	\label{tablenorms}
\end{table}

 Since the inception of NMF, there has been a particular emphasis on exploring various sparsity-driven variants of NMF. The initial substantial attempts to establish the sparse NMF model were attributed to Hoyer \cite{hoyer2004non}, and it was demonstrated that integrating the concept of sparseness enhances the parts-based representation within NMF. The initial sparse NMF model introduced by Hoyer relied on the utilization of the $L_1$ norm. Following that, the exploration of diverse sparse NMF models, primarily employing the Frobenius, $L_p$ and $L_{2,p}$ norms, commenced. For example, Kim and Park \cite{kim2007sparse} discussed that enforcing a strict sparsity criterion on the basis matrix might lead to the omission of valuable information during the gene selection process. To tackle this issue, they developed a sparse NMF model that employs the Frobenius norm to impose the sparsity on the basis matrix. The use of the $L_1$ norm is evident in research works such as NeNMF \cite{2012-6166359}, GSNMTF \cite{9737322L1}, and MM-SNMF-$L_1$ \cite{major10103209}. Besides the $L_1$ norm, there has been significant interest in the two other pseudo norms in recent years, namely, $L_0$ and $L_{1/2}$. Instances of applying these norms include: NMF$L_0$ \cite{peharz2012sparse}, SGNMF \cite{chen2023graph}, and HGNMF-FS \cite{hyper10124031}.
Several successful instances of sparse NMF models utilizing the $L_{2,1}$ and $L_{2,0}$ norms are presented in
RCLR \cite{zhao2021robust},
DSMF/ISMF \cite{jin2022sparse},
and
NMF-{$L_{2,0}$}/ONMF-{$L_{2,0}$} \cite{L209893402}.
Finally, apart from the sparsity variations discussed earlier in this section, alternative techniques for creating sparse NMF models have also been explored. These techniques include the inner-product regularization, exemplified by RMFFS \cite{qi2018unsupervised} and DR-FS-MFMR \cite{saberi2022dual}, as well as the log-based sparsity regularization, illustrated by LS-NMF/RLS-NMF \cite{peng2022logbased}.

 \paragraph{NMF with Graph Regularization}
NMF with graph regularization involves transforming the data matrix into a graph, where each node represents a data sample and edges connecting nodes indicate their similarity. The information obtained from this constructed graph is then integrated into the NMF objective function as a regularization term, aiming to uncover the underlying geometric structure within the data space.
The following list outlines some of the most widely used graph regularization techniques applied in NMF.

 \noindent\textbf{Graph Laplacian:} A popular technique for graph regularization is the graph Laplacian regularization that employs the graph Laplacian matrix to represent the interconnectivity between nodes within a graph.
One of the earliest and most successful NMF methods that use the graph Laplacian regularization is the GNMF method \cite{cai2010graph}. It works under the assumption that when two data samples are in proximity to each other in the data space, their corresponding reduced dimension vectors should also be near each other in the low-dimensional space. The problem of GNMF is established as:
	\begin{equation}\label{pgnmf}
		\min_{\mf{B}\in\mathbb{R}_+^{p\times k}, \mf{C}\in\mathbb{R}_+^{k\times q}} \Vert \mf{A}-\mf{BC}\Vert_F^2+\alpha\tr(\mf{C}\mf{L}\mf{C}^T),
	\end{equation}
	where $\mf{L}$ is the Laplacian matrix associated with the graph Laplacian constructed based on the data samples, and $\alpha>0$ is the regularization parameter.
In recent years, NMF methods have been witnessing a growing tendency toward a perspective grounded in the notion of duality between data samples and features. This viewpoint primarily involves the creation of a dual graph regularization framework that emphasizes the incorporation of geometric aspects of both data samples and features.
A noteworthy instance in this context is the DGNMF method \cite{shang2012graph}, which unifies the graph regularizations of both data samples and features within the loss function of NMF. That is to say
	\begin{equation}\label{pdgnmf}
		\min_{\mf{B}\in\mathbb{R}_+^{p\times k}, \mf{C}\in\mathbb{R}_+^{k\times q}} \Vert \mf{A}-\mf{BC}\Vert_F^2+\alpha\tr(\mf{C}\mf{L}\mf{C}^T)+\beta\tr(\mf{B}^T\widetilde{\mf{L}}\mf{B}),
	\end{equation}
where $\alpha,\beta>0$ represent the regularization parameters. Moreover, $\mathbf{L}$ and $\widetilde{\mathbf{L}}$ are the Laplacian matrices associated with the data and feature graphs, respectively.
Further discussion about some other applications of the graph Laplacian in NMF can be found in DNMTF \cite{shang2012graph}, ODGNMF \cite{tang2021orthogonal}, DGLCF \cite{9786852dual}, and MVDGCF \cite{mu2023dualgraph}.
	
\noindent\textbf{Adaptive Neighbors Graph:}
While graph Laplacian-based clustering methods deliver efficient results, they group the data samples using an unchanging data graph. This can in turn bring about the clustering outcomes being significantly influenced by the initial similarity matrix. Moreover, the graph Laplacian is computed by creating a graph using the $k$-nearest neighbors, and this approach might not effectively capture the local data connections, which potentially results in neighborhoods that are less than optimal \cite{FSASL}. To tackle these issues, rather than employing the graph Laplacian with a predetermined neighboring connection, an alternative concept called Adaptive Neighbors Graph was introduced by Nie et al. \cite{Nieadaptive}. This concept revolves around generating a probabilistic neighborhood graph matrix, which the model learns to adapt dynamically. In a formal manner, the process of constructing a regularization term using the adaptive neighbors graph involves assuming that every data sample $\mf{a}_j$, for $j = 1, 2, \ldots, q$, is considered a neighbor of $\mf{a}_i$, with a corresponding probability denoted as $\mf{z}_{ij}$. Let $\mathbf{Z}=[\mf{z}_{ij}]\in\mathbb{R}_+^{q\times q}$ be the probabilistic neighborhood graph matrix. The subsequent problem is introduced to determine the values of $\mf{z}_{ij}$ where $i, j = 1, \ldots, q$:
	\begin{align}\label{Adaptive N}
		&\min_{\mf{Z}}\sum_{i,j=1}^q\left(\Vert\mf{a}_i-\mf{a}_j\Vert_2^2\mf{z}_{ij}+\alpha\mf{z}_{ij}^2\right)\\
		&\mathrm{subject\,\,to}\quad \mf{Z}\in\mathbb{R}_+^{q\times q}, \mf{Z}\mf{1}_q=\mf{1}_q,\nonumber
	\end{align}
	in which $\alpha>0$ denotes the regularization parameter.
Lately, the development of NMF models that utilize adaptive neighbor graphs has gained significant traction by researchers. Notably, Wang et al. \cite{9134812} introduced the RBSMF method, which involves defining a new robust loss function denoted as $Hx(x)$ and learning an adaptive graph through the utilization of a bi-stochastic matrix derived from the original data.
Drawing inspiration from the strengths of the $Hx(x)$ function, Tang and Feng \cite{tang2022robust} proposed the RLNMFAG approach, where they defined an effective and resilient variant of adaptive local graphs to mitigate the impact of large noises in data and extreme outliers.
Li et al. \cite{li2023semi} introduced the ABNMTF method, which incorporates the solution to the adaptive neighbor graph problem with a block diagonal matrix to construct the graph similarity matrix. This approach enables efficient learning of the similarity matrix for the purpose of clustering non-linearly separable data. In addition to these examples, other methods employing the concept of adaptive neighbor graphs are worth pointing out, including
NMFAN \cite{huang2020regularized},
NMF-LCAG \cite{yi2020nonnegative},
and
AGRDSNMF \cite{shu2022adaptive}.	

 \noindent\textbf{Hypergraph Laplacian:}
A hypergraph expands upon the idea of a simple graph by employing hyperedges, which are an extension of edges, to link more than just two nodes \cite{zhou2006learning}.
To define a matrix representation for a hypergraph, also known as the hypergraph Laplacian matrix, let $\mathcal{N}$ and $\mathcal{E}$ be the set of nodes and the set of hyperedges, respectively. Moreover, assume that $\mf{W}_\mathcal{E}$ is the weight matrix associated with the set of hyperedges, and $\mf{M}$ is the incidence matrix of the hypergraph. With this description, the unnormalized Laplacian matrix of a hypergraph, denoted as $\mf{L}_\mathcal{H}$, is defined as:
	\begin{align}\label{Adaptive N}
		\mf{L}_\mathcal{H} = \mf{D}_\mathcal{N} - \mf{M}\mf{W}_\mathcal{E}\mf{D}_\mathcal{E}^{-1}\mf{M}^T,
	\end{align}
	where $\mf{D}_\mathcal{N}$ and $\mf{D}_\mathcal{E}$ are denoted as the degree matrices associated with the nodes degree and the hyperedges degree, respectively.
	
Hypergraphs can offer significant advantages in capturing complex, high-order relationships among data points. Consequently, there has been a growing trend in NMF research that focuses on developing models utilizing hypergraphs. For instance, the HNMF \cite{ZENG2014209}, HGSNMF \cite{xu2023hypergraph} and CHNMF \cite{9130073hyper} methods can be referred to, all of which involve the utilization of a hypergraph, rather than a conventional graph, to investigate the high-order geometric characteristics of data.
In a recent work by Ye et al. \cite{heypeyyuan}, the RCHNMTF method was introduced. It represents a dual-regularized adaptation of NMTF and employs two hypergraphs to capture the underlying geometric properties within both the data and feature manifold.
Yin et al. \cite{YIN2023109274} implemented a semi-supervised variant of SNMF known as HSSNMF. This approach utilizes an innovative hypergraph-based technique for propagating pairwise constraints. Through this strategy, HSSNMF can successfully leverage the existing supervised information to build an
improved similarity structure.

\noindent$\bullet$\,\,\textbf{Generalized NMF.} This category explores three extensions of traditional NMF, namely Semi-NMF, Nonnegative Tensor Factorization (NTF), and Deep Nonnegative Matrix Factorization (DNMF). These variations expand the capabilities of NMF to encompass a wider range of data types and structures to address the limitations of traditional NMF.

 \paragraph{Semi-NMF}\label{convsec}
When real-valued entries are present in the data matrix $\mf{A}$, a type of matrix factorization known as Semi-NMF can be defined as an extension of NMF. It is worth highlighting that the reason behind developing Semi-NMF is rooted in its clustering point of view, which arises from the relationship between $k$-means and Semi-NMF \cite{ding2008convex}. In this context, the basis vectors are regarded as the cluster centroids, with no restrictions imposed on the values of the basis matrix $\mf{B}$. Furthermore, the entries in the coefficient matrix $\mf{C}=[\mf{c}_{ij}]$ are interpreted as the cluster indicators, all of which possess nonnegative values.

 To enhance the interpretability of the basis vectors (cluster centroids) obtained through Semi-NMF, Ding et al. \cite{ding2008convex} suggested the Convex-NMF method as a type of Semi-NMF that assumes that the basis vectors $\{\mf{b}_1,\mf{b}_2,\ldots,\mf{b}_k\}$ are constructed within the conic hull of the data samples $\{\mf{a}_1,\mf{a}_2,\ldots,\mf{a}_q\}$, which implies that each basis vector $\mathbf{b}_j$ can be specified as a non-negative weighted combination (conic combination) of data samples. By definition, each basis vector $\mathbf{b}_j$ in Convex-NMF has the structure
$\mf{b}_j\in\mathrm{coni}\left(\mf{a}_{1}, \mf{a}_{2},\ldots, \mf{a}_{q}\right)$, that is to say
\begin{equation}\label{convexrel1}
	\mf{b}_j = \mf{g}_{1j}\mf{a}_1+\mf{g}_{2j}\mf{a}_2+\cdots+\mf{g}_{qj}\mf{a}_q,\quad j =1,2,\ldots,k,
\end{equation}
where each $\mf{g}_{ij}$, for $i=1,\ldots,q$, is a nonnegative number. Now, if the matrix $\mf{G}$ is defined as $\mf{G}=[\mf{g}_{ij}]$, for $i=1,\ldots,q$ and $j =1,\ldots,k$,
	it can be easily verified from \eqref{convexrel1} that
	$[\mf{b}_1,\mf{b}_2,\ldots,\mf{b}_k] = [\mf{a}_1,\mf{a}_2,\ldots,\mf{a}_q]\mf{G}$,
	which means
	\begin{equation}\label{convexrel3}
		\mf{B} = \mf{A} \mf{G}\quad\mathrm{subject\,\,to}\,\,\mf{G}\in\mathbb{R}_+^{q\times k}.
	\end{equation}
It should be mentioned that both Semi-NMF and Convex NMF can be regarded as relaxed versions of $k$-means clustering \cite{ding2008convex}.

 Lastly, in cases where the data matrix $\mf{A}$ contains only non-negative values, Xu and Gong \cite{xu2004document} put forward the idea of Concept Factorization (CF) as an extension of NMF that exhibits enhanced performance, particularly when applied to datasets that display highly non-linear distributions \cite{cf5567104}. Furthermore, akin to Convex-NMF, CF represents each basis vector as a conic combination of the data samples as shown in \eqref{convexrel1}. 
 
 In brief, the framework of Semi-NMF, Convex-NMF, and CF can be summed up as:
	\begin{equation}
		\small
		\begin{array}{rl}
			\mathrm{Semi-NMF}: & \min_{\mf{B}, \mf{C}} \Vert \mf{A}-\mf{BC}\Vert_F^2,\\
			&\mathrm{subject\,\,to}\,\,\mf{A}\in\mathbb{R}_\pm^{p\times q}, \mf{B}\in\mathbb{R}_\pm^{p\times k}, \mf{C}\in\mathbb{R}_+^{k\times q},\\[1ex]
			\mathrm{Convex-NMF}: & \min_{\mf{G}, \mf{C}} \Vert \mf{A}-\mf{AGC}\Vert_F^2,\\
			&\mathrm{subject\,\,to}\,\,\mf{A}\in\mathbb{R}_\pm^{p\times q}, \mf{G}\in\mathbb{R}_+^{q\times k}, \mf{C}\in\mathbb{R}_+^{k\times q},
\\[1ex]
			\mathrm{CF}: & \min_{\mf{G}, \mf{C}} \Vert \mf{A}-\mf{AGC}\Vert_F^2,\\
			&\mathrm{subject\,\,to}\,\,\mf{A}\in\mathbb{R}_+^{p\times q}, \mf{G}\in\mathbb{R}_+^{q\times k}, \mf{C}\in\mathbb{R}_+^{k\times q},
		\end{array}
	\end{equation}
where $k\ll\min\{p,q\}$.

 Previous studies have examined the practical applications and inherent capabilities of Semi-NMF, Convex-NMF, and CF in dimensionality reduction. As an illustration, Zhang et al. \cite{semi6819071} offered the RsNGE approach, which is a nonnegative graph embedding method that relies on a robust form of Semi-NMF. Peng et al. \cite{PENG2022106} introduced a novel Semi-NMF technique, dubbed TS-NMF, specifically designed for two-dimensional data. Taking into account the nonlinear structures within the data, TS-NMF has the capability to maintain the topological characteristics of two-dimensional data when transformed into vectors. Pei et al. \cite{Pei7748469} showcased the CFAN method, which incorporates both concepts of the graph Laplacian and adaptive neighbors graph into the CF framework.
Subsequently, inspired by CFAN, Yuan et al. \cite{yuan2020convex} employed the loss function of Convex-NMF, enhanced by the adaptive graph concept, to develop the CNAFS technique for unsupervised feature selection. The primary benefit of CNAFS lies in its simultaneous consideration of the correlation between the data and the neighborhood data structure.
Peng et al. \cite{PENG2022571} presented a semi-supervised variation of Convex-NMF, referred to as DCNMF. This approach integrates both pointwise and pairwise constraints derived from labeled samples to enhance the quality of the low-dimensional data representation. In conjunction with these methods, various instances of Semi-NMF, Convex-NMF, and CF can also be emphasized, including LCCF \cite{cf5567104}, SeH \cite{hashing8115178}, PGCNMF \cite{LI20171}, and DGSCF \cite{wang2022dual}.

\paragraph{NTF}\label{secntf}
In various real-world applications such as medical imaging \cite{fmri}, data commonly possess multi-dimensional representations, which means they exist as tensors with order greater than two (multi-dimensional arrays) rather than simple vectors.
In order to deal with higher-order tensor data, a primary challenge in the analysis of such data lies in their typical high dimensionality, which can result in significant issues, including substantial computational expenses \cite{MR-NTD}.
Therefore, it is highly advantageous to carry out dimension reduction prior to deal with tensor data, which helps in maintaining the essential information of tensor while utilizing a more condensed representation. A proficient approach for reducing dimensionality in nonnegative tensor data is the utilization of Nonnegative Tensor Factorization (NTF) \cite{ZhouCichocki}.
Two specific tensor factorizations applied for the NTF analysis are the following: the Canonical Polyadic (CP) and the Tucker decomposition \cite{baderkolda}.
Let us consider a nonnegative data tensor $\mathcal{A}\in\mathbb{R}_+^{J_1\times J_2\times \cdots \times J_q}$. In the following, an explanation is provided for each of these two types of tensor factorization.
	
	\noindent\textbf{NTF based on CP (NTF-CP).} The CP decomposition involves breaking down a tensor into the summation of rank-one tensors. Following this, the NTF-CP problem can be formulated as an optimization task to find the best approximation of $\mathcal{A}$, that is
 	\begin{align}\label{tnfprob}
		&\min_{ \mf{C}^{(1)},\ldots,\mf{C}^{(q)}}\Vert\mathcal{A} - \llbracket \mf{C}^{(1)},\ldots,\mf{C}^{(q)}\rrbracket\Vert_F^2,\\
		&\mathrm{subject\,\,to}\quad\mf{C}^{(i)}\in \mathbb{R}_+^{J_i \times k_0},\,\,\text{for}\,\, i=1,2,\ldots,q,\nonumber
	\end{align}
	in which $k_0$ is a positive specified rank, and each nonnegative component factor $\mf{C}^{(i)}$ includes the column vectors
	$\{\mf{c}_1^{(i)}, \mf{c}_2^{(i)},\ldots, \mf{c}_{k_0}^{(i)}\}$. It is important to mention that the term $\llbracket \mf{C}^{(1)},\ldots,\mf{C}^{(q)}\rrbracket$ indicates a low-rank tensor representation of the original nonnegative tensor $\mathcal{A}$ and is commonly referred to as a CP tensor described as:
	\begin{equation}\label{tnfprob2}
		\llbracket \mf{C}^{(1)},\ldots,\mf{C}^{(q)}\rrbracket :=\sum_{k=1}^{k_0} \mf{c}_k^{(1)} \medcirc \mf{c}_k^{(2)} \medcirc \cdots \medcirc \mf{c}_k^{(q)},
	\end{equation}
	where $\medcirc$ denotes the vector outer product.

In a recent development, Che et al. \cite{che10195864} proposed the SGCP approach, a variant of NTF-CP enriched with sparsity and graph manifold regularization, for the purpose of detecting misinformation on social networks.
Wang et al. \cite{wang2021sparse} suggested the sparse NCP method, which is developed by utilizing the $L_1$ norm and the proximal algorithm to reach a stationary point and address the rank deficiency issue that may arise due to sparsity.
Chen et al. \cite{CP9737358} introduced the CPUFS approach, an innovative unsupervised feature selection method grounded on the NTF-CP problem. CPUFS, in particular, retains the multi-dimensional data structure throughout the entire feature selection procedure. Furthermore, as far as our knowledge extends, CPUFS is the first work that presents an unsupervised feature selection framework derived from NTF-CP. For further illustrations of NTF-CP in different contexts, NTF-ADM \cite{cai2013nonnegative}, CNTF \cite{yin2019learning}, and HyperNTF \cite{YIN2022190} can be referred to.
	
	\noindent\textbf{NTF based on Tucker (NTF-Tucker).}
	The purpose of the NTF-Tucker approach is to break down the nonnegative data tensor $\mathcal{A}$ into a set of $q+1$ components: a nonnegative core tensor $\mathcal{C}\in\mathbb{R}_+^{I_1\times I_2\times \cdots \times I_q}$ and $q$ nonnegative factor matrices $\mf{B}^{(i)}\in \mathbb{R}_+^{J_i \times I_i}$ (for $i=1,2,\ldots,q$), corresponding to each mode.
Accordingly, the NTF-Tucker problem can be stated as follows:
 	\begin{align}\label{tnfprob3}
		&\min_{\mathcal{C},\mf{B}^{(1)},\ldots,\mf{B}^{(q)}}\Vert \mathcal{A}-\mathcal{C}\times_{1}\mf{B}^{(1)}\times_2\mf{B}^{(2)}\times_3\cdots\times_{q}\mf{B}^{(q)}\Vert_F^2,\\
		&\mathrm{subject\,\,to}\quad\mathcal{C}\in\mathbb{R}_+^{I_1\times I_2\times \cdots \times I_q},\mf{B}^{(i)}\in \mathbb{R}_+^{J_i \times I_i},\,\,\text{for}\,\, i=1,2,\ldots,q,\nonumber
	\end{align}
	in which $\times_i$ denotes the operator associated with the product along the $i$th mode.
	
Numerous dimensionality reduction techniques have been developed thus far, building upon the NTF-Tucker structure. For example,
Sun et al. \cite{sun2015heterogeneous} presented the HTD-Multinomial approach, which is a modified version of NTF-Tucker. In this variant, the orthogonality constraints are applied to the first $q-1$ modes of the Tucker decomposition, while the last mode is constrained to adhere to a multinomial manifold.
Subsequently, drawing inspiration from the idea utilized in HTD-Multinomial, the LRRHTD approach \cite{zhang2017lowrank} was presented. In LRRHTD, rather than employing the multinomial manifold constraint on the last mode of the Tucker decomposition, it is instead subject to regularization through a low-rank constraint based on the nuclear norm. 
Li et al. \cite{7460200tucker} offered the MR-NTD method, which is designed with a graph manifold regularization variant of the NTF-Tucker framework. The primary objective of MR-NTD is to uphold the geometric details within tensor data by utilizing a manifold regularization term for the core tensors established during the Tucker decomposition.
Several other instances of NTF-Tucker can be observed in
AGRNTD \cite{chen2023adaptive},
UGNTD \cite{9058984gene},
and
ARTD \cite{GONG202375}.
	
	\paragraph{DNMF}

 NMF, known as a shallow matrix factorization algorithm, is limited to linear combinations of features and does not inherently account for higher-order interactions or hierarchical relationships that may be present in complex datasets \cite{de2021survey}.
To address this limitation and extract multi-level features, researchers have explored more advanced versions of NMF, such as DNMF \cite{cichocki2006multilayer}. The core intention of DNMF is to integrate the interpretability offered by the classical NMF with the capability of extracting a series of hierarchical features facilitated by the use of multilayer architectures.

 Let $l$ indicate the number of layers of a DNMF model, and let $\mf{A}\in\mathbb{R}_+^{p\times q}$ be a data matrix with $p$ features and $q$ data samples. The subsequent parts of this section provide a description for four distinct factorization frameworks utilized to formulate the DNMF problem.
	
\noindent\textbf{DNMF through the factorization of the coefficient matrix $\mf{C}$ (DNMF-$\mf{C}$).}
DNMF-$\mf{C}$ establishes a multilayer factorization procedure for the initial data matrix $\mf{A}$, utilizing the coefficient matrix $\mf{C}$ in the following manner:
	\begin{align}
		\mf{A} \approx \mf{B}_1\mf{C}_1,\quad
		\mf{C}_1 \approx \mf{B}_2\mf{C}_2,\quad\ldots\quad
		\mf{C}_{l-1} \approx \mf{B}_{l}\mf{C}_l.\label{ldnmf0}
	\end{align}
	On account of this, the problem \eqref{ldnmf0} can be transformed into the following best approximation problem:
 	\begin{align}\label{ldnmf1}
		&\min_{\mf{B}_1,\ldots,\mf{B}_l,\mf{C}_l} \Vert \mf{A} - \mf{B}_1\ldots\mf{B}_l\mf{C}_l\Vert_F^2,\\
		&\mathrm{subject\,\,to}\quad\mf{B}_i\in\mathbb{R}_+^{k_{i-1}\times k_i},\,\,\mf{C}_l\in\mathbb{R}_+^{k_{l}\times q},\,\,\text{for}\,\,i=1,\ldots,l.\nonumber
	\end{align}
 	In the entire process of DNMF-$\mf{C}$, it is assumed that $k_l\leq k_{l-1}\leq\cdots\leq k_1\leq k_0=p$.
	
The DNMF-$\mf{C}$ framework serves as the foundation for the subsequent noteworthy deep NMF methods:
DANMF \cite{ye2018deep},
FSSDNMF \cite{chen2022link},
and
ODD-NMF \cite{LUONG20221088152022}.
	
\noindent\textbf{DNMF through the factorization of the basis matrix $\mf{B}$ (DNMF-$\mf{B}$).}
	The framework of DNMF-$\mf{B}$ follows a process similar to that of DNMF-$\mf{C}$, except that the hierarchical decomposition in DNMF-$\mf{B}$ takes place on the basis matrix $\mf{B}$ rather than the coefficient matrix $\mf{C}$. As a result, the subsequent procedure unfolds:
	\begin{align}
		\mf{A} \approx \mf{B}_1\mf{C}_1,\quad
		\mf{B}_1 \approx \mf{B}_2\mf{C}_2,\quad\ldots\quad
		\mf{B}_{l-1} \approx \mf{B}_{l}\mf{C}_l, \label{ldnmfB}
	\end{align}
	Accordingly, the optimization problem of DNMF-$\mf{B}$ can be established as:
	\begin{align}\label{ldnmf1}
		&\min_{\mf{B}_l,\mf{C}_l,\ldots,\mf{C}_1} \Vert \mf{A} - \mf{B}_l\mf{C}_l\ldots\mf{C}_1\Vert_F^2,\\
		&\mathrm{subject\,\,to}\quad\mf{B}_l\in\mathbb{R}_+^{p\times{k'_{l}}},\,\,\mf{C}_i\in\mathbb{R}_+^{k'_{i}\times k'_{i-1}},\,\,\text{for}\,\,i=1,\ldots,l,\nonumber
	\end{align}
	in which it is assumed that $k'_l\leq k'_{l-1}\leq\cdots\leq k'_1\leq k'_0=q$.
	
Based on the DNMF-$\mf{B}$ framework, the following examples of deep NMF can be mentioned:
CMLNMF \cite{chen2017multilayer}, DNBMF, RDNBMF, and RDNNBMF \cite{8943941pami2021}.

 \noindent\textbf{Deep Alternating NMF (DA-NMF).}
	The major limitation of the DNMF-$\mf{B}$ and DNMF-$\mf{C}$ methods is that their optimization problems do not explicitly incorporate the hierarchical representation of both the basis and coefficient matrices. To overcome this limitation and design a DNMF framework that employs hierarchical representation for both matrices, the DA-NMF method proposed by Sun et al. \cite{sun2022deep} breaks down the basis and coefficient matrices alternately across the layers. The decomposition process of DA-NMF, involving $l$ layers, can be briefly described as follows:
	\begin{align}
		&\mf{A} \approx \mf{B}_1\mf{C}_1,\nonumber\\
		&\mf{C}_1 \approx \mf{B}_2\mf{C}_2,\qquad\mf{B}_2 \approx \mf{B}_3\mf{C}_3,\nonumber\\
		&\vdots\label{ldnmfalter}\\
		&\mf{B}^{l-2}\approx \mf{B}^{l-1}\mf{C}^{l-1},\qquad\mf{C}^{l-1} \approx \mf{B}^{l}\mf{C}^{l}.\nonumber
	\end{align}
	Thus, the optimization problem for DA-NMF with $l$ layers can be formulated as follows:
	\begin{align}\label{ldnmfDA-NMF}
		&\min_{\mf{B}_i,\mf{C}_j} \Vert \mf{A} - \mf{B}_1 \mf{B}_3\cdots\mf{B}_{2l-1}\mf{C}_{2l-1}\cdots\mf{C}_{4}\mf{C}_{2}\Vert_F^2,\\
		&\mathrm{subject\,\,to}\quad\mf{B}_i,\mf{C}_j\geq0,\nonumber
	\end{align}
	where $i=1,3,\ldots,2l-1$ and $j=2,4,\ldots,2l-1$. Moreover,
	$\mf{B}_1 \in \mathbb{R}_+^{p\times {k_1}}$,
	$\mf{B}_i \in \mathbb{R}_+^{{k_{i-2}}\times {k_i}}$,
	$\mf{C}_2 \in \mathbb{R}_+^{{k_1}\times q}$,
	and $\mf{C}_j \in \mathbb{R}_+^{k_{j}\times {k_{j-2}}}$.

\noindent\textbf{Deep Semi-NMF (D-Semi-NMF).}
	Taking inspiration from the Semi-NMF framework, Trigeorgis et al. \cite{trigeorgis2016deep} introduced the D-Semi-NMF method which stands as a multi-layer version of Semi-NMF and disregards the requirement of nonnegativity for the data matrix $\mf{A}$. Furthermore, the hierarchical decomposition within D-Semi-NMF occurs within the coefficient matrices produced during the multi-layer process. The multi-layer procedure of D-Semi-NMF, which includes $l$ layers, can be explained as follows:
	\begin{align}\label{ldnmf1deepsemiii}
		&\min_{\mf{B}_1^\pm,\ldots,\mf{B}_l^\pm,\mf{C}_l} \Vert \mf{A}^\pm - \mf{B}_1^\pm\mf{B}_2^\pm\cdots\mf{B}_l^\pm\mf{C}_l\Vert_F^2,\\
		&\mathrm{subject\,\,to}\quad\mf{B}_i^\pm\in\mathbb{R}^{k_{i-1}\times k_i},\,\,\mf{C}_l\in\mathbb{R}_+^{k_{l}\times q},\,\,\text{for}\,\,i=1,\ldots,l.\nonumber
	\end{align}
 The following examples of deep NMF, rooted in the D-Semi-NMF framework, are remarkable:	
SGDNMF \cite{8858038semi2020},
DMRMF\_MVC \cite{LIU2023109806},
MCDS, and DSNMF \cite{WANG2023101884}.
	
\noindent$\bullet$\,\,\textbf{Robust NMF.} The goal of this section is to review techniques that empower NMF to effectively handle challenges such as dealing with potential noise that might emerge while executing the NMF procedure. In practical applications, the presence of noise or outliers in the data can considerably influence the decomposition process of NMF. Therefore, removing or assigning lower significance to these elements can lead to a more robust factorization outcome. Robust NMF endeavors to tackle this issue by incorporating mechanisms that strengthen the factorization process against such irregularities. There are various approaches to improve the robustness of the NMF algorithm, including: Sparsity Regularizations, and Robust Loss Functions.

 \noindent \textbf{Sparsity Regularizations:} In the context of NMF, regularization pertains to the concept of introducing a penalty term into the objective function, which influences the process of factorization. The regularization term basically promotes specific characteristics within the factorization matrices, such as inducing sparsity or ensuring smoothness. Incorporating a sparsity regularization term into the NMF objective function encourages the basis or coefficient matrices to exhibit sparsity. This prompts the elements within these matrices to change gradually rather than suddenly. Consequently, NMF models utilizing sparsity-based regularization are characterized through reduced sensitivity and increased resilience to minor variations in the input data. In Table \ref{tablenorms}, some of the most effective strategies for defining a sparse NMF model were discussed.
%
%

 \noindent \textbf{Robust Loss Functions:} In this case, instead of using the Frobenius norm-based loss functions, there are other robust loss functions that are less affected by noise and outliers and have been used in the loss function of NMF. Several efficient tools employed to formulate a more resilient loss function for NMF include: (1) the $L_{2,1}$ norm (e.g., $L_{2,1}$-NMF \cite{kong2011robust}), (2) the $\beta$-divergence functions (e.g., MM-SNMF-$L_1$ \cite{marmin2023majorization}), (3) the Kullback-Leibler divergence functions (e.g., FNTNMF-KLD \cite{hu2022feature}), (4) the maximum correntropy criterion (e.g., CHNMF \cite{9130073hyper}), and (5) the $Hx(x)$ function (e.g., Hx-NMF \cite{9134812}). 

	\subsubsection{NMF in Feature Selection}\label{FSNMF}
As a technique for feature representation learning, NMF can provide an efficient solution for feature selection, thanks to its ability to represent data by decomposing it into the basis and coefficient components. The main objective of employing NMF for feature selection is to detect a smaller set of features obtained from the original dataset, with the objective of keeping the most informative and pertinent features while disregarding those that are less significant. A workable strategy to define a framework for feature selection using NMF involves the construction of a basis matrix $\mathbf{B} \in \mathbb{R}_+^{p\times k}$ and a coefficient representation matrix $\mathbf{C} \in \mathbb{R}_+^{k \times q}$, which serve as a compressed representation of the features and data samples,  respectively. The purpose of this construction is to
	\begin{equation}\label{ppnmf}
		\min_{\mf{B}\in\mathbb{R}_+^{p\times k}, \mf{C}\in\mathbb{R}_+^{k\times q}} \Vert \mf{A}-\mf{BC}\Vert_F^2,
	\end{equation}
where the dimensionality reduction parameter $k$ is significantly smaller than $p,q$. 

Problem \eqref{ppnmf} has been employed as a conventional foundation for the introduction of feature selection methods based on NMF. As an illustration, Meng et al. \cite{meng2018feature} proposed the DSNMF method, wherein Problem \eqref{ppnmf} is integrated with the dual-graph structure linked to the data and feature graphs. DSNMF also applies the $L_{2,1}$ norm to enforce the sparsity on the basis matrix, signifying the significance of the chosen features. The problem for DSNMF is presented as follows:
\begin{equation}\label{DSNMFprob}
\min_{\mf{B}\in\mathbb{R}_+^{p\times k}, \mf{C}\in\mathbb{R}_+^{k\times q}} \Vert \mf{A}-\mf{BC}\Vert_F^2+\alpha\tr(\mf{C}\mf{L}\mf{C}^T)+\beta\tr(\mf{B}^T\widetilde{\mf{L}}\mf{B})+\gamma\Vert\mf{B}\Vert_{2,1},
\end{equation}
where $\alpha,\beta,\gamma>0$ represent the regularization parameters. Another illustration includes the NMF-LRSR approach \cite{shang2020double}, which is constructed by applying the concept of low-rank sparse data representation. The optimization problem for NMF-LRSR is:
\begin{align}\label{DSNMFprob}
\min_{\mf{B}\in\mathbb{R}_+^{p\times k}, \mf{C}\in\mathbb{R}_+^{k\times q}} &\Vert \mf{A}-\mf{BC}\Vert_{2,1}+\alpha\Vert \mf{C}-\mf{C}\mf{V}\Vert_F^2+\beta\Vert \mf{I}_k-\mf{C}\mf{C}^T\Vert_F^2+\gamma\Vert\mf{B}\Vert_{2,1},
\end{align}
where $\mf{V}$ indicates the low-rank representation matrix for $\mf{A}$ with the purpose of directing the learning process of the coefficient matrix $\mf{C}$ to retain the global geometric information of data.

Another perspective recently adopted for introducing feature selection methods  is to extend the idea of Concept Factorization (CF), which was discussed in \eqref{cfmethod}, to the feature space. Given the significance of the basis matrix in NMF, the primary objective of applying CF to the feature space is to efficiently learn the basis vectors $\{\mathbf{b}_{1}, \mathbf{b}_{2}, \ldots, \mathbf{b}_{k}\}$ so that they can effectively capture the fundamental characteristics of the original data, thereby generating significant features. Let us consider a non-negative dataset represented by the feature matrix $\mathbf{X} = [\mathbf{f}_1, \mathbf{f}_2, \ldots, \mathbf{f}_p] \in \mathbb{R}_+^{q \times p}$, where each feature vector $\mathbf{f}_j$, for $j = 1, \ldots, p$, exists in a $q$-dimensional data space. The general framework for a CF-based feature selection method is defined as:
	\begin{equation}\label{pppnmfnew}
	\min_{\mf{G}\in\mathbb{R}_+^{p\times k}, \mf{C}\in\mathbb{R}_+^{k\times p}} \Vert \mf{X}-\mf{XGC}\Vert_F^2,
\end{equation}
where $k\leq p$. To be more specific, in Problem \eqref{pppnmfnew}, the assumption is made that each basis vector $\mathbf{b}_{l}$ can be expressed as a conic combination of feature vectors, implying that
	\begin{equation}\label{fc0}
		\mf{b}_{l}\in\mathrm{coni}\left(\mf{f}_{1}, \mf{f}_{2},\ldots, \mf{f}_{p}\right),\qquad\text{for}\quad l=1,\ldots,k.
	\end{equation}
	Hence, it can be deduced that each basis vector $\mf{b}_{l}$ has the following form:
	\begin{equation}\label{fc1}
		\mf{b}_{l}=\mf{g}_{1l}\mf{f}_{1}+\mf{g}_{2l}\mf{f}_{2}+\cdots+\mf{g}_{pl}\mf{f}_{p},
	\end{equation}
	where $\mf{g}_{jl}\geq 0$, for $j=1,\ldots,p$. This indicates that if the feature weight matrix $\mf{G}$ is defined as $\mf{G}=[\mf{g}_{jl}]\in\mathbb{R}_+^{p\times k}$, then it can be shown that
	\begin{align}
		\mf{B}=\mf{XG}.\label{fc4}
	\end{align}

The CF-based feature selection framework \eqref{pppnmfnew} has gained significant attention from researchers in recent years. This attention has led to the development of successful feature selection methods by combining the objective function of \eqref{pppnmfnew} with various types of regularizations or incorporating linear or non-linear optimization constraints. For instance, Wang et al. \cite{wang2015subspace} introduced the MFFS method, wherein Problem \eqref{pppnmfnew} is constrained by imposing an orthogonality condition on the feature weight matrix $\mf{G}$. In other words,
\begin{align}\label{wang2015subspaceprob}
	&\min_{\mf{G}\in\mathbb{R}_+^{p\times k}, \mf{C}\in\mathbb{R}_+^{k\times p}} \Vert \mf{X}-\mf{XGC}\Vert_F^2,\\
	&\mathrm{subject\,\,to}\quad\mf{G}^T\mf{G}=\mf{I}_k.\nonumber
\end{align}
A number of feature selection methods grounded in the CP-framework \eqref{pppnmfnew} can be summarized as:
\begin{itemize}
\item \textbf{SGFS \cite{shang2016subspace}:} 
\begin{align}
	&\min_{\mf{G}\in\mathbb{R}_+^{p\times k}, \mf{C}\in\mathbb{R}_+^{k\times p}} \alpha\Vert \mf{X}-\mf{XGC}\Vert_F^2+\tr(\mf{C}\widetilde{\mathbf{L}}\mf{C}^T)+\beta\Vert\mf{G}\Vert_{2,1},\\
	&\mathrm{subject\,\,to}\quad\mf{G}^T\mf{G}=\mf{I}_k.\nonumber
\end{align}

\item \textbf{RMFFS \cite{qi2018unsupervised}:} 
\begin{align}
	&\min_{\mf{G}\in\mathbb{R}_+^{p\times k}, \mf{C}\in\mathbb{R}_+^{k\times p}}\Vert \mf{X}-\mf{XGC}\Vert_F^2+\alpha\left(\Vert\mf{G}\mf{G}^T\Vert_1-\Vert\mf{G}\Vert_F^2\right).
\end{align}

\item \textbf{SPFS \cite{wang2020structured}:} 
\begin{align}\label{wang2015subspaceprob}
	&\min_{\mf{G}\in\mathbb{R}_+^{p\times k}, \mf{C}\in\mathbb{R}_+^{k\times p}} \Vert \mf{X}-\mf{XGC}\Vert_F^2+\alpha\tr(\mf{G}^T\mf{X}^T\mathbf{L}\mf{XG}),\\
	&\mathrm{subject\,\,to}\quad\mf{G}^T\mf{G}=\mf{I}_k.\nonumber
\end{align}
\end{itemize}

Finally, Table \ref{tablefeamethods} outlines the specifics of various feature selection methods that are based on either Problem \eqref{ppnmf} or Problem \eqref{pppnmfnew}. In this table, ``Graph-Based" and ``Dual Graph-Based" indicate the use of either the graph Laplacian or the dual graph Laplacian to define the feature selection approach. ``Sparsity" signifies the utilization of a sparsity constraint in the feature selection method, while ``Orthogonality Constraint" refers to the case where the objective function of the proposed feature selection technique is constrained by orthogonality.

	\begin{table*}[!htbp]
		\centering
		\caption{A summary of a number of NMF-based feature selection methods.}
		\scalebox{0.7}{\begin{tabular}{|l||c|c|c|c|c|c|}
				\hline
				\multicolumn{1}{|>{\centering}m{2cm}||}{\textbf{Method}} &	\multicolumn{1}{>{\centering}m{3cm}|}{\textbf{Standard NMF Problem} \eqref{ppnmf}} 	& \multicolumn{1}{>{\centering}m{2.5cm}|}{\textbf{Concept Factorization (CF) Problem} \eqref{pppnmfnew}} & \multicolumn{1}{>{\centering}m{2.2cm}|}{\textbf{Graph-Based}} & \multicolumn{1}{>{\centering}m{2.5cm}|}{\textbf{Dual Graph-Based}} & \textbf{Sparsity} & \multicolumn{1}{>{\centering}m{2.5cm}|}{\textbf{Orthogonality Constraint}}\\
				\hline
				AGNMF-FS \cite{wang2015feature} &
				\inc & 
				\notc & 
				\inc & 
				\notc & 
				\notc & 
				\notc 
				\\[1ex]\hline
				MFFS \cite{wang2015subspace} &
				\notc & 
				\inc & 
				\notc & 
				\notc & 
				\notc & 
				\inc 
				\\[1ex]\hline
				MPMR \cite{wang2015unsupervised} &
				\notc & 
				\inc & 
				\notc & 
				\notc & 
				\notc & 
				\inc 
				\\[1ex]\hline
				NMFBFS \cite{ji2015nmfbfs} &
				\inc & 
				\notc & 
				\notc & 
				\notc & 
				\inc & 
				\notc 
				\\[1ex]\hline
				JDSSL \cite{zhou2016discriminative} &
				\incc & 
				\inc & 
				\incc & 
				\notc & 
				\incc & 
				\incc 
				\\[1ex]\hline
				FSNMF \cite{liang2016feature} &
				\inc & 
				\notc & 
				\notc & 
				\notc & 
				\inc & 
				\notc 
				\\[1ex]\hline
				SGFS \cite{shang2016subspace} &
				\notc & 
				\inc & 
				\inc & 
				\notc & 
				\inc & 
				\inc 
				\\[1ex]\hline
				NSSRD \cite{shang2017non} &
				\incc & 
				\notc & 
				\incc & 
				\incc & 
				\incc & 
				\incc 
				\\[1ex]\hline
				DSNMF \cite{meng2018feature} &
				\incc & 
				\notc & 
				\incc & 
				\incc & 
				\incc & 
				\notc 
				\\[1ex]\hline
				OPMF \cite{yi2018ordinal} &
				\notc & 
				\incc & 
				\incc & 
				\notc & 
				\incc & 
				\notc 
				\\[1ex]\hline
				RMFFS \cite{qi2018unsupervised} &
				\notc & 
				\incc & 
				\notc & 
				\notc & 
				\incc & 
				\notc 
				\\[1ex]\hline
				SC-CNMF$_1$/SC-CNMF$_2$ \cite{cui2018subspace} 
                    &
				\incc & 
				\incc & 
				\incc & 
				\notc & 
				\incc & 
				\notc 
				\\[1ex]\hline
				RHNMF \cite{yu2019robust} &
				\incc & 
				\notc & 
				\incc & 
				\notc & 
				\notc & 
				\notc 
				\\[1ex]\hline
				SFS-BMF1/SFS-BMF2 \cite{saberi2020supervised} &
				\notc & 
				\incc & 
				\notc & 
				\notc & 
				\notc & 
				\incc 
				\\[1ex]\hline
				SFSMF \cite{zare2019supervised} &
				\notc & 
				\incc & 
				\notc & 
				\notc & 
				\notc & 
				\incc 
				\\[1ex]\hline
				NMF-LRSR \cite{shang2020double} &
				\incc & 
				\notc & 
				\notc & 
				\notc & 
				\incc & 
				\incc 
				\\[1ex]\hline
				SLARS \cite{shang2020subspace} &
				\notc & 
				\incc & 
				\incc & 
				\notc & 
				\incc & 
				\incc 
				\\[1ex]\hline
				SLSDR \cite{shang2020sparse} &
				\notc & 
				\incc & 
				\incc & 
				\incc & 
				\incc & 
				\incc 
				\\[1ex]\hline
				SPFS \cite{wang2020structured} &
				\notc & 
				\incc & 
				\incc & 
				\notc & 
				\notc & 
				\incc 
				\\[1ex]\hline
				B-MFFS/B-MPMR/B-SGFS \cite{mehrpooya2022high} &
				\notc & 
				\incc & 
				\incc & 
				\notc & 
				\incc & 
				\incc 
				\\[1ex]\hline
				DGSLFS \cite{sheng2021dual} &
				\notc & 
				\incc & 
				\incc & 
				\incc & 
				\incc & 
				\incc 
				\\[1ex]\hline
				CNAFS \cite{yuan2020convex} &
				\incc & 
				\incc & 
				\incc & 
				\incc & 
				\incc & 
				\incc 
				\\[1ex]\hline
				DR-FS-MFMR \cite{saberi2022dual} &
				\notc & 
				\incc & 
				\incc & 
				\notc & 
				\incc & 
				\notc 
				\\[1ex]\hline
				OCM-UFS \cite{luo2022orthogonally} &
				\incc & 
				\notc & 
				\incc & 
				\notc & 
				\incc & 
				\incc 
				\\[1ex]\hline
				RMFRASL \cite{lai2022rmfrasl} &
				\incc & 
				\incc & 
				\incc & 
				\notc & 
				\notc & 
				\incc 
				\\[1ex]\hline
				UFGOR \cite{jahani2023unsupervised} &
				\notc & 
				\incc & 
				\incc & 
				\notc & 
				\notc & 
				\incc 
				\\[1ex]\hline
				SLNMF \cite{zhou2023soft} &
				\notc & 
				\incc & 
				\notc & 
				\notc & 
				\incc & 
				\incc 
				\\[1ex]\hline
				VCSDFS \cite{KARAMI2023} &
				\incc & 
				\notc & 
				\notc & 
				\notc & 
				\incc & 
				\notc 
				\\[1ex]\hline
				OCLSP \cite{lin2021unsupervised} &
				\incc & 
				\notc & 
				\incc & 
				\notc & 
				\incc & 
				\incc 
				\\ \hline
		\end{tabular}}
		\label{tablefeamethods}
	\end{table*}

 \section{Future Direction for NMF in Dimensional Reduction}\label{sec4}
\subsection{Semi-Supervised NMF for Dimensionality Reduction}

 Semi-supervised NMF endeavors to achieve factorization process by leveraging both labeled and unlabeled data which can lead to more effective and accurate low-dimensional representations. It takes the advantages of matrix factorization and semi-supervised learning to effectively learn part-based representations and improve the capabilities of NMF in dimensionality reduction when dealing with limited labeled data and abundant unlabeled data. Future research on semi-Supervised NMF in dimensionality reduction could focus on developing new algorithms to integrate different types of supervision information into the NMF framework, designing deep semi-supervised NMF architectures, presenting semi-supervised NMF algorithms for feature selection, exploiting the intrinsic geometrical structure of data and utilizing a hypergraph for capturing high-order relationships among samples.

 \subsection{Hypergraph-Based NMF for Dimensionality Reduction}

 Hypergraphs, distinct from traditional graph structures, provide greater flexibility in modeling relationships among data points, enabling richer representations of interactions. Integrating hypergraphs with NMF enhances dimensionality reduction by capturing higher-order relationships, benefiting data understanding. Hypergraph-Based NMF uncovers intricate relationships and latent features, accommodating diverse interactions beyond pairwise connections. Future research in this field could refine hypergraph construction techniques, optimize algorithms, and explore applications across domains, advancing NMF in dimensionality reduction.

 \subsection{Sparse NMF with the $L_{2,0}$ norm}
While numerous sparse NMF techniques have been introduced using various matrix norms, the exploration of utilizing the $L_{2,0}$ norm for constructing a sparse NMF model remains relatively limited in terms of both theoretical and numerical investigations. In a recent study, Min et al. \cite{L209893402} showcased the NMF-$L_{2,0}$ method and its corresponding orthogonal variation, ONMF-$L_{2,0}$, wherein the $L_{2,0}$ norm can effectively promote sparsity. However, the incorporation of the $L_{2,0}$ norm in NMF and its diverse adaptations, coupled with the development of efficient algorithms for solving $L_{2,0}$ norm-based NMF models, emerges as an intriguing avenue of research.

 \subsection{Nonnegative Tensor Factorization in Feature Selection}
As highlighted in Section \ref{secntf}, the CPUFS approach \cite{CP9737358} marks the first attempt to introduce an unsupervised feature selection framework originating from Nonnegative Tensor Factorization (NTF) through the use of Canonical Polyadic (CP) decomposition. Hence, delving into the development of effective supervised, unsupervised, or semi-supervised feature selection techniques grounded in the NTF-CP decomposition and enriched with diverse forms of regularization can offer an innovative research direction for scholars to explore.

 \subsection{Deep NMF in Feature Selection}
Though the utility of NMF in feature selection is evident, to the best of our knowledge, the multi-layer version of NMF, known as Deep NMF (DNMF), has not yet found application in this context. Additionally, exploring DNMF-based feature selection techniques that incorporate both the local and global geometric information of data at each layer could serve as a significant area of research interest.

 \subsection{Adaptive Learning NMF for Dimensionality Reduction }

 Adaptive learning involves techniques where a model behavior and performance automatically adjust in response to incoming data, enabling continuous improvement and better adaptation to evolving patterns and conditions. In certain Nonnegative Matrix Factorization (NMF) variants, such as SNMF and GNMF, reliance on static matrices, like similarity matrices or predefined graphs, initially posed limitations. These matrices remained fixed throughout the learning process. However, recent advancements are steering towards a more flexible approach, eliminating the need for fixed inputs. The focus is on enabling NMF to autonomously learn and adapt to various inputs during the learning process. Adaptive SNMF and Adaptive Graph Learning within graph-regularized NMF are emerging as significant trends. These approaches promise to enhance NMF effectiveness by dynamically adjusting to data characteristics and evolving patterns, ultimately leading to more optimized learning outcomes.

 \subsection{Constructing Objective Function based on Non-Frobenius norm}\label{sec4}
Exploring new loss functions that are tailored to specific tasks or data types can open up new possibilities for NMF applications. For instance, incorporating domain-specific loss functions for tasks like image segmentation or text clustering can improve the performance of NMF in these contexts.


 \section{Conclusion}\label{sec5}

This paper presents an extensive overview of NMF
in the realm of dimensionality reduction, distinguishing
between feature extraction and selection. It conducts an in-depth analysis of NMF in feature extraction, categorizing it into four principal groups: Variety, Regularization, Generalized, and Robust. The exploration of NMF for feature selection spans six distinct perspectives. Lastly, we outline six potential future avenues for advancing NMF in dimensionality reduction techniques.

\bibliographystyle{ACM-Reference-Format}
\bibliography{ManuscriptNMF}


\begin{thebibliography}{192}


\ifx \showCODEN    \undefined \def \showCODEN     #1{\unskip}     \fi
\ifx \showDOI      \undefined \def \showDOI       #1{#1}\fi
\ifx \showISBNx    \undefined \def \showISBNx     #1{\unskip}     \fi
\ifx \showISBNxiii \undefined \def \showISBNxiii  #1{\unskip}     \fi
\ifx \showISSN     \undefined \def \showISSN      #1{\unskip}     \fi
\ifx \showLCCN     \undefined \def \showLCCN      #1{\unskip}     \fi
\ifx \shownote     \undefined \def \shownote      #1{#1}          \fi
\ifx \showarticletitle \undefined \def \showarticletitle #1{#1}   \fi
\ifx \showURL      \undefined \def \showURL       {\relax}        \fi
\providecommand\bibfield[2]{#2}
\providecommand\bibinfo[2]{#2}
\providecommand\natexlab[1]{#1}
\providecommand\showeprint[2][]{arXiv:#2}

\bibitem[Alhassan and Zainon(2021)]%
        {alhassan2021review}
\bibfield{author}{\bibinfo{person}{Afnan~M Alhassan} {and} \bibinfo{person}{Wan Mohd Nazmee~Wan Zainon}.} \bibinfo{year}{2021}\natexlab{}.
\newblock \showarticletitle{Review of feature selection, dimensionality reduction and classification for chronic disease diagnosis}.
\newblock \bibinfo{journal}{\emph{IEEE Access}}  \bibinfo{volume}{9} (\bibinfo{year}{2021}), \bibinfo{pages}{87310--87317}.
\newblock


\bibitem[Alpaydin(2014)]%
        {alpaydin2014introduction}
\bibfield{author}{\bibinfo{person}{Ethem Alpaydin}.} \bibinfo{year}{2014}\natexlab{}.
\newblock \bibinfo{title}{Introduction to machine learning, 3rd editio. ed}.
\newblock
\newblock


\bibitem[Anowar et~al\mbox{.}(2021)]%
        {anowar2021conceptual}
\bibfield{author}{\bibinfo{person}{Farzana Anowar}, \bibinfo{person}{Samira Sadaoui}, {and} \bibinfo{person}{Bassant Selim}.} \bibinfo{year}{2021}\natexlab{}.
\newblock \showarticletitle{Conceptual and empirical comparison of dimensionality reduction algorithms (pca, kpca, lda, mds, svd, lle, isomap, le, ica, t-sne)}.
\newblock \bibinfo{journal}{\emph{Computer Science Review}}  \bibinfo{volume}{40} (\bibinfo{year}{2021}), \bibinfo{pages}{100378}.
\newblock


\bibitem[Asteris et~al\mbox{.}(2015)]%
        {asteris2015orthogonal}
\bibfield{author}{\bibinfo{person}{Michail~Vlachos Asteris}, \bibinfo{person}{Dimitris Papailiopoulos}, {and} \bibinfo{person}{Alexandros~G. Dimakis}.} \bibinfo{year}{2015}\natexlab{}.
\newblock \showarticletitle{Orthogonal {NMF} through Subspace Exploration}. In \bibinfo{booktitle}{\emph{Advances in Neural Information Processing Systems}}, Vol.~\bibinfo{volume}{28}.
\newblock


\bibitem[Ayesha et~al\mbox{.}(2020)]%
        {ayesha2020overview}
\bibfield{author}{\bibinfo{person}{Shaeela Ayesha}, \bibinfo{person}{Muhammad~Kashif Hanif}, {and} \bibinfo{person}{Ramzan Talib}.} \bibinfo{year}{2020}\natexlab{}.
\newblock \showarticletitle{Overview and comparative study of dimensionality reduction techniques for high dimensional data}.
\newblock \bibinfo{journal}{\emph{Information Fusion}}  \bibinfo{volume}{59} (\bibinfo{year}{2020}), \bibinfo{pages}{44--58}.
\newblock


\bibitem[Babaee et~al\mbox{.}(2016)]%
        {babaee2016discriminative}
\bibfield{author}{\bibinfo{person}{Mohammadreza Babaee}, \bibinfo{person}{Stefanos Tsoukalas}, \bibinfo{person}{Mahdad Babaee}, \bibinfo{person}{Gerhard Rigoll}, {and} \bibinfo{person}{Mihai Datcu}.} \bibinfo{year}{2016}\natexlab{}.
\newblock \showarticletitle{Discriminative nonnegative matrix factorization for dimensionality reduction}.
\newblock \bibinfo{journal}{\emph{Neurocomputing}}  \bibinfo{volume}{173} (\bibinfo{year}{2016}), \bibinfo{pages}{212--223}.
\newblock


\bibitem[Belachew(2019)]%
        {belachew2019efficient}
\bibfield{author}{\bibinfo{person}{M.~T. Belachew}.} \bibinfo{year}{2019}\natexlab{}.
\newblock \showarticletitle{Efficient Algorithm for Sparse Symmetric Nonnegative Matrix Factorization}.
\newblock \bibinfo{journal}{\emph{Pattern Recognition Letters}}  \bibinfo{volume}{125} (\bibinfo{year}{2019}), \bibinfo{pages}{735--741}.
\newblock


\bibitem[Belachew and Del~Buono(2020)]%
        {belachew2020hybrid}
\bibfield{author}{\bibinfo{person}{M.~T. Belachew} {and} \bibinfo{person}{N. Del~Buono}.} \bibinfo{year}{2020}\natexlab{}.
\newblock \showarticletitle{Hybrid projective nonnegative matrix factorization based on $\alpha$-divergence and the alternating least squares algorithm}.
\newblock \bibinfo{journal}{\emph{Appl. Math. Comput.}}  \bibinfo{volume}{369} (\bibinfo{year}{2020}), \bibinfo{pages}{124825}.
\newblock


\bibitem[Cai et~al\mbox{.}(2011)]%
        {cf5567104}
\bibfield{author}{\bibinfo{person}{Deng Cai}, \bibinfo{person}{Xiaofei He}, {and} \bibinfo{person}{Jiawei Han}.} \bibinfo{year}{2011}\natexlab{}.
\newblock \showarticletitle{Locally Consistent Concept Factorization for Document Clustering}.
\newblock \bibinfo{journal}{\emph{IEEE Transactions on Knowledge and Data Engineering}} \bibinfo{volume}{23}, \bibinfo{number}{6} (\bibinfo{year}{2011}), \bibinfo{pages}{902--913}.
\newblock


\bibitem[Cai et~al\mbox{.}(2010)]%
        {cai2010graph}
\bibfield{author}{\bibinfo{person}{Deng Cai}, \bibinfo{person}{Xiaofei He}, \bibinfo{person}{Jiawei Han}, {and} \bibinfo{person}{Thomas~S Huang}.} \bibinfo{year}{2010}\natexlab{}.
\newblock \showarticletitle{Graph regularized nonnegative matrix factorization for data representation}.
\newblock \bibinfo{journal}{\emph{IEEE Transactions on Pattern Analysis and Machine Intelligence}} \bibinfo{volume}{33}, \bibinfo{number}{8} (\bibinfo{year}{2010}), \bibinfo{pages}{1548--1560}.
\newblock


\bibitem[Cai et~al\mbox{.}(2013)]%
        {cai2013nonnegative}
\bibfield{author}{\bibinfo{person}{X. Cai}, \bibinfo{person}{Y. Chen}, {and} \bibinfo{person}{D. Han}.} \bibinfo{year}{2013}\natexlab{}.
\newblock \showarticletitle{Nonnegative tensor factorizations using an alternating direction method}.
\newblock \bibinfo{journal}{\emph{Frontiers of Mathematics in China}}  \bibinfo{volume}{8} (\bibinfo{year}{2013}), \bibinfo{pages}{3--18}.
\newblock


\bibitem[Chao et~al\mbox{.}(2019)]%
        {chao2019recent}
\bibfield{author}{\bibinfo{person}{Guoqing Chao}, \bibinfo{person}{Yuan Luo}, {and} \bibinfo{person}{Weiping Ding}.} \bibinfo{year}{2019}\natexlab{}.
\newblock \showarticletitle{Recent advances in supervised dimension reduction: A survey}.
\newblock \bibinfo{journal}{\emph{Machine learning and knowledge extraction}} \bibinfo{volume}{1}, \bibinfo{number}{1} (\bibinfo{year}{2019}), \bibinfo{pages}{341--358}.
\newblock


\bibitem[Charikar and Hu(2021)]%
        {pmlr-v130-charikar21a}
\bibfield{author}{\bibinfo{person}{Moses Charikar} {and} \bibinfo{person}{Lunjia Hu}.} \bibinfo{year}{2021}\natexlab{}.
\newblock \showarticletitle{Approximation Algorithms for Orthogonal Non-negative Matrix Factorization}. In \bibinfo{booktitle}{\emph{Proceedings of The 24th International Conference on Artificial Intelligence and Statistics}} \emph{(\bibinfo{series}{Proceedings of Machine Learning Research}, Vol.~\bibinfo{volume}{130})}, \bibfield{editor}{\bibinfo{person}{Arindam Banerjee} {and} \bibinfo{person}{Kenji Fukumizu}} (Eds.). \bibinfo{publisher}{PMLR}, \bibinfo{pages}{2728--2736}.
\newblock


\bibitem[Chavoshinejad et~al\mbox{.}(2023)]%
        {CHAVOSHINEJAD2023109282}
\bibfield{author}{\bibinfo{person}{Jovan Chavoshinejad}, \bibinfo{person}{Seyed~Amjad Seyedi}, \bibinfo{person}{Fardin {Akhlaghian Tab}}, {and} \bibinfo{person}{Navid Salahian}.} \bibinfo{year}{2023}\natexlab{}.
\newblock \showarticletitle{Self-supervised semi-supervised nonnegative matrix factorization for data clustering}.
\newblock \bibinfo{journal}{\emph{Pattern Recognition}}  \bibinfo{volume}{137} (\bibinfo{year}{2023}), \bibinfo{pages}{109282}.
\newblock


\bibitem[Che et~al\mbox{.}(2023)]%
        {che10195864}
\bibfield{author}{\bibinfo{person}{Hangjun Che}, \bibinfo{person}{Baicheng Pan}, \bibinfo{person}{Man-Fai Leung}, \bibinfo{person}{Yuting Cao}, {and} \bibinfo{person}{Zheng Yan}.} \bibinfo{year}{2023}\natexlab{}.
\newblock \showarticletitle{Tensor Factorization With Sparse and Graph Regularization for Fake News Detection on Social Networks}.
\newblock \bibinfo{journal}{\emph{IEEE Transactions on Computational Social Systems}} (\bibinfo{year}{2023}), \bibinfo{pages}{1--11}.
\newblock


\bibitem[Chen et~al\mbox{.}(2023b)]%
        {CP9737358}
\bibfield{author}{\bibinfo{person}{Bilian Chen}, \bibinfo{person}{Jiewen Guan}, {and} \bibinfo{person}{Zhening Li}.} \bibinfo{year}{2023}\natexlab{b}.
\newblock \showarticletitle{Unsupervised Feature Selection via Graph Regularized Nonnegative {CP} Decomposition}.
\newblock \bibinfo{journal}{\emph{IEEE Transactions on Pattern Analysis and Machine Intelligence}} \bibinfo{volume}{45}, \bibinfo{number}{2} (\bibinfo{year}{2023}), \bibinfo{pages}{2582--2594}.
\newblock
\urldef\tempurl%
\url{https://doi.org/10.1109/TPAMI.2022.3160205}
\showDOI{\tempurl}


\bibitem[Chen et~al\mbox{.}(2023d)]%
        {chen2023adaptive}
\bibfield{author}{\bibinfo{person}{D. Chen}, \bibinfo{person}{G. Zhou}, \bibinfo{person}{Y. Qiu}, {and} \bibinfo{person}{Y. Yu}.} \bibinfo{year}{2023}\natexlab{d}.
\newblock \showarticletitle{Adaptive graph regularized non-negative {T}ucker decomposition for multiway dimensionality reduction}.
\newblock \bibinfo{journal}{\emph{Multimedia Tools and Applications}} (\bibinfo{year}{2023}), \bibinfo{pages}{1--22}.
\newblock


\bibitem[Chen et~al\mbox{.}(2022)]%
        {chen2022link}
\bibfield{author}{\bibinfo{person}{G. Chen}, \bibinfo{person}{H. Wang}, \bibinfo{person}{Y. Fang}, {and} \bibinfo{person}{L. Jiang}.} \bibinfo{year}{2022}\natexlab{}.
\newblock \showarticletitle{Link prediction by deep non-negative matrix factorization}.
\newblock \bibinfo{journal}{\emph{Expert Systems with Applications}}  \bibinfo{volume}{188} (\bibinfo{year}{2022}), \bibinfo{pages}{115991}.
\newblock


\bibitem[Chen et~al\mbox{.}(2023a)]%
        {chen2023graph}
\bibfield{author}{\bibinfo{person}{K. Chen}, \bibinfo{person}{H. Che}, \bibinfo{person}{X. Li}, {and} \bibinfo{person}{et al.}} \bibinfo{year}{2023}\natexlab{a}.
\newblock \showarticletitle{Graph non-negative matrix factorization with alternative smoothed $L_0$ regularizations}.
\newblock \bibinfo{journal}{\emph{Neural Computing \& Applications}}  \bibinfo{volume}{35} (\bibinfo{year}{2023}), \bibinfo{pages}{9995–10009}.
\newblock


\bibitem[Chen et~al\mbox{.}(2017)]%
        {chen2017multilayer}
\bibfield{author}{\bibinfo{person}{L. Chen}, \bibinfo{person}{S. Chen}, {and} \bibinfo{person}{X. Guo}.} \bibinfo{year}{2017}\natexlab{}.
\newblock \showarticletitle{Multilayer {NMF} for blind unmixing of hyperspectral imagery with additional constraints}.
\newblock \bibinfo{journal}{\emph{Photogrammetric Engineering \& Remote Sensing}} \bibinfo{volume}{83}, \bibinfo{number}{4} (\bibinfo{year}{2017}), \bibinfo{pages}{307--316}.
\newblock


\bibitem[Chen et~al\mbox{.}(2023c)]%
        {chen2023bilevel}
\bibfield{author}{\bibinfo{person}{W.~S. Chen}, \bibinfo{person}{Z. Lian}, \bibinfo{person}{B. Pan}, {and} \bibinfo{person}{B. Chen}.} \bibinfo{year}{2023}\natexlab{c}.
\newblock \showarticletitle{Bi-level Optimization-based Projective Non-negative Matrix Factorization}.
\newblock \bibinfo{journal}{\emph{International Journal of Wavelets, Multiresolution and Information Processing}} \bibinfo{volume}{21}, \bibinfo{number}{01} (\bibinfo{year}{2023}), \bibinfo{pages}{2250041}.
\newblock


\bibitem[Chen et~al\mbox{.}(2018)]%
        {hashing8115178}
\bibfield{author}{\bibinfo{person}{Yong Chen}, \bibinfo{person}{Hui Zhang}, \bibinfo{person}{Xiaopeng Zhang}, {and} \bibinfo{person}{Rui Liu}.} \bibinfo{year}{2018}\natexlab{}.
\newblock \showarticletitle{Regularized Semi-non-negative Matrix Factorization for {H}ashing}.
\newblock \bibinfo{journal}{\emph{IEEE Transactions on Multimedia}} \bibinfo{volume}{20}, \bibinfo{number}{7} (\bibinfo{year}{2018}), \bibinfo{pages}{1823--1836}.
\newblock


\bibitem[Cichocki et~al\mbox{.}(2008)]%
        {cichocki2008nonnegative}
\bibfield{author}{\bibinfo{person}{Andrzej Cichocki}, \bibinfo{person}{Haesun Lee}, \bibinfo{person}{Yoojin~D Kim}, {and} \bibinfo{person}{Sheng Choi}.} \bibinfo{year}{2008}\natexlab{}.
\newblock \showarticletitle{Non-negative matrix factorization with $\alpha$-divergence}.
\newblock \bibinfo{journal}{\emph{Pattern Recognition Letters}} \bibinfo{volume}{29}, \bibinfo{number}{9} (\bibinfo{year}{2008}), \bibinfo{pages}{1433--1440}.
\newblock


\bibitem[Cichocki and Zdunek(2006)]%
        {cichocki2006multilayer}
\bibfield{author}{\bibinfo{person}{Andrzej Cichocki} {and} \bibinfo{person}{Rafal Zdunek}.} \bibinfo{year}{2006}\natexlab{}.
\newblock \showarticletitle{Multilayer nonnegative matrix factorisation}.
\newblock \bibinfo{journal}{\emph{ELECTRONICS LETTERS-IEE}} \bibinfo{volume}{42}, \bibinfo{number}{16} (\bibinfo{year}{2006}), \bibinfo{pages}{947}.
\newblock


\bibitem[Cui et~al\mbox{.}(2018)]%
        {cui2018subspace}
\bibfield{author}{\bibinfo{person}{Guosheng Cui}, \bibinfo{person}{Xuelong Li}, {and} \bibinfo{person}{Yongsheng Dong}.} \bibinfo{year}{2018}\natexlab{}.
\newblock \showarticletitle{Subspace clustering guided convex nonnegative matrix factorization}.
\newblock \bibinfo{journal}{\emph{Neurocomputing}}  \bibinfo{volume}{292} (\bibinfo{year}{2018}), \bibinfo{pages}{38--48}.
\newblock


\bibitem[Cunningham and Ghahramani(2015)]%
        {cunningham2015linear}
\bibfield{author}{\bibinfo{person}{John~P Cunningham} {and} \bibinfo{person}{Zoubin Ghahramani}.} \bibinfo{year}{2015}\natexlab{}.
\newblock \showarticletitle{Linear dimensionality reduction: Survey, insights, and generalizations}.
\newblock \bibinfo{journal}{\emph{The Journal of Machine Learning Research}} \bibinfo{volume}{16}, \bibinfo{number}{1} (\bibinfo{year}{2015}), \bibinfo{pages}{2859--2900}.
\newblock


\bibitem[De~Handschutter et~al\mbox{.}(2021)]%
        {de2021survey}
\bibfield{author}{\bibinfo{person}{Pierre De~Handschutter}, \bibinfo{person}{Nicolas Gillis}, {and} \bibinfo{person}{Xavier Siebert}.} \bibinfo{year}{2021}\natexlab{}.
\newblock \showarticletitle{A survey on deep matrix factorizations}.
\newblock \bibinfo{journal}{\emph{Computer Science Review}}  \bibinfo{volume}{42} (\bibinfo{year}{2021}), \bibinfo{pages}{100423}.
\newblock


\bibitem[Del~Buono and Pio(2015)]%
        {delbuono2015nonnegative}
\bibfield{author}{\bibinfo{person}{N. Del~Buono} {and} \bibinfo{person}{G. Pio}.} \bibinfo{year}{2015}\natexlab{}.
\newblock \showarticletitle{Non-negative Matrix Tri-factorization for Co-clustering: {A}n Analysis of the Block Matrix}.
\newblock \bibinfo{journal}{\emph{Information Sciences}}  \bibinfo{volume}{301} (\bibinfo{year}{2015}), \bibinfo{pages}{13--26}.
\newblock


\bibitem[Deng and Moore(1998)]%
        {deng1998greediness}
\bibfield{author}{\bibinfo{person}{Kan Deng} {and} \bibinfo{person}{Andrew~W Moore}.} \bibinfo{year}{1998}\natexlab{}.
\newblock \bibinfo{booktitle}{\emph{On Greediness of Feature Selection Algorithms}}.
\newblock \bibinfo{publisher}{Citeseer}.
\newblock


\bibitem[Deng et~al\mbox{.}(2023)]%
        {9737322L1}
\bibfield{author}{\bibinfo{person}{Ping Deng}, \bibinfo{person}{Tianrui Li}, \bibinfo{person}{Hongjun Wang}, \bibinfo{person}{Dexian Wang}, \bibinfo{person}{Shi-Jinn Horng}, {and} \bibinfo{person}{Rui Liu}.} \bibinfo{year}{2023}\natexlab{}.
\newblock \showarticletitle{Graph Regularized Sparse Non-Negative Matrix Factorization for Clustering}.
\newblock \bibinfo{journal}{\emph{IEEE Transactions on Computational Social Systems}} \bibinfo{volume}{10}, \bibinfo{number}{3} (\bibinfo{year}{2023}), \bibinfo{pages}{910--921}.
\newblock


\bibitem[Ding et~al\mbox{.}(2005)]%
        {ding2005equivalence}
\bibfield{author}{\bibinfo{person}{Chris Ding}, \bibinfo{person}{Xiaofeng He}, {and} \bibinfo{person}{Horst~D Simon}.} \bibinfo{year}{2005}\natexlab{}.
\newblock \showarticletitle{On the equivalence of nonnegative matrix factorization and spectral clustering}. In \bibinfo{booktitle}{\emph{Proceedings of the 2005 SIAM International Conference on Data Mining}}. SIAM, \bibinfo{pages}{606--610}.
\newblock


\bibitem[Ding et~al\mbox{.}(2006)]%
        {ding2006orthogonal}
\bibfield{author}{\bibinfo{person}{Chris Ding}, \bibinfo{person}{Tao Li}, \bibinfo{person}{Wei Peng}, {and} \bibinfo{person}{Hanghang Park}.} \bibinfo{year}{2006}\natexlab{}.
\newblock \showarticletitle{Orthogonal nonnegative matrix t-factorizations for clustering}. In \bibinfo{booktitle}{\emph{Proceedings of the 12th ACM SIGKDD international conference on Knowledge discovery and data mining}}. \bibinfo{pages}{126--135}.
\newblock


\bibitem[Ding and Peng(2005)]%
        {ding2005minimum}
\bibfield{author}{\bibinfo{person}{Chris Ding} {and} \bibinfo{person}{Hanchuan Peng}.} \bibinfo{year}{2005}\natexlab{}.
\newblock \showarticletitle{Minimum redundancy feature selection from microarray gene expression data}.
\newblock \bibinfo{journal}{\emph{Journal of bioinformatics and computational biology}} \bibinfo{volume}{3}, \bibinfo{number}{02} (\bibinfo{year}{2005}), \bibinfo{pages}{185--205}.
\newblock


\bibitem[Ding et~al\mbox{.}(2008)]%
        {ding2008convex}
\bibfield{author}{\bibinfo{person}{Chris~HQ Ding}, \bibinfo{person}{Tao Li}, {and} \bibinfo{person}{Michael~I Jordan}.} \bibinfo{year}{2008}\natexlab{}.
\newblock \showarticletitle{Convex and semi-nonnegative matrix factorizations}.
\newblock \bibinfo{journal}{\emph{IEEE Transactions on Pattern Analysis and Machine Intelligence}} \bibinfo{volume}{32}, \bibinfo{number}{1} (\bibinfo{year}{2008}), \bibinfo{pages}{45--55}.
\newblock


\bibitem[Ding et~al\mbox{.}(2016)]%
        {ding2016locality}
\bibfield{author}{\bibinfo{person}{Jie Ding}, \bibinfo{person}{Changyun Wen}, \bibinfo{person}{Guoqi Li}, {and} \bibinfo{person}{Chin~Seng Chua}.} \bibinfo{year}{2016}\natexlab{}.
\newblock \showarticletitle{Locality sensitive batch feature extraction for high-dimensional data}.
\newblock \bibinfo{journal}{\emph{Neurocomputing}}  \bibinfo{volume}{171} (\bibinfo{year}{2016}), \bibinfo{pages}{664--672}.
\newblock


\bibitem[Du and Shen(2015)]%
        {FSASL}
\bibfield{author}{\bibinfo{person}{Liang Du} {and} \bibinfo{person}{Yi-Dong Shen}.} \bibinfo{year}{2015}\natexlab{}.
\newblock \showarticletitle{Unsupervised feature selection with adaptive structure learning}. In \bibinfo{booktitle}{\emph{Proceedings of the 21st ACM SIGKDD International Conference on Knowledge Discovery and Data Mining}}. ACM, \bibinfo{pages}{1903--1912}.
\newblock


\bibitem[Eskandari and Seifaddini(2023)]%
        {eskandari2023online}
\bibfield{author}{\bibinfo{person}{S Eskandari} {and} \bibinfo{person}{M Seifaddini}.} \bibinfo{year}{2023}\natexlab{}.
\newblock \showarticletitle{Online and offline streaming feature selection methods with bat algorithm for redundancy analysis}.
\newblock \bibinfo{journal}{\emph{Pattern Recognition}}  \bibinfo{volume}{133} (\bibinfo{year}{2023}), \bibinfo{pages}{109007}.
\newblock


\bibitem[Fan et~al\mbox{.}(2021)]%
        {fan2021adaptive}
\bibfield{author}{\bibinfo{person}{Mingyu Fan}, \bibinfo{person}{Xiaoqin Zhang}, \bibinfo{person}{Jie Hu}, \bibinfo{person}{Nannan Gu}, {and} \bibinfo{person}{Dacheng Tao}.} \bibinfo{year}{2021}\natexlab{}.
\newblock \showarticletitle{Adaptive data structure regularized multiclass discriminative feature selection}.
\newblock \bibinfo{journal}{\emph{IEEE Transactions on Neural Networks and Learning Systems}} \bibinfo{volume}{33}, \bibinfo{number}{10} (\bibinfo{year}{2021}), \bibinfo{pages}{5859--5872}.
\newblock


\bibitem[F{\'e}votte and Idier(2011)]%
        {fevotte2011algorithms}
\bibfield{author}{\bibinfo{person}{C{\'e}dric F{\'e}votte} {and} \bibinfo{person}{J{\'e}r{\^o}me Idier}.} \bibinfo{year}{2011}\natexlab{}.
\newblock \showarticletitle{Algorithms for nonnegative matrix factorization with the $\beta$-divergence}.
\newblock \bibinfo{journal}{\emph{Neural Computation}} \bibinfo{volume}{23}, \bibinfo{number}{9} (\bibinfo{year}{2011}), \bibinfo{pages}{2421--2456}.
\newblock


\bibitem[F\'{e}votte and Idier(2011)]%
        {10.1162NECOa00168}
\bibfield{author}{\bibinfo{person}{C\'{e}dric F\'{e}votte} {and} \bibinfo{person}{J\'{e}r\^{o}me Idier}.} \bibinfo{year}{2011}\natexlab{}.
\newblock \showarticletitle{{Algorithms for Nonnegative Matrix Factorization with the $\beta$-Divergence}}.
\newblock \bibinfo{journal}{\emph{Neural Computation}} \bibinfo{volume}{23}, \bibinfo{number}{9} (\bibinfo{year}{2011}), \bibinfo{pages}{2421--2456}.
\newblock


\bibitem[Friedman(2001)]%
        {friedman2001greedy}
\bibfield{author}{\bibinfo{person}{Jerome~H Friedman}.} \bibinfo{year}{2001}\natexlab{}.
\newblock \showarticletitle{Greedy function approximation: a gradient boosting machine}.
\newblock \bibinfo{journal}{\emph{Annals of statistics}} (\bibinfo{year}{2001}), \bibinfo{pages}{1189--1232}.
\newblock


\bibitem[Gan et~al\mbox{.}(2021)]%
        {gan2021nonnegative}
\bibfield{author}{\bibinfo{person}{J. Gan}, \bibinfo{person}{T. Liu}, \bibinfo{person}{L. Li}, {and} \bibinfo{person}{J. Zhang}.} \bibinfo{year}{2021}\natexlab{}.
\newblock \showarticletitle{Non-negative matrix factorization: {A} survey}.
\newblock \bibinfo{journal}{\emph{Comput. J.}} \bibinfo{volume}{64}, \bibinfo{number}{7} (\bibinfo{year}{2021}), \bibinfo{pages}{1080--1092}.
\newblock


\bibitem[Gong et~al\mbox{.}(2023)]%
        {GONG202375}
\bibfield{author}{\bibinfo{person}{Wenwu Gong}, \bibinfo{person}{Zhejun Huang}, {and} \bibinfo{person}{Lili Yang}.} \bibinfo{year}{2023}\natexlab{}.
\newblock \showarticletitle{Accurate regularized {T}ucker decomposition for image restoration}.
\newblock \bibinfo{journal}{\emph{Applied Mathematical Modelling}}  \bibinfo{volume}{123} (\bibinfo{year}{2023}), \bibinfo{pages}{75--86}.
\newblock


\bibitem[Gu and Zhou(2009)]%
        {gu2009co}
\bibfield{author}{\bibinfo{person}{Quanquan Gu} {and} \bibinfo{person}{Jie Zhou}.} \bibinfo{year}{2009}\natexlab{}.
\newblock \showarticletitle{Co-clustering on manifolds}. In \bibinfo{booktitle}{\emph{Proceedings of the 15th ACM SIGKDD international conference on Knowledge discovery and data mining}}.
\newblock


\bibitem[Guan et~al\mbox{.}(2012)]%
        {2012-6166359}
\bibfield{author}{\bibinfo{person}{Naiyang Guan}, \bibinfo{person}{Dacheng Tao}, \bibinfo{person}{Zhigang Luo}, {and} \bibinfo{person}{Bo Yuan}.} \bibinfo{year}{2012}\natexlab{}.
\newblock \showarticletitle{{N}e{NMF}: An Optimal Gradient Method for Nonnegative Matrix Factorization}.
\newblock \bibinfo{journal}{\emph{IEEE Transactions on Signal Processing}} \bibinfo{volume}{60}, \bibinfo{number}{6} (\bibinfo{year}{2012}), \bibinfo{pages}{2882--2898}.
\newblock


\bibitem[Guyon et~al\mbox{.}(2002)]%
        {guyon2002gene}
\bibfield{author}{\bibinfo{person}{Isabelle Guyon}, \bibinfo{person}{Jason Weston}, \bibinfo{person}{Stephen Barnhill}, {and} \bibinfo{person}{Vladimir Vapnik}.} \bibinfo{year}{2002}\natexlab{}.
\newblock \showarticletitle{Gene selection for cancer classification using support vector machines}.
\newblock \bibinfo{journal}{\emph{Machine learning}}  \bibinfo{volume}{46} (\bibinfo{year}{2002}), \bibinfo{pages}{389--422}.
\newblock


\bibitem[Hall(1999)]%
        {hall1999correlation}
\bibfield{author}{\bibinfo{person}{Mark~A Hall}.} \bibinfo{year}{1999}\natexlab{}.
\newblock \emph{\bibinfo{title}{Correlation-based feature selection for machine learning}}.
\newblock \bibinfo{thesistype}{Ph.\,D. Dissertation}. \bibinfo{school}{The University of Waikato}.
\newblock


\bibitem[He et~al\mbox{.}(2020)]%
        {he2020lowrank}
\bibfield{author}{\bibinfo{person}{Pengfei He}, \bibinfo{person}{Xinyi Xu}, \bibinfo{person}{Jundong Ding}, {and} \bibinfo{person}{Biao Fan}.} \bibinfo{year}{2020}\natexlab{}.
\newblock \showarticletitle{Low-Rank Nonnegative Matrix Factorization on Stiefel Manifold}.
\newblock \bibinfo{journal}{\emph{Information Sciences}}  \bibinfo{volume}{514} (\bibinfo{year}{2020}), \bibinfo{pages}{131--148}.
\newblock


\bibitem[He et~al\mbox{.}(2011)]%
        {he2011symmetric}
\bibfield{author}{\bibinfo{person}{Zhaofan He}, \bibinfo{person}{Shengli Xie}, \bibinfo{person}{Rafa{\l} Zdunek}, \bibinfo{person}{Guoxu Zhou}, {and} \bibinfo{person}{Andrzej Cichocki}.} \bibinfo{year}{2011}\natexlab{}.
\newblock \showarticletitle{Symmetric nonnegative matrix factorization: Algorithms and applications to probabilistic clustering}.
\newblock \bibinfo{journal}{\emph{IEEE Transactions on Neural Networks}}  \bibinfo{volume}{22} (\bibinfo{year}{2011}), \bibinfo{pages}{2117–2131}.
\newblock


\bibitem[Hoseinipour et~al\mbox{.}(2023)]%
        {hoseinipour2023orthogonal}
\bibfield{author}{\bibinfo{person}{S. Hoseinipour}, \bibinfo{person}{M. Aminghafari}, {and} \bibinfo{person}{A. Mohammadpour}.} \bibinfo{year}{2023}\natexlab{}.
\newblock \showarticletitle{Orthogonal parametric non-negative matrix tri-factorization with $\alpha$-divergence for co-clustering}.
\newblock \bibinfo{journal}{\emph{Expert Systems with Applications}} (\bibinfo{year}{2023}), \bibinfo{pages}{120680}.
\newblock


\bibitem[Hoyer(2004)]%
        {hoyer2004non}
\bibfield{author}{\bibinfo{person}{Patrik~O Hoyer}.} \bibinfo{year}{2004}\natexlab{}.
\newblock \showarticletitle{Non-negative matrix factorization with sparseness constraints}.
\newblock \bibinfo{journal}{\emph{Journal of Machine Learning Research}} \bibinfo{volume}{5}, \bibinfo{number}{9} (\bibinfo{year}{2004}).
\newblock


\bibitem[Hu et~al\mbox{.}(2014)]%
        {hu2014convergent}
\bibfield{author}{\bibinfo{person}{L. Hu}, \bibinfo{person}{L. Dai}, {and} \bibinfo{person}{J. Wu}.} \bibinfo{year}{2014}\natexlab{}.
\newblock \showarticletitle{Convergent projective non-negative matrix factorization with Kullback--Leibler divergence}.
\newblock \bibinfo{journal}{\emph{Pattern Recognition Letters}}  \bibinfo{volume}{36} (\bibinfo{year}{2014}), \bibinfo{pages}{15--21}.
\newblock


\bibitem[Hu et~al\mbox{.}(2022)]%
        {hu2022feature}
\bibfield{author}{\bibinfo{person}{Lianzhang Hu}, \bibinfo{person}{Nan Wu}, {and} \bibinfo{person}{Xuelong Li}.} \bibinfo{year}{2022}\natexlab{}.
\newblock \showarticletitle{Feature nonlinear transformation non-negative matrix factorization with Kullback-Leibler divergence}.
\newblock \bibinfo{journal}{\emph{Pattern Recognition}}  \bibinfo{volume}{132} (\bibinfo{year}{2022}), \bibinfo{pages}{108906}.
\newblock


\bibitem[Huang et~al\mbox{.}(2023)]%
        {huang2023unsupervised}
\bibfield{author}{\bibinfo{person}{Pei Huang}, \bibinfo{person}{Mengying Xie}, {and} \bibinfo{person}{Xiaowei Yang}.} \bibinfo{year}{2023}\natexlab{}.
\newblock \showarticletitle{Unsupervised Feature Selection via Controllable Adaptive Graph Learning and Discriminative Feature Learning}.
\newblock \bibinfo{journal}{\emph{IEEE Transactions on Neural Networks and Learning Systems}} (\bibinfo{year}{2023}).
\newblock


\bibitem[Huang et~al\mbox{.}(2020)]%
        {huang2020regularized}
\bibfield{author}{\bibinfo{person}{S. Huang}, \bibinfo{person}{Z. Xu}, \bibinfo{person}{Z. Kang}, {and} \bibinfo{person}{Y. Ren}.} \bibinfo{year}{2020}\natexlab{}.
\newblock \showarticletitle{Regularized nonnegative matrix factorization with adaptive local structure learning}.
\newblock \bibinfo{journal}{\emph{Neurocomputing}}  \bibinfo{volume}{382} (\bibinfo{year}{2020}), \bibinfo{pages}{196--209}.
\newblock


\bibitem[Huang(2015)]%
        {huang2015supervised}
\bibfield{author}{\bibinfo{person}{Samuel~H Huang}.} \bibinfo{year}{2015}\natexlab{}.
\newblock \showarticletitle{Supervised feature selection: A tutorial.}
\newblock \bibinfo{journal}{\emph{Artif. Intell. Res.}} \bibinfo{volume}{4}, \bibinfo{number}{2} (\bibinfo{year}{2015}), \bibinfo{pages}{22--37}.
\newblock


\bibitem[Jahani et~al\mbox{.}(2023)]%
        {jahani2023unsupervised}
\bibfield{author}{\bibinfo{person}{Mahsa~Samareh Jahani}, \bibinfo{person}{Gholamreza Aghamollaei}, \bibinfo{person}{Mahdi Eftekhari}, {and} \bibinfo{person}{Farid Saberi-Movahed}.} \bibinfo{year}{2023}\natexlab{}.
\newblock \showarticletitle{Unsupervised feature selection guided by orthogonal representation of feature space}.
\newblock \bibinfo{journal}{\emph{Neurocomputing}}  \bibinfo{volume}{516} (\bibinfo{year}{2023}), \bibinfo{pages}{61--76}.
\newblock


\bibitem[Ji et~al\mbox{.}(2015)]%
        {ji2015nmfbfs}
\bibfield{author}{\bibinfo{person}{Zhiwei Ji}, \bibinfo{person}{Guanmin Meng}, \bibinfo{person}{Deshuang Huang}, \bibinfo{person}{Xiaoqiang Yue}, \bibinfo{person}{Bing Wang}, {et~al\mbox{.}}} \bibinfo{year}{2015}\natexlab{}.
\newblock \showarticletitle{{NMFBFS}: A {NMF}-based feature selection method in identifying pivotal clinical symptoms of hepatocellular carcinoma}.
\newblock \bibinfo{journal}{\emph{Computational and Mathematical Methods in Medicine}}  \bibinfo{volume}{2015} (\bibinfo{year}{2015}).
\newblock


\bibitem[Jia et~al\mbox{.}(2019)]%
        {jia2019semi}
\bibfield{author}{\bibinfo{person}{Y. Jia}, \bibinfo{person}{S. Kwong}, \bibinfo{person}{J. Hou}, {and} \bibinfo{person}{W. Wu}.} \bibinfo{year}{2019}\natexlab{}.
\newblock \showarticletitle{Semi-supervised non-negative matrix factorization with dissimilarity and similarity regularization}.
\newblock \bibinfo{journal}{\emph{IEEE Transactions on Neural Networks and Learning Systems}} \bibinfo{volume}{31}, \bibinfo{number}{7} (\bibinfo{year}{2019}), \bibinfo{pages}{2510--2521}.
\newblock


\bibitem[Jia et~al\mbox{.}(2021)]%
        {jia2021selfsupervised}
\bibfield{author}{\bibinfo{person}{Y. Jia}, \bibinfo{person}{H. Liu}, \bibinfo{person}{J. Hou}, \bibinfo{person}{S. Kwong}, {and} \bibinfo{person}{Q. Zhang}.} \bibinfo{year}{2021}\natexlab{}.
\newblock \showarticletitle{Self-supervised symmetric nonnegative matrix factorization}.
\newblock \bibinfo{journal}{\emph{IEEE Transactions on Circuits and Systems for Video Technology}} \bibinfo{volume}{32}, \bibinfo{number}{7} (\bibinfo{year}{2021}), \bibinfo{pages}{4526--4537}.
\newblock


\bibitem[Jin et~al\mbox{.}(2019)]%
        {Jin2019}
\bibfield{author}{\bibinfo{person}{H. Jin}, \bibinfo{person}{W. Yu}, {and} \bibinfo{person}{S. Li}.} \bibinfo{year}{2019}\natexlab{}.
\newblock \showarticletitle{Graph regularized nonnegative matrix tri-factorization for overlapping community detection}.
\newblock \bibinfo{journal}{\emph{Physica A: Statistical Mechanics and its Applications}}  \bibinfo{volume}{515} (\bibinfo{year}{2019}), \bibinfo{pages}{376--387}.
\newblock


\bibitem[Jin et~al\mbox{.}(2022)]%
        {jin2022sparse}
\bibfield{author}{\bibinfo{person}{X. Jin}, \bibinfo{person}{J. Miao}, \bibinfo{person}{Q. Wang}, \bibinfo{person}{G. Geng}, {and} \bibinfo{person}{K. Huang}.} \bibinfo{year}{2022}\natexlab{}.
\newblock \showarticletitle{Sparse matrix factorization with $L_{2,1}$ norm for matrix completion}.
\newblock \bibinfo{journal}{\emph{Pattern Recognition}}  \bibinfo{volume}{127} (\bibinfo{year}{2022}), \bibinfo{pages}{108655}.
\newblock


\bibitem[Karami et~al\mbox{.}(2023)]%
        {KARAMI2023}
\bibfield{author}{\bibinfo{person}{Saeed Karami}, \bibinfo{person}{Farid Saberi-Movahed}, \bibinfo{person}{Prayag Tiwari}, \bibinfo{person}{Pekka Marttinen}, {and} \bibinfo{person}{Sahar Vahdati}.} \bibinfo{year}{2023}\natexlab{}.
\newblock \showarticletitle{Unsupervised feature selection based on variance–covariance subspace distance}.
\newblock \bibinfo{journal}{\emph{Neural Networks}}  \bibinfo{volume}{166} (\bibinfo{year}{2023}), \bibinfo{pages}{188--203}.
\newblock


\bibitem[Karimi et~al\mbox{.}(2023)]%
        {karimi2023semiaco}
\bibfield{author}{\bibinfo{person}{Fereshteh Karimi}, \bibinfo{person}{Mohammad~Bagher Dowlatshahi}, {and} \bibinfo{person}{Amin Hashemi}.} \bibinfo{year}{2023}\natexlab{}.
\newblock \showarticletitle{SemiACO: A semi-supervised feature selection based on ant colony optimization}.
\newblock \bibinfo{journal}{\emph{Expert Systems with Applications}}  \bibinfo{volume}{214} (\bibinfo{year}{2023}), \bibinfo{pages}{119130}.
\newblock


\bibitem[Kim and Park(2007)]%
        {kim2007sparse}
\bibfield{author}{\bibinfo{person}{H. Kim} {and} \bibinfo{person}{H. Park}.} \bibinfo{year}{2007}\natexlab{}.
\newblock \showarticletitle{Sparse non-negative matrix factorizations via alternating non-negativity-constrained least squares for microarray data analysis}.
\newblock \bibinfo{journal}{\emph{Bioinformatics}} \bibinfo{volume}{23}, \bibinfo{number}{12} (\bibinfo{year}{2007}), \bibinfo{pages}{1495--1502}.
\newblock


\bibitem[Kolda and Bader(2009)]%
        {baderkolda}
\bibfield{author}{\bibinfo{person}{Tamara~G. Kolda} {and} \bibinfo{person}{Brett~W. Bader}.} \bibinfo{year}{2009}\natexlab{}.
\newblock \showarticletitle{Tensor decompositions and applications}.
\newblock \bibinfo{journal}{\emph{SIAM Rev.}} \bibinfo{volume}{51}, \bibinfo{number}{3} (\bibinfo{year}{2009}), \bibinfo{pages}{455--500}.
\newblock


\bibitem[Kong et~al\mbox{.}(2011)]%
        {kong2011robust}
\bibfield{author}{\bibinfo{person}{D. Kong}, \bibinfo{person}{C. Ding}, {and} \bibinfo{person}{H. Huang}.} \bibinfo{year}{2011}\natexlab{}.
\newblock \showarticletitle{Robust nonnegative matrix factorization using $l_{21}$-norm}. In \bibinfo{booktitle}{\emph{Proceedings of the 20th ACM international conference on Information and knowledge management}}. \bibinfo{pages}{673--682}.
\newblock


\bibitem[Kuang et~al\mbox{.}(2015)]%
        {kuang2015symnmf}
\bibfield{author}{\bibinfo{person}{D. Kuang}, \bibinfo{person}{S. Yun}, {and} \bibinfo{person}{H. Park}.} \bibinfo{year}{2015}\natexlab{}.
\newblock \showarticletitle{{SYMNMF}: Nonnegative lowrank approximation of a similarity matrix for graph clustering}.
\newblock \bibinfo{journal}{\emph{J. Global Optim.}} \bibinfo{volume}{62}, \bibinfo{number}{3} (\bibinfo{year}{2015}), \bibinfo{pages}{545--574}.
\newblock


\bibitem[Lai et~al\mbox{.}(2022)]%
        {lai2022rmfrasl}
\bibfield{author}{\bibinfo{person}{Shumin Lai}, \bibinfo{person}{Longjun Huang}, \bibinfo{person}{Ping Li}, \bibinfo{person}{Zhenzhen Luo}, \bibinfo{person}{Jianzhong Wang}, {and} \bibinfo{person}{Yugen Yi}.} \bibinfo{year}{2022}\natexlab{}.
\newblock \showarticletitle{{RMFRASL}: Robust Matrix Factorization with Robust Adaptive Structure Learning for Feature Selection}.
\newblock \bibinfo{journal}{\emph{Algorithms}} \bibinfo{volume}{16}, \bibinfo{number}{1} (\bibinfo{year}{2022}), \bibinfo{pages}{14}.
\newblock


\bibitem[Latif et~al\mbox{.}(2021)]%
        {latif2021survey}
\bibfield{author}{\bibinfo{person}{Siddique Latif}, \bibinfo{person}{Rajib Rana}, \bibinfo{person}{Sara Khalifa}, \bibinfo{person}{Raja Jurdak}, \bibinfo{person}{Junaid Qadir}, {and} \bibinfo{person}{Bjoern~W Schuller}.} \bibinfo{year}{2021}\natexlab{}.
\newblock \showarticletitle{Survey of deep representation learning for speech emotion recognition}.
\newblock \bibinfo{journal}{\emph{IEEE Transactions on Affective Computing}} (\bibinfo{year}{2021}).
\newblock


\bibitem[Le(1990)]%
        {le1990optimal}
\bibfield{author}{\bibinfo{person}{Cun Le}.} \bibinfo{year}{1990}\natexlab{}.
\newblock \showarticletitle{Optimal brain damage}.
\newblock \bibinfo{journal}{\emph{Advances in Neural Information Processing Systems}}  \bibinfo{volume}{2} (\bibinfo{year}{1990}), \bibinfo{pages}{598--605}.
\newblock


\bibitem[Lee and Seung(2000)]%
        {lee2000algorithms}
\bibfield{author}{\bibinfo{person}{Daniel Lee} {and} \bibinfo{person}{H~Sebastian Seung}.} \bibinfo{year}{2000}\natexlab{}.
\newblock \showarticletitle{Algorithms for non-negative matrix factorization}. In \bibinfo{booktitle}{\emph{Advances in Neural Information Processing Systems}}, Vol.~\bibinfo{volume}{13}.
\newblock


\bibitem[Li et~al\mbox{.}(2017b)]%
        {LI20171}
\bibfield{author}{\bibinfo{person}{Guopeng Li}, \bibinfo{person}{Xinyu Zhang}, \bibinfo{person}{Siyi Zheng}, {and} \bibinfo{person}{Deyi Li}.} \bibinfo{year}{2017}\natexlab{b}.
\newblock \showarticletitle{Semi-supervised convex nonnegative matrix factorizations with graph regularized for image representation}.
\newblock \bibinfo{journal}{\emph{Neurocomputing}}  \bibinfo{volume}{237} (\bibinfo{year}{2017}), \bibinfo{pages}{1--11}.
\newblock


\bibitem[Li et~al\mbox{.}(2022)]%
        {9786852dual}
\bibfield{author}{\bibinfo{person}{Ning Li}, \bibinfo{person}{Chengcai Leng}, \bibinfo{person}{Irene Cheng}, \bibinfo{person}{Anup Basu}, {and} \bibinfo{person}{Licheng Jiao}.} \bibinfo{year}{2022}\natexlab{}.
\newblock \showarticletitle{Dual-Graph Global and Local Concept Factorization for Data Clustering}.
\newblock \bibinfo{journal}{\emph{IEEE Transactions on Neural Networks and Learning Systems}} (\bibinfo{year}{2022}), \bibinfo{pages}{1--14}.
\newblock


\bibitem[Li et~al\mbox{.}(2014)]%
        {li2014discriminative}
\bibfield{author}{\bibinfo{person}{Ping Li}, \bibinfo{person}{Jiajun Bu}, \bibinfo{person}{Yi Yang}, \bibinfo{person}{Rongrong Ji}, \bibinfo{person}{Chun Chen}, {and} \bibinfo{person}{Deng Cai}.} \bibinfo{year}{2014}\natexlab{}.
\newblock \showarticletitle{Discriminative Orthogonal Nonnegative Matrix Factorization with Flexibility for Data Representation}.
\newblock \bibinfo{journal}{\emph{Expert Systems with Applications}} \bibinfo{volume}{41}, \bibinfo{number}{4} (\bibinfo{year}{2014}), \bibinfo{pages}{1283--1293}.
\newblock


\bibitem[Li et~al\mbox{.}(2023a)]%
        {li2023semi}
\bibfield{author}{\bibinfo{person}{S. Li}, \bibinfo{person}{W. Li}, \bibinfo{person}{H. Lu}, {and} \bibinfo{person}{Y. Li}.} \bibinfo{year}{2023}\natexlab{a}.
\newblock \showarticletitle{Semi-supervised non-negative matrix tri-factorization with adaptive neighbors and block-diagonal learning}.
\newblock \bibinfo{journal}{\emph{Engineering Applications of Artificial Intelligence}}  \bibinfo{volume}{121} (\bibinfo{year}{2023}), \bibinfo{pages}{106043}.
\newblock


\bibitem[Li et~al\mbox{.}(2020)]%
        {li2020fast}
\bibfield{author}{\bibinfo{person}{Wenbo Li}, \bibinfo{person}{Jicheng Li}, \bibinfo{person}{Xuenian Liu}, {and} \bibinfo{person}{Liqiang Dong}.} \bibinfo{year}{2020}\natexlab{}.
\newblock \showarticletitle{Two Fast Vector-wise Update Algorithms for Orthogonal Nonnegative Matrix Factorization with Sparsity Constraint}.
\newblock \bibinfo{journal}{\emph{J. Comput. Appl. Math.}}  \bibinfo{volume}{375} (\bibinfo{year}{2020}), \bibinfo{pages}{112785}.
\newblock


\bibitem[Li et~al\mbox{.}(2016)]%
        {MR-NTD}
\bibfield{author}{\bibinfo{person}{Xutao Li} {et~al\mbox{.}}} \bibinfo{year}{2016}\natexlab{}.
\newblock \showarticletitle{{MR-NTD}: Manifold regularization nonnegative Tucker decomposition for tensor data dimension reduction and representation}.
\newblock \bibinfo{journal}{\emph{IEEE Transactions on Neural Networks and Learning Systems}} \bibinfo{volume}{28}, \bibinfo{number}{8} (\bibinfo{year}{2016}), \bibinfo{pages}{1787--1800}.
\newblock


\bibitem[Li et~al\mbox{.}(2017a)]%
        {7460200tucker}
\bibfield{author}{\bibinfo{person}{Xutao Li}, \bibinfo{person}{Michael~K. Ng}, \bibinfo{person}{Gao Cong}, \bibinfo{person}{Yunming Ye}, {and} \bibinfo{person}{Qingyao Wu}.} \bibinfo{year}{2017}\natexlab{a}.
\newblock \showarticletitle{{MR-NTD}: Manifold Regularization Nonnegative Tucker Decomposition for Tensor Data Dimension Reduction and Representation}.
\newblock \bibinfo{journal}{\emph{IEEE Transactions on Neural Networks and Learning Systems}} \bibinfo{volume}{28}, \bibinfo{number}{8} (\bibinfo{year}{2017}), \bibinfo{pages}{1787--1800}.
\newblock


\bibitem[Li et~al\mbox{.}(2021)]%
        {li2021semisupervised}
\bibfield{author}{\bibinfo{person}{Xuelong Li}, \bibinfo{person}{Yunxing Zhang}, {and} \bibinfo{person}{Rui Zhang}.} \bibinfo{year}{2021}\natexlab{}.
\newblock \showarticletitle{Semisupervised feature selection via generalized uncorrelated constraint and manifold embedding}.
\newblock \bibinfo{journal}{\emph{IEEE transactions on neural networks and learning systems}} \bibinfo{volume}{33}, \bibinfo{number}{9} (\bibinfo{year}{2021}), \bibinfo{pages}{5070--5079}.
\newblock


\bibitem[Li et~al\mbox{.}(2023b)]%
        {9606619}
\bibfield{author}{\bibinfo{person}{Xiao Li}, \bibinfo{person}{Zhihui Zhu}, \bibinfo{person}{Qiuwei Li}, {and} \bibinfo{person}{Kai Liu}.} \bibinfo{year}{2023}\natexlab{b}.
\newblock \showarticletitle{A Provable Splitting Approach for Symmetric Nonnegative Matrix Factorization}.
\newblock \bibinfo{journal}{\emph{IEEE Transactions on Knowledge and Data Engineering}} \bibinfo{volume}{35}, \bibinfo{number}{3} (\bibinfo{year}{2023}), \bibinfo{pages}{2206--2219}.
\newblock


\bibitem[Li et~al\mbox{.}(2018)]%
        {7902136}
\bibfield{author}{\bibinfo{person}{Zechao Li}, \bibinfo{person}{Jinhui Tang}, {and} \bibinfo{person}{Xiaofei He}.} \bibinfo{year}{2018}\natexlab{}.
\newblock \showarticletitle{Robust Structured Nonnegative Matrix Factorization for Image Representation}.
\newblock \bibinfo{journal}{\emph{IEEE Transactions on Neural Networks and Learning Systems}} \bibinfo{volume}{29}, \bibinfo{number}{5} (\bibinfo{year}{2018}), \bibinfo{pages}{1947--1960}.
\newblock


\bibitem[Liang et~al\mbox{.}(2016)]%
        {liang2016feature}
\bibfield{author}{\bibinfo{person}{Lin Liang}, \bibinfo{person}{Fei Liu}, \bibinfo{person}{Maolin Li}, \bibinfo{person}{Kangkang He}, {and} \bibinfo{person}{Guanghua Xu}.} \bibinfo{year}{2016}\natexlab{}.
\newblock \showarticletitle{Feature selection for machine fault diagnosis using clustering of non-negation matrix factorization}.
\newblock \bibinfo{journal}{\emph{Measurement}}  \bibinfo{volume}{94} (\bibinfo{year}{2016}), \bibinfo{pages}{295--305}.
\newblock


\bibitem[Liang et~al\mbox{.}(2020)]%
        {liang2020multi}
\bibfield{author}{\bibinfo{person}{Nan Liang}, \bibinfo{person}{Zhenqiu Yang}, \bibinfo{person}{Zilong Li}, \bibinfo{person}{Wei Sun}, {and} \bibinfo{person}{Shengli Xie}.} \bibinfo{year}{2020}\natexlab{}.
\newblock \showarticletitle{Multi-view Clustering by Non-negative Matrix Factorization with Co-orthogonal Constraints}.
\newblock \bibinfo{journal}{\emph{Knowledge-Based Systems}}  \bibinfo{volume}{194} (\bibinfo{year}{2020}), \bibinfo{pages}{105582}.
\newblock


\bibitem[Lin et~al\mbox{.}(2019)]%
        {lin2019overview}
\bibfield{author}{\bibinfo{person}{Renjie Lin}, \bibinfo{person}{Shiping Wang}, {and} \bibinfo{person}{Wenzhong Guo}.} \bibinfo{year}{2019}\natexlab{}.
\newblock \showarticletitle{An overview of co-clustering via matrix factorization}.
\newblock \bibinfo{journal}{\emph{IEEE Access}}  \bibinfo{volume}{7} (\bibinfo{year}{2019}), \bibinfo{pages}{33481--33493}.
\newblock


\bibitem[Lin et~al\mbox{.}(2021)]%
        {lin2021unsupervised}
\bibfield{author}{\bibinfo{person}{Xutao Lin}, \bibinfo{person}{Jian Guan}, \bibinfo{person}{Bing Chen}, {and} \bibinfo{person}{Yang Zeng}.} \bibinfo{year}{2021}\natexlab{}.
\newblock \showarticletitle{Unsupervised feature selection via orthogonal basis clustering and local structure preserving}.
\newblock \bibinfo{journal}{\emph{IEEE Transactions on Neural Networks and Learning Systems}} \bibinfo{volume}{33}, \bibinfo{number}{11} (\bibinfo{year}{2021}), \bibinfo{pages}{6881--6892}.
\newblock


\bibitem[Liu et~al\mbox{.}(2011)]%
        {liu2011constrained}
\bibfield{author}{\bibinfo{person}{Hao Liu}, \bibinfo{person}{Zhongqiu Wu}, \bibinfo{person}{Xuelong Li}, \bibinfo{person}{Deng Cai}, {and} \bibinfo{person}{Thomas~S Huang}.} \bibinfo{year}{2011}\natexlab{}.
\newblock \showarticletitle{Constrained nonnegative matrix factorization for image representation}.
\newblock \bibinfo{journal}{\emph{IEEE Transactions on Pattern Analysis and Machine Intelligence}} \bibinfo{volume}{34}, \bibinfo{number}{7} (\bibinfo{year}{2011}), \bibinfo{pages}{1299--1311}.
\newblock


\bibitem[Liu et~al\mbox{.}(2023a)]%
        {LIU2023109806}
\bibfield{author}{\bibinfo{person}{Xiangnan Liu}, \bibinfo{person}{Shifei Ding}, \bibinfo{person}{Xiao Xu}, {and} \bibinfo{person}{Lijuan Wang}.} \bibinfo{year}{2023}\natexlab{a}.
\newblock \showarticletitle{Deep manifold regularized semi-nonnegative matrix factorization for Multi-view Clustering}.
\newblock \bibinfo{journal}{\emph{Applied Soft Computing}}  \bibinfo{volume}{132} (\bibinfo{year}{2023}), \bibinfo{pages}{109806}.
\newblock


\bibitem[Liu et~al\mbox{.}(2023b)]%
        {10072010}
\bibfield{author}{\bibinfo{person}{Zhigang Liu}, \bibinfo{person}{Xin Luo}, {and} \bibinfo{person}{Mengchu Zhou}.} \bibinfo{year}{2023}\natexlab{b}.
\newblock \showarticletitle{Symmetry and Graph Bi-Regularized Non-Negative Matrix Factorization for Precise Community Detection}.
\newblock \bibinfo{journal}{\emph{IEEE Transactions on Automation Science and Engineering}} (\bibinfo{year}{2023}), \bibinfo{pages}{1--15}.
\newblock


\bibitem[Lu et~al\mbox{.}(2023)]%
        {lu2023robust}
\bibfield{author}{\bibinfo{person}{G. Lu}, \bibinfo{person}{C. Leng}, \bibinfo{person}{B. Li}, \bibinfo{person}{L. Jiao}, {and} \bibinfo{person}{A. Basu}.} \bibinfo{year}{2023}\natexlab{}.
\newblock \showarticletitle{Robust dual-graph discriminative {NMF} for data classification}.
\newblock \bibinfo{journal}{\emph{Knowledge-Based Systems}}  \bibinfo{volume}{268} (\bibinfo{year}{2023}), \bibinfo{pages}{110465}.
\newblock


\bibitem[Lu et~al\mbox{.}(2016)]%
        {lu2016projective}
\bibfield{author}{\bibinfo{person}{Yuwu Lu}, \bibinfo{person}{Zhihui Lai}, \bibinfo{person}{Yong Xu}, \bibinfo{person}{Jane You}, \bibinfo{person}{Xuelong Li}, {and} \bibinfo{person}{Chun Yuan}.} \bibinfo{year}{2016}\natexlab{}.
\newblock \showarticletitle{Projective robust nonnegative factorization}.
\newblock \bibinfo{journal}{\emph{Information Sciences}}  \bibinfo{volume}{364} (\bibinfo{year}{2016}), \bibinfo{pages}{16--32}.
\newblock


\bibitem[Luo et~al\mbox{.}(2022)]%
        {luo2022orthogonally}
\bibfield{author}{\bibinfo{person}{Chuan Luo}, \bibinfo{person}{Jian Zheng}, \bibinfo{person}{Tianrui Li}, \bibinfo{person}{Hongmei Chen}, \bibinfo{person}{Yanyong Huang}, {and} \bibinfo{person}{Xi Peng}.} \bibinfo{year}{2022}\natexlab{}.
\newblock \showarticletitle{Orthogonally constrained matrix factorization for robust unsupervised feature selection with local preserving}.
\newblock \bibinfo{journal}{\emph{Information Sciences}}  \bibinfo{volume}{586} (\bibinfo{year}{2022}), \bibinfo{pages}{662--675}.
\newblock


\bibitem[Luo et~al\mbox{.}(2021)]%
        {luo2021symmetric}
\bibfield{author}{\bibinfo{person}{X. Luo}, \bibinfo{person}{Z. Liu}, \bibinfo{person}{L. Jin}, \bibinfo{person}{Y. Zhou}, {and} \bibinfo{person}{M. Zhou}.} \bibinfo{year}{2021}\natexlab{}.
\newblock \showarticletitle{Symmetric Nonnegative Matrix Factorization-Based Community Detection Models and Their Convergence Analysis}.
\newblock \bibinfo{journal}{\emph{IEEE Transactions on Neural Networks and Learning Systems}} \bibinfo{volume}{33}, \bibinfo{number}{3} (\bibinfo{year}{2021}), \bibinfo{pages}{1203--1215}.
\newblock


\bibitem[Luong et~al\mbox{.}(2022)]%
        {LUONG20221088152022}
\bibfield{author}{\bibinfo{person}{Khanh Luong}, \bibinfo{person}{Richi Nayak}, \bibinfo{person}{Thirunavukarasu Balasubramaniam}, {and} \bibinfo{person}{Md~Abul Bashar}.} \bibinfo{year}{2022}\natexlab{}.
\newblock \showarticletitle{Multi-layer manifold learning for deep non-negative matrix factorization-based multi-view clustering}.
\newblock \bibinfo{journal}{\emph{Pattern Recognition}}  \bibinfo{volume}{131} (\bibinfo{year}{2022}), \bibinfo{pages}{108815}.
\newblock


\bibitem[Marmin et~al\mbox{.}(2023a)]%
        {marmin2023majorization}
\bibfield{author}{\bibinfo{person}{Arthur Marmin}, \bibinfo{person}{João~Henrique de Morais~Goulart}, {and} \bibinfo{person}{Cédric Févotte}.} \bibinfo{year}{2023}\natexlab{a}.
\newblock \showarticletitle{Majorization-minimization for Sparse Nonnegative Matrix Factorization with the $\beta$-divergence}.
\newblock \bibinfo{journal}{\emph{IEEE Transactions on Signal Processing}} (\bibinfo{year}{2023}).
\newblock


\bibitem[Marmin et~al\mbox{.}(2023b)]%
        {major10103209}
\bibfield{author}{\bibinfo{person}{Arthur Marmin}, \bibinfo{person}{Jose Henrique de~Morais Goulart}, {and} \bibinfo{person}{Cedric Fevotte}.} \bibinfo{year}{2023}\natexlab{b}.
\newblock \showarticletitle{Majorization-Minimization for Sparse Nonnegative Matrix Factorization With the $\beta$-Divergence}.
\newblock \bibinfo{journal}{\emph{IEEE Transactions on Signal Processing}}  \bibinfo{volume}{71} (\bibinfo{year}{2023}), \bibinfo{pages}{1435--1447}.
\newblock


\bibitem[McInnes et~al\mbox{.}(2018)]%
        {mcinnes2018umap}
\bibfield{author}{\bibinfo{person}{Leland McInnes}, \bibinfo{person}{John Healy}, {and} \bibinfo{person}{James Melville}.} \bibinfo{year}{2018}\natexlab{}.
\newblock \showarticletitle{Umap: Uniform manifold approximation and projection for dimension reduction}.
\newblock \bibinfo{journal}{\emph{arXiv preprint arXiv:1802.03426}} (\bibinfo{year}{2018}).
\newblock


\bibitem[Mehrpooya et~al\mbox{.}(2022)]%
        {mehrpooya2022high}
\bibfield{author}{\bibinfo{person}{Adel Mehrpooya}, \bibinfo{person}{Farid Saberi-Movahed}, \bibinfo{person}{Najmeh Azizizadeh}, \bibinfo{person}{Mohammad Rezaei-Ravari}, \bibinfo{person}{Farshad Saberi-Movahed}, \bibinfo{person}{Mahdi Eftekhari}, {and} \bibinfo{person}{Iman Tavassoly}.} \bibinfo{year}{2022}\natexlab{}.
\newblock \showarticletitle{High dimensionality reduction by matrix factorization for systems pharmacology}.
\newblock \bibinfo{journal}{\emph{Briefings in Bioinformatics}} \bibinfo{volume}{23}, \bibinfo{number}{1} (\bibinfo{year}{2022}), \bibinfo{pages}{bbab410}.
\newblock


\bibitem[Meng et~al\mbox{.}(2018)]%
        {meng2018feature}
\bibfield{author}{\bibinfo{person}{Yang Meng}, \bibinfo{person}{Ronghua Shang}, \bibinfo{person}{Licheng Jiao}, \bibinfo{person}{Wenya Zhang}, \bibinfo{person}{Yijing Yuan}, {and} \bibinfo{person}{Shuyuan Yang}.} \bibinfo{year}{2018}\natexlab{}.
\newblock \showarticletitle{Feature selection based dual-graph sparse non-negative matrix factorization for local discriminative clustering}.
\newblock \bibinfo{journal}{\emph{Neurocomputing}}  \bibinfo{volume}{290} (\bibinfo{year}{2018}), \bibinfo{pages}{87--99}.
\newblock


\bibitem[Meng et~al\mbox{.}(2020)]%
        {8858038semi2020}
\bibfield{author}{\bibinfo{person}{Yang Meng}, \bibinfo{person}{Ronghua Shang}, \bibinfo{person}{Fanhua Shang}, \bibinfo{person}{Licheng Jiao}, \bibinfo{person}{Shuyuan Yang}, {and} \bibinfo{person}{Rustam Stolkin}.} \bibinfo{year}{2020}\natexlab{}.
\newblock \showarticletitle{Semi-Supervised Graph Regularized Deep {NMF} With Bi-Orthogonal Constraints for Data Representation}.
\newblock \bibinfo{journal}{\emph{IEEE Transactions on Neural Networks and Learning Systems}} \bibinfo{volume}{31}, \bibinfo{number}{9} (\bibinfo{year}{2020}), \bibinfo{pages}{3245--3258}.
\newblock


\bibitem[Min et~al\mbox{.}(2023)]%
        {L209893402}
\bibfield{author}{\bibinfo{person}{Wenwen Min}, \bibinfo{person}{Taosheng Xu}, \bibinfo{person}{Xiang Wan}, {and} \bibinfo{person}{Tsung-Hui Chang}.} \bibinfo{year}{2023}\natexlab{}.
\newblock \showarticletitle{Structured Sparse Non-Negative Matrix Factorization With $\ell _{2,0}$-Norm}.
\newblock \bibinfo{journal}{\emph{IEEE Transactions on Knowledge and Data Engineering}} \bibinfo{volume}{35}, \bibinfo{number}{8} (\bibinfo{year}{2023}), \bibinfo{pages}{8584--8595}.
\newblock


\bibitem[Mu et~al\mbox{.}(2023)]%
        {mu2023dualgraph}
\bibfield{author}{\bibinfo{person}{J. Mu}, \bibinfo{person}{P. Song}, \bibinfo{person}{X. Liu}, {and} \bibinfo{person}{S. Li}.} \bibinfo{year}{2023}\natexlab{}.
\newblock \showarticletitle{Dual-graph regularized concept factorization for multi-view clustering}.
\newblock \bibinfo{journal}{\emph{Expert Systems with Applications}}  \bibinfo{volume}{223} (\bibinfo{year}{2023}), \bibinfo{pages}{119949}.
\newblock


\bibitem[Nie et~al\mbox{.}(2020)]%
        {nie2020unsupervised}
\bibfield{author}{\bibinfo{person}{Feiping Nie}, \bibinfo{person}{Xia Dong}, {and} \bibinfo{person}{Xuelong Li}.} \bibinfo{year}{2020}\natexlab{}.
\newblock \showarticletitle{Unsupervised and semisupervised projection with graph optimization}.
\newblock \bibinfo{journal}{\emph{IEEE transactions on neural networks and learning systems}} \bibinfo{volume}{32}, \bibinfo{number}{4} (\bibinfo{year}{2020}), \bibinfo{pages}{1547--1559}.
\newblock


\bibitem[Nie et~al\mbox{.}(2022)]%
        {nie2022discrete}
\bibfield{author}{\bibinfo{person}{Feiping Nie}, \bibinfo{person}{Sisi Wang}, \bibinfo{person}{Zheng Wang}, \bibinfo{person}{Rong Wang}, {and} \bibinfo{person}{Xuelong Li}.} \bibinfo{year}{2022}\natexlab{}.
\newblock \showarticletitle{Discrete robust principal component analysis via binary weights self-learning}.
\newblock \bibinfo{journal}{\emph{IEEE Transactions on Neural Networks and Learning Systems}} (\bibinfo{year}{2022}).
\newblock


\bibitem[Nie et~al\mbox{.}(2021)]%
        {nie2021adaptive}
\bibfield{author}{\bibinfo{person}{Feiping Nie}, \bibinfo{person}{Zheng Wang}, \bibinfo{person}{Rong Wang}, {and} \bibinfo{person}{Xuelong Li}.} \bibinfo{year}{2021}\natexlab{}.
\newblock \showarticletitle{Adaptive local embedding learning for semi-supervised dimensionality reduction}.
\newblock \bibinfo{journal}{\emph{IEEE Transactions on Knowledge and Data Engineering}} \bibinfo{volume}{34}, \bibinfo{number}{10} (\bibinfo{year}{2021}), \bibinfo{pages}{4609--4621}.
\newblock


\bibitem[Nie et~al\mbox{.}(2016)]%
        {Nieadaptive}
\bibfield{author}{\bibinfo{person}{Feiping Nie}, \bibinfo{person}{Wei Zhu}, {and} \bibinfo{person}{Xuelong Li}.} \bibinfo{year}{2016}\natexlab{}.
\newblock \showarticletitle{Unsupervised feature selection with structured graph optimization}. In \bibinfo{booktitle}{\emph{Proceedings of the AAAI Conference on Artificial Intelligence}}, Vol.~\bibinfo{volume}{30}.
\newblock


\bibitem[Omanovic et~al\mbox{.}(2023)]%
        {tri10156842}
\bibfield{author}{\bibinfo{person}{Amra Omanovic}, \bibinfo{person}{Polona Oblak}, {and} \bibinfo{person}{Tomaz Curk}.} \bibinfo{year}{2023}\natexlab{}.
\newblock \showarticletitle{Matrix Tri-Factorization Over the Tropical Semiring}.
\newblock \bibinfo{journal}{\emph{IEEE Access}}  \bibinfo{volume}{11} (\bibinfo{year}{2023}), \bibinfo{pages}{69022--69032}.
\newblock


\bibitem[Paatero and Tapper(1994)]%
        {paatero1994positive}
\bibfield{author}{\bibinfo{person}{Pentti Paatero} {and} \bibinfo{person}{Unto Tapper}.} \bibinfo{year}{1994}\natexlab{}.
\newblock \showarticletitle{Positive matrix factorization: A non-negative factor model with optimal utilization of error estimates of data values}.
\newblock \bibinfo{journal}{\emph{Environmetrics}} \bibinfo{volume}{5}, \bibinfo{number}{2} (\bibinfo{year}{1994}), \bibinfo{pages}{111--126}.
\newblock


\bibitem[Peharz and Pernkopf(2012)]%
        {peharz2012sparse}
\bibfield{author}{\bibinfo{person}{Robert Peharz} {and} \bibinfo{person}{Franz Pernkopf}.} \bibinfo{year}{2012}\natexlab{}.
\newblock \showarticletitle{Sparse nonnegative matrix factorization with $\ell_0$-constraints}.
\newblock \bibinfo{journal}{\emph{Neurocomputing}}  \bibinfo{volume}{80} (\bibinfo{year}{2012}), \bibinfo{pages}{38--46}.
\newblock


\bibitem[Pei et~al\mbox{.}(2018)]%
        {Pei7748469}
\bibfield{author}{\bibinfo{person}{Xiaobing Pei}, \bibinfo{person}{Chuanbo Chen}, {and} \bibinfo{person}{Weihua Gong}.} \bibinfo{year}{2018}\natexlab{}.
\newblock \showarticletitle{Concept Factorization With Adaptive Neighbors for Document Clustering}.
\newblock \bibinfo{journal}{\emph{IEEE Transactions on Neural Networks and Learning Systems}} \bibinfo{volume}{29}, \bibinfo{number}{2} (\bibinfo{year}{2018}), \bibinfo{pages}{343--352}.
\newblock


\bibitem[Peng et~al\mbox{.}(2022c)]%
        {peng2022logbased}
\bibfield{author}{\bibinfo{person}{C. Peng}, \bibinfo{person}{Y. Zhang}, \bibinfo{person}{Y. Chen}, \bibinfo{person}{Z. Kang}, \bibinfo{person}{C. Chen}, {and} \bibinfo{person}{Q. Cheng}.} \bibinfo{year}{2022}\natexlab{c}.
\newblock \showarticletitle{Log-based sparse nonnegative matrix factorization for data representation}.
\newblock \bibinfo{journal}{\emph{Knowledge-Based Systems}}  \bibinfo{volume}{251} (\bibinfo{year}{2022}), \bibinfo{pages}{109127}.
\newblock


\bibitem[Peng et~al\mbox{.}(2022b)]%
        {PENG2022106}
\bibfield{author}{\bibinfo{person}{Chong Peng}, \bibinfo{person}{Zhilu Zhang}, \bibinfo{person}{Chenglizhao Chen}, \bibinfo{person}{Zhao Kang}, {and} \bibinfo{person}{Qiang Cheng}.} \bibinfo{year}{2022}\natexlab{b}.
\newblock \showarticletitle{Two-dimensional semi-nonnegative matrix factorization for clustering}.
\newblock \bibinfo{journal}{\emph{Information Sciences}}  \bibinfo{volume}{590} (\bibinfo{year}{2022}), \bibinfo{pages}{106--141}.
\newblock


\bibitem[Peng et~al\mbox{.}(2020)]%
        {peng2020robust}
\bibfield{author}{\bibinfo{person}{Siyuan Peng}, \bibinfo{person}{Wee Ser}, \bibinfo{person}{Badong Chen}, {and} \bibinfo{person}{Zhiping Lin}.} \bibinfo{year}{2020}\natexlab{}.
\newblock \showarticletitle{Robust orthogonal nonnegative matrix tri-factorization for data representation}.
\newblock \bibinfo{journal}{\emph{Knowledge-Based Systems}}  \bibinfo{volume}{201} (\bibinfo{year}{2020}), \bibinfo{pages}{106054}.
\newblock


\bibitem[Peng et~al\mbox{.}(2022a)]%
        {PENG2022571}
\bibfield{author}{\bibinfo{person}{Siyuan Peng}, \bibinfo{person}{Zhijing Yang}, \bibinfo{person}{Bingo Wing-Kuen Ling}, \bibinfo{person}{Badong Chen}, {and} \bibinfo{person}{Zhiping Lin}.} \bibinfo{year}{2022}\natexlab{a}.
\newblock \showarticletitle{Dual semi-supervised convex nonnegative matrix factorization for data representation}.
\newblock \bibinfo{journal}{\emph{Information Sciences}}  \bibinfo{volume}{585} (\bibinfo{year}{2022}), \bibinfo{pages}{571--593}.
\newblock


\bibitem[Qi et~al\mbox{.}(2018)]%
        {qi2018unsupervised}
\bibfield{author}{\bibinfo{person}{Miao Qi}, \bibinfo{person}{Ting Wang}, \bibinfo{person}{Fucong Liu}, \bibinfo{person}{Baoxue Zhang}, \bibinfo{person}{Jianzhong Wang}, {and} \bibinfo{person}{Yugen Yi}.} \bibinfo{year}{2018}\natexlab{}.
\newblock \showarticletitle{Unsupervised feature selection by regularized matrix factorization}.
\newblock \bibinfo{journal}{\emph{Neurocomputing}}  \bibinfo{volume}{273} (\bibinfo{year}{2018}), \bibinfo{pages}{593--610}.
\newblock


\bibitem[Qin et~al\mbox{.}(2023)]%
        {9543530}
\bibfield{author}{\bibinfo{person}{Yalan Qin}, \bibinfo{person}{Guorui Feng}, \bibinfo{person}{Yanli Ren}, {and} \bibinfo{person}{Xinpeng Zhang}.} \bibinfo{year}{2023}\natexlab{}.
\newblock \showarticletitle{Block-Diagonal Guided Symmetric Nonnegative Matrix Factorization}.
\newblock \bibinfo{journal}{\emph{IEEE Transactions on Knowledge and Data Engineering}} \bibinfo{volume}{35}, \bibinfo{number}{3} (\bibinfo{year}{2023}), \bibinfo{pages}{2313--2325}.
\newblock


\bibitem[Qiu et~al\mbox{.}(2022)]%
        {9058984gene}
\bibfield{author}{\bibinfo{person}{Yuning Qiu}, \bibinfo{person}{Guoxu Zhou}, \bibinfo{person}{Yanjiao Wang}, \bibinfo{person}{Yu Zhang}, {and} \bibinfo{person}{Shengli Xie}.} \bibinfo{year}{2022}\natexlab{}.
\newblock \showarticletitle{A Generalized Graph Regularized Non-Negative {T}ucker Decomposition Framework for Tensor Data Representation}.
\newblock \bibinfo{journal}{\emph{IEEE Transactions on Cybernetics}} \bibinfo{volume}{52}, \bibinfo{number}{1} (\bibinfo{year}{2022}), \bibinfo{pages}{594--607}.
\newblock


\bibitem[Qu et~al\mbox{.}(2023)]%
        {hyper10124031}
\bibfield{author}{\bibinfo{person}{Kewen Qu}, \bibinfo{person}{Zhenqing Li}, \bibinfo{person}{Chenyang Wang}, \bibinfo{person}{Fangzhou Luo}, {and} \bibinfo{person}{Wenxing Bao}.} \bibinfo{year}{2023}\natexlab{}.
\newblock \showarticletitle{Hyperspectral Unmixing Using Higher-Order Graph Regularized {NMF} With Adaptive Feature Selection}.
\newblock \bibinfo{journal}{\emph{IEEE Transactions on Geoscience and Remote Sensing}}  \bibinfo{volume}{61} (\bibinfo{year}{2023}), \bibinfo{pages}{1--15}.
\newblock


\bibitem[Ran et~al\mbox{.}(2020)]%
        {ran2020general}
\bibfield{author}{\bibinfo{person}{Ruisheng Ran}, \bibinfo{person}{Ji Feng}, \bibinfo{person}{Shougui Zhang}, {and} \bibinfo{person}{Bin Fang}.} \bibinfo{year}{2020}\natexlab{}.
\newblock \showarticletitle{A general matrix function dimensionality reduction framework and extension for manifold learning}.
\newblock \bibinfo{journal}{\emph{IEEE Transactions on Cybernetics}} \bibinfo{volume}{52}, \bibinfo{number}{4} (\bibinfo{year}{2020}), \bibinfo{pages}{2137--2148}.
\newblock


\bibitem[Saberi-Movahed et~al\mbox{.}(2020)]%
        {saberi2020supervised}
\bibfield{author}{\bibinfo{person}{Farid Saberi-Movahed}, \bibinfo{person}{Mahdi Eftekhari}, {and} \bibinfo{person}{Mohammad Mohtashami}.} \bibinfo{year}{2020}\natexlab{}.
\newblock \showarticletitle{Supervised feature selection by constituting a basis for the original space of features and matrix factorization}.
\newblock \bibinfo{journal}{\emph{International Journal of Machine Learning and Cybernetics}}  \bibinfo{volume}{11} (\bibinfo{year}{2020}), \bibinfo{pages}{1405--1421}.
\newblock


\bibitem[Saberi-Movahed et~al\mbox{.}(2022)]%
        {saberi2022dual}
\bibfield{author}{\bibinfo{person}{Farid Saberi-Movahed}, \bibinfo{person}{Mehrdad Rostami}, \bibinfo{person}{Kamal Berahmand}, \bibinfo{person}{Saeed Karami}, \bibinfo{person}{Prayag Tiwari}, \bibinfo{person}{Mourad Oussalah}, {and} \bibinfo{person}{Shahab~S Band}.} \bibinfo{year}{2022}\natexlab{}.
\newblock \showarticletitle{Dual regularized unsupervised feature selection based on matrix factorization and minimum redundancy with application in gene selection}.
\newblock \bibinfo{journal}{\emph{Knowledge-Based Systems}}  \bibinfo{volume}{256} (\bibinfo{year}{2022}), \bibinfo{pages}{109884}.
\newblock


\bibitem[Saeys et~al\mbox{.}(2007)]%
        {saeys2007review}
\bibfield{author}{\bibinfo{person}{Yvan Saeys}, \bibinfo{person}{Inaki Inza}, {and} \bibinfo{person}{Pedro Larranaga}.} \bibinfo{year}{2007}\natexlab{}.
\newblock \showarticletitle{A review of feature selection techniques in bioinformatics}.
\newblock \bibinfo{journal}{\emph{bioinformatics}} \bibinfo{volume}{23}, \bibinfo{number}{19} (\bibinfo{year}{2007}), \bibinfo{pages}{2507--2517}.
\newblock


\bibitem[Shang et~al\mbox{.}(2012)]%
        {shang2012graph}
\bibfield{author}{\bibinfo{person}{Fanhua Shang}, \bibinfo{person}{LC Jiao}, {and} \bibinfo{person}{Fei Wang}.} \bibinfo{year}{2012}\natexlab{}.
\newblock \showarticletitle{Graph dual regularization non-negative matrix factorization for co-clustering}.
\newblock \bibinfo{journal}{\emph{Pattern Recognition}} \bibinfo{volume}{45}, \bibinfo{number}{6} (\bibinfo{year}{2012}), \bibinfo{pages}{2237--2250}.
\newblock


\bibitem[Shang et~al\mbox{.}(2020a)]%
        {shang2020double}
\bibfield{author}{\bibinfo{person}{Ronghua Shang}, \bibinfo{person}{Jiuzheng Song}, \bibinfo{person}{Licheng Jiao}, {and} \bibinfo{person}{Yangyang Li}.} \bibinfo{year}{2020}\natexlab{a}.
\newblock \showarticletitle{Double feature selection algorithm based on low-rank sparse non-negative matrix factorization}.
\newblock \bibinfo{journal}{\emph{International Journal of Machine Learning and Cybernetics}}  \bibinfo{volume}{11} (\bibinfo{year}{2020}), \bibinfo{pages}{1891--1908}.
\newblock


\bibitem[Shang et~al\mbox{.}(2016)]%
        {shang2016subspace}
\bibfield{author}{\bibinfo{person}{Ronghua Shang}, \bibinfo{person}{Wenbing Wang}, \bibinfo{person}{Rustam Stolkin}, {and} \bibinfo{person}{Licheng Jiao}.} \bibinfo{year}{2016}\natexlab{}.
\newblock \showarticletitle{Subspace learning-based graph regularized feature selection}.
\newblock \bibinfo{journal}{\emph{Knowledge-Based Systems}}  \bibinfo{volume}{112} (\bibinfo{year}{2016}), \bibinfo{pages}{152--165}.
\newblock


\bibitem[Shang et~al\mbox{.}(2017)]%
        {shang2017non}
\bibfield{author}{\bibinfo{person}{Ronghua Shang}, \bibinfo{person}{Wenbing Wang}, \bibinfo{person}{Rustam Stolkin}, {and} \bibinfo{person}{Licheng Jiao}.} \bibinfo{year}{2017}\natexlab{}.
\newblock \showarticletitle{Non-negative spectral learning and sparse regression-based dual-graph regularized feature selection}.
\newblock \bibinfo{journal}{\emph{IEEE Transactions on Cybernetics}} \bibinfo{volume}{48}, \bibinfo{number}{2} (\bibinfo{year}{2017}), \bibinfo{pages}{793--806}.
\newblock


\bibitem[Shang et~al\mbox{.}(2020b)]%
        {shang2020subspace}
\bibfield{author}{\bibinfo{person}{Ronghua Shang}, \bibinfo{person}{Kaiming Xu}, {and} \bibinfo{person}{Licheng Jiao}.} \bibinfo{year}{2020}\natexlab{b}.
\newblock \showarticletitle{Subspace learning for unsupervised feature selection via adaptive structure learning and rank approximation}.
\newblock \bibinfo{journal}{\emph{Neurocomputing}}  \bibinfo{volume}{413} (\bibinfo{year}{2020}), \bibinfo{pages}{72--84}.
\newblock


\bibitem[Shang et~al\mbox{.}(2020c)]%
        {shang2020sparse}
\bibfield{author}{\bibinfo{person}{Ronghua Shang}, \bibinfo{person}{Kaiming Xu}, \bibinfo{person}{Fanhua Shang}, {and} \bibinfo{person}{Licheng Jiao}.} \bibinfo{year}{2020}\natexlab{c}.
\newblock \showarticletitle{Sparse and low-redundant subspace learning-based dual-graph regularized robust feature selection}.
\newblock \bibinfo{journal}{\emph{Knowledge-Based Systems}}  \bibinfo{volume}{187} (\bibinfo{year}{2020}), \bibinfo{pages}{104830}.
\newblock


\bibitem[Sheikhpour(2023)]%
        {sheikhpour2023local}
\bibfield{author}{\bibinfo{person}{Razieh Sheikhpour}.} \bibinfo{year}{2023}\natexlab{}.
\newblock \showarticletitle{A local spline regression-based framework for semi-supervised sparse feature selection}.
\newblock \bibinfo{journal}{\emph{Knowledge-Based Systems}}  \bibinfo{volume}{262} (\bibinfo{year}{2023}), \bibinfo{pages}{110265}.
\newblock


\bibitem[Sheikhpour et~al\mbox{.}(2023)]%
        {sheikhpour2023hessian}
\bibfield{author}{\bibinfo{person}{Razieh Sheikhpour}, \bibinfo{person}{Kamal Berahmand}, {and} \bibinfo{person}{Saman Forouzandeh}.} \bibinfo{year}{2023}\natexlab{}.
\newblock \showarticletitle{Hessian-based semi-supervised feature selection using generalized uncorrelated constraint}.
\newblock \bibinfo{journal}{\emph{Knowledge-Based Systems}}  \bibinfo{volume}{269} (\bibinfo{year}{2023}), \bibinfo{pages}{110521}.
\newblock


\bibitem[Sheikhpour et~al\mbox{.}(2020)]%
        {sheikhpour2020robust}
\bibfield{author}{\bibinfo{person}{Razieh Sheikhpour}, \bibinfo{person}{Mehdi~Agha Sarram}, \bibinfo{person}{Sajjad Gharaghani}, {and} \bibinfo{person}{Mohammad Ali~Zare Chahooki}.} \bibinfo{year}{2020}\natexlab{}.
\newblock \showarticletitle{A robust graph-based semi-supervised sparse feature selection method}.
\newblock \bibinfo{journal}{\emph{Information Sciences}}  \bibinfo{volume}{531} (\bibinfo{year}{2020}), \bibinfo{pages}{13--30}.
\newblock


\bibitem[Sheng et~al\mbox{.}(2021)]%
        {sheng2021dual}
\bibfield{author}{\bibinfo{person}{Chao Sheng}, \bibinfo{person}{Peng Song}, \bibinfo{person}{Weijian Zhang}, {and} \bibinfo{person}{Dongliang Chen}.} \bibinfo{year}{2021}\natexlab{}.
\newblock \showarticletitle{Dual-graph regularized subspace learning based feature selection}.
\newblock \bibinfo{journal}{\emph{Digital Signal Processing}}  \bibinfo{volume}{117} (\bibinfo{year}{2021}), \bibinfo{pages}{103175}.
\newblock


\bibitem[Shu et~al\mbox{.}(2022)]%
        {shu2022adaptive}
\bibfield{author}{\bibinfo{person}{Z. Shu}, \bibinfo{person}{Y. Sun}, \bibinfo{person}{J. Tang}, {and} \bibinfo{person}{C. You}.} \bibinfo{year}{2022}\natexlab{}.
\newblock \showarticletitle{Adaptive graph regularized deep semi-nonnegative matrix factorization for data representation}.
\newblock \bibinfo{journal}{\emph{Neural Processing Letters}} \bibinfo{volume}{54}, \bibinfo{number}{6} (\bibinfo{year}{2022}), \bibinfo{pages}{5721--5739}.
\newblock


\bibitem[Sugiyama(2006)]%
        {sugiyama2006local}
\bibfield{author}{\bibinfo{person}{Masashi Sugiyama}.} \bibinfo{year}{2006}\natexlab{}.
\newblock \showarticletitle{Local fisher discriminant analysis for supervised dimensionality reduction}. In \bibinfo{booktitle}{\emph{Proceedings of the 23rd international conference on Machine learning}}. \bibinfo{pages}{905--912}.
\newblock


\bibitem[Sun et~al\mbox{.}(2022)]%
        {sun2022deep}
\bibfield{author}{\bibinfo{person}{Jianyong Sun}, \bibinfo{person}{Qingming Kong}, {and} \bibinfo{person}{Zongben Xu}.} \bibinfo{year}{2022}\natexlab{}.
\newblock \showarticletitle{Deep alternating non-negative matrix factorisation}.
\newblock \bibinfo{journal}{\emph{Knowledge-Based Systems}}  \bibinfo{volume}{251} (\bibinfo{year}{2022}), \bibinfo{pages}{109210}.
\newblock


\bibitem[Sun et~al\mbox{.}(2015)]%
        {sun2015heterogeneous}
\bibfield{author}{\bibinfo{person}{Yanfeng Sun} {et~al\mbox{.}}} \bibinfo{year}{2015}\natexlab{}.
\newblock \showarticletitle{Heterogeneous tensor decomposition for clustering via manifold optimization}.
\newblock \bibinfo{journal}{\emph{IEEE Transactions on Pattern Analysis and Machine Intelligence}} \bibinfo{volume}{38}, \bibinfo{number}{3} (\bibinfo{year}{2015}), \bibinfo{pages}{476--489}.
\newblock


\bibitem[Tang and Feng(2022)]%
        {tang2022robust}
\bibfield{author}{\bibinfo{person}{J. Tang} {and} \bibinfo{person}{H. Feng}.} \bibinfo{year}{2022}\natexlab{}.
\newblock \showarticletitle{Robust local-coordinate non-negative matrix factorization with adaptive graph for robust clustering}.
\newblock \bibinfo{journal}{\emph{Information Sciences}}  \bibinfo{volume}{610} (\bibinfo{year}{2022}), \bibinfo{pages}{1058--1077}.
\newblock


\bibitem[Tang and Wan(2021)]%
        {tang2021orthogonal}
\bibfield{author}{\bibinfo{person}{J. Tang} {and} \bibinfo{person}{Z. Wan}.} \bibinfo{year}{2021}\natexlab{}.
\newblock \showarticletitle{Orthogonal Dual Graph-Regularized Nonnegative Matrix Factorization for Co-clustering}.
\newblock \bibinfo{journal}{\emph{Journal of Scientific Computing}} \bibinfo{volume}{87}, \bibinfo{number}{3} (\bibinfo{year}{2021}), \bibinfo{pages}{66}.
\newblock


\bibitem[Tolic et~al\mbox{.}(2018)]%
        {tolic2018nonlinear}
\bibfield{author}{\bibinfo{person}{D. Tolic}, \bibinfo{person}{N. Antulov-Fantulin}, {and} \bibinfo{person}{I. Kopriva}.} \bibinfo{year}{2018}\natexlab{}.
\newblock \showarticletitle{A Nonlinear Orthogonal Non-negative Matrix Factorization Approach to Subspace Clustering}.
\newblock \bibinfo{journal}{\emph{Pattern Recognition}}  \bibinfo{volume}{82} (\bibinfo{year}{2018}), \bibinfo{pages}{40--55}.
\newblock


\bibitem[Trigeorgis et~al\mbox{.}(2017)]%
        {trigeorgis2016deep}
\bibfield{author}{\bibinfo{person}{George Trigeorgis}, \bibinfo{person}{Konstantinos Bousmalis}, \bibinfo{person}{Stefanos Zafeiriou}, {and} \bibinfo{person}{Björn~W. Schuller}.} \bibinfo{year}{2017}\natexlab{}.
\newblock \showarticletitle{A deep matrix factorization method for learning attribute representations}.
\newblock \bibinfo{journal}{\emph{IEEE Transactions on Pattern Analysis and Machine Intelligence}} \bibinfo{volume}{39}, \bibinfo{number}{3} (\bibinfo{year}{2017}), \bibinfo{pages}{417--429}.
\newblock


\bibitem[Wang et~al\mbox{.}(2021c)]%
        {wang2021semisupervised}
\bibfield{author}{\bibinfo{person}{Chen Wang}, \bibinfo{person}{Xiaojun Chen}, \bibinfo{person}{Guowen Yuan}, \bibinfo{person}{Feiping Nie}, {and} \bibinfo{person}{Min Yang}.} \bibinfo{year}{2021}\natexlab{c}.
\newblock \showarticletitle{Semisupervised feature selection with sparse discriminative least squares regression}.
\newblock \bibinfo{journal}{\emph{IEEE Transactions on Cybernetics}} \bibinfo{volume}{52}, \bibinfo{number}{8} (\bibinfo{year}{2021}), \bibinfo{pages}{8413--8424}.
\newblock


\bibitem[Wang et~al\mbox{.}(2021a)]%
        {wang2021sparse}
\bibfield{author}{\bibinfo{person}{D. Wang}, \bibinfo{person}{Z. Chang}, {and} \bibinfo{person}{F. Cong}.} \bibinfo{year}{2021}\natexlab{a}.
\newblock \showarticletitle{Sparse nonnegative tensor decomposition using proximal algorithm and inexact block coordinate descent scheme}.
\newblock \bibinfo{journal}{\emph{Neural Computing and Applications}} \bibinfo{volume}{33}, \bibinfo{number}{24} (\bibinfo{year}{2021}), \bibinfo{pages}{17369--17387}.
\newblock


\bibitem[Wang et~al\mbox{.}(2022b)]%
        {wang2022dual}
\bibfield{author}{\bibinfo{person}{D. Wang}, \bibinfo{person}{T. Li}, \bibinfo{person}{P. Deng}, \bibinfo{person}{H. Wang}, {and} \bibinfo{person}{P. Zhang}.} \bibinfo{year}{2022}\natexlab{b}.
\newblock \showarticletitle{Dual graph-regularized sparse concept factorization for clustering}.
\newblock \bibinfo{journal}{\emph{Information Sciences}}  \bibinfo{volume}{607} (\bibinfo{year}{2022}), \bibinfo{pages}{1074--1088}.
\newblock


\bibitem[Wang et~al\mbox{.}(2023)]%
        {WANG2023101884}
\bibfield{author}{\bibinfo{person}{Dexian Wang}, \bibinfo{person}{Tianrui Li}, \bibinfo{person}{Wei Huang}, \bibinfo{person}{Zhipeng Luo}, \bibinfo{person}{Ping Deng}, \bibinfo{person}{Pengfei Zhang}, {and} \bibinfo{person}{Minbo Ma}.} \bibinfo{year}{2023}\natexlab{}.
\newblock \showarticletitle{A multi-view clustering algorithm based on deep semi-{NMF}}.
\newblock \bibinfo{journal}{\emph{Information Fusion}}  \bibinfo{volume}{99} (\bibinfo{year}{2023}), \bibinfo{pages}{101884}.
\newblock


\bibitem[Wang et~al\mbox{.}(2021d)]%
        {wang2021joint}
\bibfield{author}{\bibinfo{person}{Jingyu Wang}, \bibinfo{person}{Lin Wang}, \bibinfo{person}{Feiping Nie}, {and} \bibinfo{person}{Xuelong Li}.} \bibinfo{year}{2021}\natexlab{d}.
\newblock \showarticletitle{Joint feature selection and extraction with sparse unsupervised projection}.
\newblock \bibinfo{journal}{\emph{IEEE Transactions on Neural Networks and Learning Systems}} (\bibinfo{year}{2021}).
\newblock


\bibitem[Wang et~al\mbox{.}(2021e)]%
        {wang2021novel}
\bibfield{author}{\bibinfo{person}{Jingyu Wang}, \bibinfo{person}{Lin Wang}, \bibinfo{person}{Feiping Nie}, {and} \bibinfo{person}{Xuelong Li}.} \bibinfo{year}{2021}\natexlab{e}.
\newblock \showarticletitle{A novel formulation of trace ratio linear discriminant analysis}.
\newblock \bibinfo{journal}{\emph{IEEE Transactions on Neural Networks and Learning Systems}} \bibinfo{volume}{33}, \bibinfo{number}{10} (\bibinfo{year}{2021}), \bibinfo{pages}{5568--5578}.
\newblock


\bibitem[Wang et~al\mbox{.}(2021f)]%
        {wang2021unsupervised}
\bibfield{author}{\bibinfo{person}{Jingyu Wang}, \bibinfo{person}{Fangyuan Xie}, \bibinfo{person}{Feiping Nie}, {and} \bibinfo{person}{Xuelong Li}.} \bibinfo{year}{2021}\natexlab{f}.
\newblock \showarticletitle{Unsupervised adaptive embedding for dimensionality reduction}.
\newblock \bibinfo{journal}{\emph{IEEE Transactions on Neural Networks and Learning Systems}} \bibinfo{volume}{33}, \bibinfo{number}{11} (\bibinfo{year}{2021}), \bibinfo{pages}{6844--6855}.
\newblock


\bibitem[Wang et~al\mbox{.}(2015a)]%
        {wang2015feature}
\bibfield{author}{\bibinfo{person}{Jim Jing-Yan Wang}, \bibinfo{person}{Jianhua~Z Huang}, \bibinfo{person}{Yijun Sun}, {and} \bibinfo{person}{Xin Gao}.} \bibinfo{year}{2015}\natexlab{a}.
\newblock \showarticletitle{Feature selection and multi-kernel learning for adaptive graph regularized nonnegative matrix factorization}.
\newblock \bibinfo{journal}{\emph{Expert Systems with Applications}} \bibinfo{volume}{42}, \bibinfo{number}{3} (\bibinfo{year}{2015}), \bibinfo{pages}{1278--1286}.
\newblock


\bibitem[Wang et~al\mbox{.}(2022a)]%
        {9134812}
\bibfield{author}{\bibinfo{person}{Qi Wang}, \bibinfo{person}{Xiang He}, \bibinfo{person}{Xu Jiang}, {and} \bibinfo{person}{Xuelong Li}.} \bibinfo{year}{2022}\natexlab{a}.
\newblock \showarticletitle{Robust Bi-Stochastic Graph Regularized Matrix Factorization for Data Clustering}.
\newblock \bibinfo{journal}{\emph{IEEE Transactions on Pattern Analysis and Machine Intelligence}} \bibinfo{volume}{44}, \bibinfo{number}{1} (\bibinfo{year}{2022}), \bibinfo{pages}{390--403}.
\newblock


\bibitem[Wang et~al\mbox{.}(2021b)]%
        {wang2021clustering}
\bibfield{author}{\bibinfo{person}{Shuai Wang}, \bibinfo{person}{Tsung{-}Han Chang}, \bibinfo{person}{Yue Cui}, {and} \bibinfo{person}{Jun{-}Seng Pang}.} \bibinfo{year}{2021}\natexlab{b}.
\newblock \showarticletitle{Clustering by Orthogonal NMF Model and Non-Convex Penalty Optimization}.
\newblock \bibinfo{journal}{\emph{IEEE Transactions on Signal Processing}}  \bibinfo{volume}{69} (\bibinfo{year}{2021}), \bibinfo{pages}{5273--5288}.
\newblock


\bibitem[Wang et~al\mbox{.}(2020)]%
        {wang2020structured}
\bibfield{author}{\bibinfo{person}{Shiping Wang}, \bibinfo{person}{Jiawei Chen}, \bibinfo{person}{Wenzhong Guo}, {and} \bibinfo{person}{Genggeng Liu}.} \bibinfo{year}{2020}\natexlab{}.
\newblock \showarticletitle{Structured learning for unsupervised feature selection with high-order matrix factorization}.
\newblock \bibinfo{journal}{\emph{Expert Systems with Applications}}  \bibinfo{volume}{140} (\bibinfo{year}{2020}), \bibinfo{pages}{112878}.
\newblock


\bibitem[Wang and Guo(2017)]%
        {wang2017robust}
\bibfield{author}{\bibinfo{person}{S. Wang} {and} \bibinfo{person}{W. Guo}.} \bibinfo{year}{2017}\natexlab{}.
\newblock \showarticletitle{Robust Co-clustering via Dual Local Learning and High-order Matrix Factorization}.
\newblock \bibinfo{journal}{\emph{Knowledge-Based Systems}}  \bibinfo{volume}{138} (\bibinfo{year}{2017}), \bibinfo{pages}{176--187}.
\newblock


\bibitem[Wang and Huang(2017)]%
        {wang2017penalized}
\bibfield{author}{\bibinfo{person}{Shuai Wang} {and} \bibinfo{person}{Ailong Huang}.} \bibinfo{year}{2017}\natexlab{}.
\newblock \showarticletitle{Penalized nonnegative matrix tri-factorization for co-clustering}.
\newblock \bibinfo{journal}{\emph{Expert Systems with Applications}}  \bibinfo{volume}{78} (\bibinfo{year}{2017}), \bibinfo{pages}{64--73}.
\newblock


\bibitem[Wang et~al\mbox{.}(2022c)]%
        {wang2022robust}
\bibfield{author}{\bibinfo{person}{Sisi Wang}, \bibinfo{person}{Feiping Nie}, \bibinfo{person}{Zheng Wang}, \bibinfo{person}{Rong Wang}, {and} \bibinfo{person}{Xuelong Li}.} \bibinfo{year}{2022}\natexlab{c}.
\newblock \showarticletitle{Robust Principal Component Analysis via Joint Reconstruction and Projection}.
\newblock \bibinfo{journal}{\emph{IEEE Transactions on Neural Networks and Learning Systems}} (\bibinfo{year}{2022}).
\newblock


\bibitem[Wang et~al\mbox{.}(2015b)]%
        {wang2015subspace}
\bibfield{author}{\bibinfo{person}{Shiping Wang}, \bibinfo{person}{Witold Pedrycz}, \bibinfo{person}{Qingxin Zhu}, {and} \bibinfo{person}{William Zhu}.} \bibinfo{year}{2015}\natexlab{b}.
\newblock \showarticletitle{Subspace learning for unsupervised feature selection via matrix factorization}.
\newblock \bibinfo{journal}{\emph{Pattern Recognition}} \bibinfo{volume}{48}, \bibinfo{number}{1} (\bibinfo{year}{2015}), \bibinfo{pages}{10--19}.
\newblock


\bibitem[Wang et~al\mbox{.}(2015c)]%
        {wang2015unsupervised}
\bibfield{author}{\bibinfo{person}{Shiping Wang}, \bibinfo{person}{Witold Pedrycz}, \bibinfo{person}{Qingxin Zhu}, {and} \bibinfo{person}{William Zhu}.} \bibinfo{year}{2015}\natexlab{c}.
\newblock \showarticletitle{Unsupervised feature selection via maximum projection and minimum redundancy}.
\newblock \bibinfo{journal}{\emph{Knowledge-Based Systems}}  \bibinfo{volume}{75} (\bibinfo{year}{2015}), \bibinfo{pages}{19--29}.
\newblock


\bibitem[Xing et~al\mbox{.}(2021)]%
        {xing2021discriminative}
\bibfield{author}{\bibinfo{person}{Z. Xing}, \bibinfo{person}{M. Wen}, \bibinfo{person}{J. Peng}, {and} \bibinfo{person}{J. Feng}.} \bibinfo{year}{2021}\natexlab{}.
\newblock \showarticletitle{Discriminative semi-supervised non-negative matrix factorization for data clustering}.
\newblock \bibinfo{journal}{\emph{Engineering Applications of Artificial Intelligence}}  \bibinfo{volume}{103} (\bibinfo{year}{2021}), \bibinfo{pages}{104289}.
\newblock


\bibitem[Xu and Gong(2004)]%
        {xu2004document}
\bibfield{author}{\bibinfo{person}{Wei Xu} {and} \bibinfo{person}{Yihong Gong}.} \bibinfo{year}{2004}\natexlab{}.
\newblock \showarticletitle{Document clustering by concept factorization}. In \bibinfo{booktitle}{\emph{Proceedings of the 27th annual international ACM SIGIR conference on Research and development in information retrieval}}. \bibinfo{pages}{202--209}.
\newblock


\bibitem[Xu et~al\mbox{.}(2023a)]%
        {xu2023graph}
\bibfield{author}{\bibinfo{person}{Weihua Xu}, \bibinfo{person}{Man Huang}, \bibinfo{person}{Zongying Jiang}, {and} \bibinfo{person}{Yuhua Qian}.} \bibinfo{year}{2023}\natexlab{a}.
\newblock \showarticletitle{Graph-Based Unsupervised Feature Selection for Interval-Valued Information System}.
\newblock \bibinfo{journal}{\emph{IEEE Transactions on Neural Networks and Learning Systems}} (\bibinfo{year}{2023}).
\newblock


\bibitem[Xu et~al\mbox{.}(2019)]%
        {xu2019review}
\bibfield{author}{\bibinfo{person}{Xinzheng Xu}, \bibinfo{person}{Tianming Liang}, \bibinfo{person}{Jiong Zhu}, \bibinfo{person}{Dong Zheng}, {and} \bibinfo{person}{Tongfeng Sun}.} \bibinfo{year}{2019}\natexlab{}.
\newblock \showarticletitle{Review of classical dimensionality reduction and sample selection methods for large-scale data processing}.
\newblock \bibinfo{journal}{\emph{Neurocomputing}}  \bibinfo{volume}{328} (\bibinfo{year}{2019}), \bibinfo{pages}{5--15}.
\newblock


\bibitem[Xu et~al\mbox{.}(2021)]%
        {xu2021general}
\bibfield{author}{\bibinfo{person}{Xueyuan Xu}, \bibinfo{person}{Xia Wu}, \bibinfo{person}{Fulin Wei}, \bibinfo{person}{Wei Zhong}, {and} \bibinfo{person}{Feiping Nie}.} \bibinfo{year}{2021}\natexlab{}.
\newblock \showarticletitle{A general framework for feature selection under orthogonal regression with global redundancy minimization}.
\newblock \bibinfo{journal}{\emph{IEEE Transactions on Knowledge and Data Engineering}} \bibinfo{volume}{34}, \bibinfo{number}{11} (\bibinfo{year}{2021}), \bibinfo{pages}{5056--5069}.
\newblock


\bibitem[Xu et~al\mbox{.}(2023b)]%
        {xu2023hypergraph}
\bibfield{author}{\bibinfo{person}{Y. Xu}, \bibinfo{person}{L. Lu}, \bibinfo{person}{Q. Liu}, {and} \bibinfo{person}{Z. Chen}.} \bibinfo{year}{2023}\natexlab{b}.
\newblock \showarticletitle{Hypergraph-Regularized $L_p$ Smooth Nonnegative Matrix Factorization for Data Representation}.
\newblock \bibinfo{journal}{\emph{Mathematics}} \bibinfo{volume}{11}, \bibinfo{number}{13} (\bibinfo{year}{2023}).
\newblock


\bibitem[Yang and Oja(2010)]%
        {yang2010linear}
\bibfield{author}{\bibinfo{person}{Zhenghua Yang} {and} \bibinfo{person}{Erkki Oja}.} \bibinfo{year}{2010}\natexlab{}.
\newblock \showarticletitle{Linear and nonlinear projective nonnegative matrix factorization}.
\newblock \bibinfo{journal}{\emph{IEEE Transactions on Neural Networks}} \bibinfo{volume}{21}, \bibinfo{number}{5} (\bibinfo{year}{2010}), \bibinfo{pages}{734--749}.
\newblock


\bibitem[Yang et~al\mbox{.}(2007)]%
        {yang2007projective}
\bibfield{author}{\bibinfo{person}{Zhenghua Yang}, \bibinfo{person}{Zhikang Yuan}, {and} \bibinfo{person}{Jorma Laaksonen}.} \bibinfo{year}{2007}\natexlab{}.
\newblock \showarticletitle{Projective non-negative matrix factorization with applications to facial image processing}.
\newblock \bibinfo{journal}{\emph{International Journal of Pattern Recognition and Artificial Intelligence}} \bibinfo{volume}{21}, \bibinfo{number}{8} (\bibinfo{year}{2007}), \bibinfo{pages}{1353--1362}.
\newblock


\bibitem[Ye et~al\mbox{.}(2018)]%
        {ye2018deep}
\bibfield{author}{\bibinfo{person}{F. Ye}, \bibinfo{person}{C. Chen}, {and} \bibinfo{person}{Z. Zheng}.} \bibinfo{year}{2018}\natexlab{}.
\newblock \showarticletitle{Deep autoencoder-like nonnegative matrix factorization for community detection}. In \bibinfo{booktitle}{\emph{Proceedings of the 27th ACM international conference on information and knowledge management}}. \bibinfo{pages}{1393--1402}.
\newblock


\bibitem[Yi et~al\mbox{.}(2020)]%
        {yi2020nonnegative}
\bibfield{author}{\bibinfo{person}{Y. Yi}, \bibinfo{person}{J. Wang}, \bibinfo{person}{W. Zhou}, \bibinfo{person}{C. Zheng}, \bibinfo{person}{J. Kong}, {and} \bibinfo{person}{S. Qiao}.} \bibinfo{year}{2020}\natexlab{}.
\newblock \showarticletitle{Non-Negative Matrix Factorization With Locality Constrained Adaptive Graph}.
\newblock \bibinfo{journal}{\emph{IEEE Transactions on Circuits and Systems for Video Technology}} \bibinfo{volume}{30}, \bibinfo{number}{2} (\bibinfo{year}{2020}), \bibinfo{pages}{427--441}.
\newblock


\bibitem[Yi et~al\mbox{.}(2018)]%
        {yi2018ordinal}
\bibfield{author}{\bibinfo{person}{Yugen Yi}, \bibinfo{person}{Wei Zhou}, \bibinfo{person}{Qinghua Liu}, \bibinfo{person}{Guoliang Luo}, \bibinfo{person}{Jianzhong Wang}, \bibinfo{person}{Yuming Fang}, {and} \bibinfo{person}{Caixia Zheng}.} \bibinfo{year}{2018}\natexlab{}.
\newblock \showarticletitle{Ordinal preserving matrix factorization for unsupervised feature selection}.
\newblock \bibinfo{journal}{\emph{Signal Processing: Image Communication}}  \bibinfo{volume}{67} (\bibinfo{year}{2018}), \bibinfo{pages}{118--131}.
\newblock


\bibitem[Yin et~al\mbox{.}(2023)]%
        {YIN2023109274}
\bibfield{author}{\bibinfo{person}{Jingxing Yin}, \bibinfo{person}{Siyuan Peng}, \bibinfo{person}{Zhijing Yang}, \bibinfo{person}{Badong Chen}, {and} \bibinfo{person}{Zhiping Lin}.} \bibinfo{year}{2023}\natexlab{}.
\newblock \showarticletitle{Hypergraph based semi-supervised symmetric nonnegative matrix factorization for image clustering}.
\newblock \bibinfo{journal}{\emph{Pattern Recognition}}  \bibinfo{volume}{137} (\bibinfo{year}{2023}), \bibinfo{pages}{109274}.
\newblock


\bibitem[Yin et~al\mbox{.}(2019)]%
        {yin2019learning}
\bibfield{author}{\bibinfo{person}{K. Yin}, \bibinfo{person}{D. Qian}, \bibinfo{person}{W.~K. Cheung}, \bibinfo{person}{B.~C. Fung}, {and} \bibinfo{person}{J. Poon}.} \bibinfo{year}{2019}\natexlab{}.
\newblock \showarticletitle{Learning phenotypes and dynamic patient representations via {RNN} regularized collective non-negative tensor factorization}. In \bibinfo{booktitle}{\emph{Proceedings of the AAAI Conference on Artificial Intelligence}}, Vol.~\bibinfo{volume}{33}. \bibinfo{pages}{1246--1253}.
\newblock


\bibitem[Yin et~al\mbox{.}(2022)]%
        {YIN2022190}
\bibfield{author}{\bibinfo{person}{Wanguang Yin}, \bibinfo{person}{Youzhi Qu}, \bibinfo{person}{Zhengming Ma}, {and} \bibinfo{person}{Quanying Liu}.} \bibinfo{year}{2022}\natexlab{}.
\newblock \showarticletitle{Hyper{NTF}: A hypergraph regularized nonnegative tensor factorization for dimensionality reduction}.
\newblock \bibinfo{journal}{\emph{Neurocomputing}}  \bibinfo{volume}{512} (\bibinfo{year}{2022}), \bibinfo{pages}{190--202}.
\newblock


\bibitem[Yu et~al\mbox{.}(2023)]%
        {heypeyyuan}
\bibfield{author}{\bibinfo{person}{Jiyang Yu}, \bibinfo{person}{Baicheng Pan}, \bibinfo{person}{Shanshan Yu}, {and} \bibinfo{person}{Man-Fai Leung}.} \bibinfo{year}{2023}\natexlab{}.
\newblock \showarticletitle{Robust capped norm dual hyper-graph regularized non-negative matrix tri-factorization}.
\newblock \bibinfo{journal}{\emph{Mathematical Biosciences and Engineering}} \bibinfo{volume}{20}, \bibinfo{number}{7} (\bibinfo{year}{2023}), \bibinfo{pages}{12486--12509}.
\newblock


\bibitem[Yu et~al\mbox{.}(2019)]%
        {yu2019robust}
\bibfield{author}{\bibinfo{person}{Na Yu}, \bibinfo{person}{Ying-Lian Gao}, \bibinfo{person}{Jin-Xing Liu}, \bibinfo{person}{Juan Wang}, {and} \bibinfo{person}{Junliang Shang}.} \bibinfo{year}{2019}\natexlab{}.
\newblock \showarticletitle{Robust hypergraph regularized non-negative matrix factorization for sample clustering and feature selection in multi-view gene expression data}.
\newblock \bibinfo{journal}{\emph{Human genomics}} \bibinfo{volume}{13}, \bibinfo{number}{1} (\bibinfo{year}{2019}), \bibinfo{pages}{1--10}.
\newblock


\bibitem[Yu et~al\mbox{.}(2021)]%
        {9130073hyper}
\bibfield{author}{\bibinfo{person}{Na Yu}, \bibinfo{person}{Ming-Juan Wu}, \bibinfo{person}{Jin-Xing Liu}, \bibinfo{person}{Chun-Hou Zheng}, {and} \bibinfo{person}{Yong Xu}.} \bibinfo{year}{2021}\natexlab{}.
\newblock \showarticletitle{Correntropy-Based Hypergraph Regularized {NMF} for Clustering and Feature Selection on Multi-Cancer Integrated Data}.
\newblock \bibinfo{journal}{\emph{IEEE Transactions on Cybernetics}} \bibinfo{volume}{51}, \bibinfo{number}{8} (\bibinfo{year}{2021}), \bibinfo{pages}{3952--3963}.
\newblock


\bibitem[Yuan et~al\mbox{.}(2022)]%
        {yuan2020convex}
\bibfield{author}{\bibinfo{person}{Aihong Yuan}, \bibinfo{person}{Mengbo You}, \bibinfo{person}{Dongjian He}, {and} \bibinfo{person}{Xuelong Li}.} \bibinfo{year}{2022}\natexlab{}.
\newblock \showarticletitle{Convex Non-Negative Matrix Factorization With Adaptive Graph for Unsupervised Feature Selection}.
\newblock \bibinfo{journal}{\emph{IEEE Transactions on Cybernetics}} \bibinfo{volume}{52}, \bibinfo{number}{6} (\bibinfo{year}{2022}), \bibinfo{pages}{5522--5534}.
\newblock


\bibitem[Yuan et~al\mbox{.}(2023)]%
        {yuan2023beta}
\bibfield{author}{\bibinfo{person}{Ruiqing Yuan}, \bibinfo{person}{Chenlei Leng}, \bibinfo{person}{Bo Li}, {and} \bibinfo{person}{Anima Basu}.} \bibinfo{year}{2023}\natexlab{}.
\newblock \showarticletitle{$\beta$-divergence NMF with biorthogonal regularization for data representation}.
\newblock \bibinfo{journal}{\emph{Engineering Applications of Artificial Intelligence}}  \bibinfo{volume}{121} (\bibinfo{year}{2023}), \bibinfo{pages}{106014}.
\newblock


\bibitem[Yuan(2009)]%
        {yuan2009advances}
\bibfield{author}{\bibinfo{person}{Z. Yuan}.} \bibinfo{year}{2009}\natexlab{}.
\newblock \emph{\bibinfo{title}{Advances in Independent Component Analysis and Nonnegative Matrix Factorization}}.
\newblock Ph.D. thesis. \bibinfo{school}{Helsinki University of Technology}.
\newblock
\newblock
\shownote{Dissertations in Information and Computer Science, Finland}.


\bibitem[Yuan and Oja(2005)]%
        {yuan2005projective}
\bibfield{author}{\bibinfo{person}{Z. Yuan} {and} \bibinfo{person}{E. Oja}.} \bibinfo{year}{2005}\natexlab{}.
\newblock \showarticletitle{Projective Nonnegative Matrix Factorization for Image Compression and Feature Extraction}. In \bibinfo{booktitle}{\emph{Image Analysis: 14th Scandinavian Conference, SCIA 2005, Joensuu, Finland, June 19-22, 2005. Proceedings}}, Vol.~\bibinfo{volume}{14}. \bibinfo{publisher}{Springer Berlin Heidelberg}, \bibinfo{address}{Joensuu, Finland}, \bibinfo{pages}{333--342}.
\newblock


\bibitem[Zare et~al\mbox{.}(2019)]%
        {zare2019supervised}
\bibfield{author}{\bibinfo{person}{Masoumeh Zare}, \bibinfo{person}{Mahdi Eftekhari}, {and} \bibinfo{person}{Gholamreza Aghamollaei}.} \bibinfo{year}{2019}\natexlab{}.
\newblock \showarticletitle{Supervised feature selection via matrix factorization based on singular value decomposition}.
\newblock \bibinfo{journal}{\emph{Chemometrics and Intelligent Laboratory Systems}}  \bibinfo{volume}{185} (\bibinfo{year}{2019}), \bibinfo{pages}{105--113}.
\newblock


\bibitem[Zeng et~al\mbox{.}(2014)]%
        {ZENG2014209}
\bibfield{author}{\bibinfo{person}{Kun Zeng}, \bibinfo{person}{Jun Yu}, \bibinfo{person}{Cuihua Li}, \bibinfo{person}{Jane You}, {and} \bibinfo{person}{Taisong Jin}.} \bibinfo{year}{2014}\natexlab{}.
\newblock \showarticletitle{Image clustering by hyper-graph regularized non-negative matrix factorization}.
\newblock \bibinfo{journal}{\emph{Neurocomputing}}  \bibinfo{volume}{138} (\bibinfo{year}{2014}), \bibinfo{pages}{209--217}.
\newblock


\bibitem[Zhang et~al\mbox{.}(2014b)]%
        {semi6819071}
\bibfield{author}{\bibinfo{person}{Hanwang Zhang}, \bibinfo{person}{Zheng-Jun Zha}, \bibinfo{person}{Yang Yang}, \bibinfo{person}{Shuicheng Yan}, {and} \bibinfo{person}{Tat-Seng Chua}.} \bibinfo{year}{2014}\natexlab{b}.
\newblock \showarticletitle{Robust (Semi) Nonnegative Graph Embedding}.
\newblock \bibinfo{journal}{\emph{IEEE Transactions on Image Processing}} \bibinfo{volume}{23}, \bibinfo{number}{7} (\bibinfo{year}{2014}), \bibinfo{pages}{2996--3012}.
\newblock


\bibitem[Zhang et~al\mbox{.}(2017)]%
        {zhang2017lowrank}
\bibfield{author}{\bibinfo{person}{Jing Zhang} {et~al\mbox{.}}} \bibinfo{year}{2017}\natexlab{}.
\newblock \showarticletitle{Low-rank regularized heterogeneous tensor decomposition for subspace clustering}.
\newblock \bibinfo{journal}{\emph{IEEE Signal Processing Letters}} \bibinfo{volume}{25}, \bibinfo{number}{3} (\bibinfo{year}{2017}), \bibinfo{pages}{333--337}.
\newblock


\bibitem[Zhang et~al\mbox{.}(2020a)]%
        {zhang2020unsupervisedfeature}
\bibfield{author}{\bibinfo{person}{Rui Zhang}, \bibinfo{person}{Yunxing Zhang}, {and} \bibinfo{person}{Xuelong Li}.} \bibinfo{year}{2020}\natexlab{a}.
\newblock \showarticletitle{Unsupervised feature selection via adaptive graph learning and constraint}.
\newblock \bibinfo{journal}{\emph{IEEE Transactions on Neural Networks and Learning Systems}} \bibinfo{volume}{33}, \bibinfo{number}{3} (\bibinfo{year}{2020}), \bibinfo{pages}{1355--1362}.
\newblock


\bibitem[Zhang et~al\mbox{.}(2020b)]%
        {zhang2020unsupervised}
\bibfield{author}{\bibinfo{person}{Rui Zhang}, \bibinfo{person}{Yunxing Zhang}, {and} \bibinfo{person}{Xuelong Li}.} \bibinfo{year}{2020}\natexlab{b}.
\newblock \showarticletitle{Unsupervised feature selection via adaptive graph learning and constraint}.
\newblock \bibinfo{journal}{\emph{IEEE Transactions on Neural Networks and Learning Systems}} \bibinfo{volume}{33}, \bibinfo{number}{3} (\bibinfo{year}{2020}), \bibinfo{pages}{1355--1362}.
\newblock


\bibitem[Zhang et~al\mbox{.}(2014a)]%
        {zhang2014boxconstrained}
\bibfield{author}{\bibinfo{person}{Xiang Zhang}, \bibinfo{person}{Naiyang Guan}, \bibinfo{person}{Long Lan}, \bibinfo{person}{Dacheng Tao}, {and} \bibinfo{person}{Zhigang Luo}.} \bibinfo{year}{2014}\natexlab{a}.
\newblock \showarticletitle{Box-constrained Projective Nonnegative Matrix Factorization via Augmented {L}agrangian Method}. In \bibinfo{booktitle}{\emph{International Joint Conference on Neural Networks (IJCNN)}}. IEEE, \bibinfo{pages}{1900--1906}.
\newblock


\bibitem[Zhao and Liu(2021)]%
        {zhao2021robust}
\bibfield{author}{\bibinfo{person}{M. Zhao} {and} \bibinfo{person}{J. Liu}.} \bibinfo{year}{2021}\natexlab{}.
\newblock \showarticletitle{Robust clustering with sparse corruption via $\ell_{2,1}$, $\ell_1$ norm constraint and {L}aplacian regularization}.
\newblock \bibinfo{journal}{\emph{Expert Systems with Applications}}  \bibinfo{volume}{186} (\bibinfo{year}{2021}), \bibinfo{pages}{115704}.
\newblock


\bibitem[Zhao et~al\mbox{.}(2019)]%
        {fmri}
\bibfield{author}{\bibinfo{person}{Y. Zhao}, \bibinfo{person}{X. Li}, \bibinfo{person}{H. Huang}, \bibinfo{person}{W. Zhang}, \bibinfo{person}{S. Zhao}, \bibinfo{person}{M. Makkie}, \bibinfo{person}{M. Zhang}, \bibinfo{person}{Q. Li}, {and} \bibinfo{person}{T. Liu}.} \bibinfo{year}{2019}\natexlab{}.
\newblock \showarticletitle{Four-dimensional modeling of fMRI data via spatio-temporal convolutional neural networks (ST-CNNs)}.
\newblock \bibinfo{journal}{\emph{IEEE Transactions on Cognitive and Developmental Systems}} \bibinfo{volume}{12}, \bibinfo{number}{3} (\bibinfo{year}{2019}), \bibinfo{pages}{451--460}.
\newblock


\bibitem[Zhao et~al\mbox{.}(2021)]%
        {8943941pami2021}
\bibfield{author}{\bibinfo{person}{Yang Zhao}, \bibinfo{person}{Huiyang Wang}, {and} \bibinfo{person}{Jihong Pei}.} \bibinfo{year}{2021}\natexlab{}.
\newblock \showarticletitle{Deep Non-Negative Matrix Factorization Architecture Based on Underlying Basis Images Learning}.
\newblock \bibinfo{journal}{\emph{IEEE Transactions on Pattern Analysis and Machine Intelligence}} \bibinfo{volume}{43}, \bibinfo{number}{6} (\bibinfo{year}{2021}), \bibinfo{pages}{1897--1913}.
\newblock


\bibitem[Zhou et~al\mbox{.}(2006)]%
        {zhou2006learning}
\bibfield{author}{\bibinfo{person}{D. Zhou}, \bibinfo{person}{J. Huang}, {and} \bibinfo{person}{B. Sch{\"o}lkopf}.} \bibinfo{year}{2006}\natexlab{}.
\newblock \showarticletitle{Learning with hypergraphs: {C}lustering, classification, and embedding}. In \bibinfo{booktitle}{\emph{Advances in Neural Information Processing Systems}}, Vol.~\bibinfo{volume}{19}.
\newblock


\bibitem[Zhou et~al\mbox{.}(2014)]%
        {ZhouCichocki}
\bibfield{author}{\bibinfo{person}{G. Zhou}, \bibinfo{person}{A. Cichocki}, \bibinfo{person}{Q. Zhao}, {and} \bibinfo{person}{S. Xie}.} \bibinfo{year}{2014}\natexlab{}.
\newblock \showarticletitle{Nonnegative matrix and tensor factorizations: An algorithmic perspective}.
\newblock \bibinfo{journal}{\emph{IEEE Signal Processing Magazine}} \bibinfo{volume}{31}, \bibinfo{number}{3} (\bibinfo{year}{2014}), \bibinfo{pages}{54--65}.
\newblock


\bibitem[Zhou et~al\mbox{.}(2016)]%
        {zhou2016discriminative}
\bibfield{author}{\bibinfo{person}{Nan Zhou}, \bibinfo{person}{Hong Cheng}, \bibinfo{person}{Witold Pedrycz}, \bibinfo{person}{Yong Zhang}, {and} \bibinfo{person}{Huaping Liu}.} \bibinfo{year}{2016}\natexlab{}.
\newblock \showarticletitle{Discriminative sparse subspace learning and its application to unsupervised feature selection}.
\newblock \bibinfo{journal}{\emph{ISA transactions}}  \bibinfo{volume}{61} (\bibinfo{year}{2016}), \bibinfo{pages}{104--118}.
\newblock


\bibitem[Zhou et~al\mbox{.}(2023)]%
        {zhou2023soft}
\bibfield{author}{\bibinfo{person}{Shixuan Zhou}, \bibinfo{person}{Peng Song}, \bibinfo{person}{Zihao Song}, {and} \bibinfo{person}{Liang Ji}.} \bibinfo{year}{2023}\natexlab{}.
\newblock \showarticletitle{Soft-label guided non-negative matrix factorization for unsupervised feature selection}.
\newblock \bibinfo{journal}{\emph{Expert Systems with Applications}}  \bibinfo{volume}{216} (\bibinfo{year}{2023}), \bibinfo{pages}{119468}.
\newblock


\bibitem[Zhu et~al\mbox{.}(2018)]%
        {zhu2018dropping}
\bibfield{author}{\bibinfo{person}{Z. Zhu}, \bibinfo{person}{X. Li}, \bibinfo{person}{K. Liu}, {and} \bibinfo{person}{Q. Li}.} \bibinfo{year}{2018}\natexlab{}.
\newblock \showarticletitle{Dropping Symmetry for Fast Symmetric Nonnegative Matrix Factorization}. In \bibinfo{booktitle}{\emph{Advances in Neural Information Processing Systems}}, Vol.~\bibinfo{volume}{31}.
\newblock


\end{thebibliography}

\end{document}